\documentclass[10pt,twocolumn,letterpaper]{article}

\usepackage{preprint}

\usepackage{amsmath}
\usepackage{amssymb}
\usepackage{booktabs}
\usepackage{multirow}
\usepackage{graphicx}
\usepackage{xspace}
\usepackage{xcolor}

\newcommand{\saysee}{SaySee\xspace}
\newcommand{\say}{Say\xspace}
\newcommand{\see}{See\xspace}
\newcommand{\E}{\mathbb{E}}
\newcommand{\ind}{\mathbf{1}}
\newcommand{\gen}{\mathrm{gen}}
\newcommand{\real}{\mathrm{real}}
\newcommand{\thickrule}{\midrule[0.9pt]}
\DeclareMathOperator*{\softmax}{softmax}

\usepackage{pifont}

\usepackage{placeins}
\usepackage{array}
\usepackage{float}
\makeatletter
\newcommand{\suppfullcol}{\global\@colroom\@colht\global\vsize\@colht}
\makeatother

\newsavebox{\supptabbox}
\newcommand{\fittab}[2][\linewidth]{%
  \sbox{\supptabbox}{#2}%
  \ifdim\wd\supptabbox>#1\relax\resizebox{#1}{!}{\usebox{\supptabbox}}\else\usebox{\supptabbox}\fi}

\definecolor{linkblue}{rgb}{0.21,0.49,0.74}
\usepackage[pagebackref,breaklinks,colorlinks,allcolors=linkblue]{hyperref}
\usepackage[capitalize]{cleveref}
\crefname{section}{Sec.}{Secs.}
\Crefname{section}{Section}{Sections}
\Crefname{table}{Table}{Tables}
\crefname{table}{Tab.}{Tabs.}
\hypersetup{
  pdftitle={Must You See Fakes to Catch Fakes? Language Says What to Look For, Real Images What to Ignore},
  pdfauthor={Tianchi Xu, Yuxuan Zhou, Yuheng Zhang, Yanan Guan, Tao Dai, Shu-Tao Xia},
  pdfsubject={AI-generated image detection},
  pdfkeywords={SaySee, image authenticity, vision-language models}
}

\title{Must You See Fakes to Catch Fakes?\\Language Says What to Look For, Real Images What to Ignore}

\author{%
Tianchi Xu\textsuperscript{1}\quad
Yuxuan Zhou\textsuperscript{1}\quad
Yuheng Zhang\textsuperscript{1}\quad
Yanan Guan\textsuperscript{2}\quad
Tao Dai\textsuperscript{3}\quad
Shu-Tao Xia\textsuperscript{1}\\[0.35em]
\textsuperscript{1}Tsinghua University\qquad
\textsuperscript{2}Nankai University\qquad
\textsuperscript{3}Shenzhen University
}

\begin{document}
\maketitle
\begin{abstract}
Must a detector see generated images in order to catch them?
Detection of AI-generated images has long faced two problems.
The cues a detector learns from the few generators in its training set carry traces specific to those generators and transfer poorly to new ones, which limits generalization.
The real and generated training images also differ in format and content, so the detector takes these differences as evidence of authenticity, and its decisions change with factors that have nothing to do with authenticity, which harms robustness as well as generalization.
Both problems stem from training on generated images.
A detector never trained on them can largely avoid both at the source.
Vision foundation models have already learned the concept of ``AI-generated'' during pretraining, and text can point to it directly.
Building on this, we propose \saysee, a training-free detector: \say builds a generation direction from sets of paired descriptions that differ only in authenticity and scores an image by the projection of its embedding onto it, and \see uses real images to correct the score offsets caused by format, image type and content.
On 24 evaluation sets that span standard benchmarks, in-the-wild data and the latest generators, \saysee reaches a mean AUC of 98.2\% with MoE-ViE, 4.2 points above the best existing detector.
When format is decoupled from authenticity, it remains the best of all methods.
Code will be released.
\end{abstract}

\begin{figure}[t]
  \centering
  \includegraphics[width=\columnwidth]{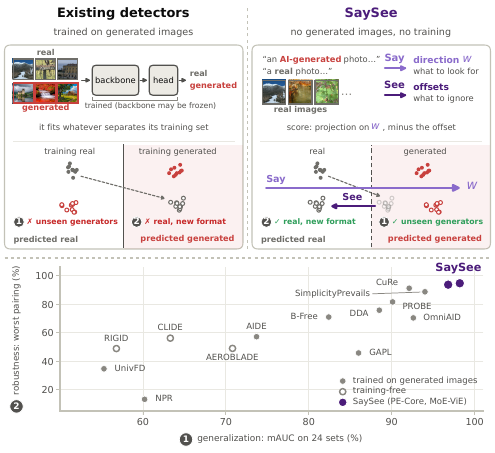}
  \caption{\textbf{The two problems of existing detectors, and \saysee.}
  Top: a trained detector fits its training set, so \ding{172} images of unseen generators fall on the real side and \ding{173} real images in a new format cross the boundary; \saysee uses no generated images: \say gives the generation direction $w$, and \see subtracts the offsets of conditions, measured on real images.
  Bottom: one point per detector; the horizontal axis is the mAUC over the 24 evaluation sets (\ding{172}), the vertical axis the mAUC of the worst pairing (\ding{173}, \cref{sec:robust}).}
  \label{fig:overview}
\end{figure}

\section{Introduction}
\label{sec:intro}

Images produced by diffusion and multimodal generative models have become hard to tell apart from real photographs~\cite{rombach2022ldm,betker2023dalle3,esser2024sd3}, which makes the detection of AI-generated images increasingly important~\cite{epstein2023art,lin2024detecting}.
Two problems have long been discussed in this field.
The first is generalization: the traces a generator leaves in its images are partly specific to it~\cite{marra2019gans,yu2019attributing,wang2020cnnspot,ricker2024towards}, a decision rule learned from the few generators of a training set inevitably carries this specific part, and it degrades on new generators~\cite{ojha2023univfd,zhu2023genimage}.
The second is robustness: the real and generated images of a training set are not aligned in format and content (real images are mostly JPEG, generated images mostly PNG~\cite{grommelt2024fake}), detectors learn these differences as evidence of authenticity as well~\cite{grommelt2024fake,zheng2024breaking}, and their decisions therefore change under compression and other processing unrelated to authenticity.
Aligning the training data by reconstructing real images with a generative model~\cite{rajan2025aligned,chen2024drct,guillaro2025bfree,chen2025dda} in turn brings in traces of the reconstruction process and of the generator used for it.

Both problems arise because the detector learns from generated images, and not training on them largely avoids both, provided that the backbone itself knows what ``AI-generated'' means.
A linear classifier on the frozen features of a vision foundation model is already a strong detector~\cite{ojha2023univfd,cozzolino2024raising,simplicityprevails2026}, which indicates that these models have learned the concept of ``AI-generated'' during pretraining.
We therefore try to read this concept out directly instead of learning it from generated images.

We build a generation direction in the embedding space of a vision--language model from a pair of descriptions that differ only in authenticity, and score an image by the projection of its embedding onto this direction.
Surprisingly, the direction obtained from the two sentences ``an AI-generated image'' and ``a real image'' alone already outperforms most existing detectors on both PE-Core and MoE-ViE (\cref{sec:direction}).
The direction, however, also absorbs factors unrelated to authenticity, such as image type, content and format (\cref{sec:bias}); the first two can be expressed in language, while format cannot.
\saysee (\cref{fig:overview}) therefore consists of two modules.
\say builds the generation direction from many paired descriptions that state the image type and content and differ only in authenticity; \see measures, on real images, the score offset caused by each kind of condition and subtracts it from the score.

On 24 evaluation sets covering standard benchmarks, in-the-wild data and the latest generators, \saysee reaches, without any training, a mean AUC of 98.2\% and 96.8\% with MoE-ViE and PE, well above existing detectors (best 94.0\%) and above a linear classifier and LoRA fine-tuning trained on the same backbones.
Its lead is largest on the latest generators (96.3\% with MoE-ViE, at most 90.0\% for existing detectors), and when format is decoupled from authenticity it is still the best of all methods under the least favorable format pairing (94.6\% with MoE-ViE, at most 91.1\% for existing detectors).
These two results correspond to the generalization and the robustness problem.

Our contributions are as follows.
\begin{itemize}
\item We show that a generation direction built from descriptions that differ only in authenticity is by itself a strong detector, and we analyze its bias.
\item We propose \saysee, a training-free detector in which factors that language can express are paired in the descriptions and factors it cannot express are corrected with real images.
\item On 24 evaluation sets, \saysee outperforms all compared methods, including those of 2026, and it also performs best when format is decoupled from authenticity.
\end{itemize}

\section{Related Work}
\label{sec:related}

\paragraph{Generalizable detection of AI-generated images.}
Early work trained detectors on images of a single generator and relied on data augmentation to reach generators of the same family~\cite{wang2020cnnspot}.
For face deepfake detection, CRDA~\cite{zhou2026crda} uses reinforcement learning to adapt data augmentation during training and improve cross-domain generalization.
Other studies searched for more universal generation traces, such as GAN fingerprints~\cite{marra2019gans,yu2019attributing}, spectral artifacts~\cite{frank2020leveraging} and relations between neighboring pixels~\cite{tan2024npr}, or extended the training data to thousands of generators~\cite{park2025cf}.
Benchmarks such as GenImage~\cite{zhu2023genimage} showed that such traces often no longer hold for generators unseen in training.
With diffusion models, reconstruction error became an important cue: DIRE~\cite{wang2023dire} trains a classifier on diffusion reconstruction errors, and AEROBLADE~\cite{ricker2024aeroblade} scores images directly by the reconstruction error of an autoencoder.
CLIDE~\cite{clide2026} models real images only and flags images that deviate from the model.
Like \saysee, the last two do not train on generated images, but their scores are hand-designed statistics that carry no knowledge about generated images.

\paragraph{Dataset bias and alignment.}
Detectors exploit differences between real and generated training images that are unrelated to authenticity, such as JPEG compression~\cite{corvi2023detection,grommelt2024fake}, image size~\cite{rajan2025aligned} and semantic content~\cite{zheng2024breaking}.
One line of work therefore reconstructs real images with a generative model to build aligned generated images, either through the autoencoder alone~\cite{rajan2025aligned} or through a full diffusion reconstruction~\cite{chen2024drct,guillaro2025bfree}; DDA~\cite{chen2025dda} also compresses each reconstruction with the JPEG quality of its real counterpart.
A reconstruction, however, first encodes a real image, and a diffusion reconstruction also adds noise, whereas actual generation only denoises and decodes, so the two still differ beyond generation traces; the generative model used for reconstruction also becomes something the detector learns.
\saysee uses no generated images and needs no alignment.

\paragraph{Detection with vision foundation models.}
Ojha \etal~\cite{ojha2023univfd} found that linear classification on frozen CLIP~\cite{radford2021clip} features generalizes better than detectors trained end to end, and Cozzolino \etal~\cite{cozzolino2024raising} showed that a linear classifier trained on CLIP features with a few paired images also beats specialized detectors.
Later work either improves how frozen features are read out or fine-tunes the backbone, for example with prompt learning or LoRA~\cite{chen2025dda,cure2026}.
SimplicityPrevails~\cite{simplicityprevails2026} found that a linear head on a frozen vision foundation model of the latest generation already surpasses earlier detectors, and attributes this to the many generated images these models have seen during pretraining.
All these methods need labeled generated images.
\saysee instead needs no training and reads out what the foundation model already knows.

\begin{figure*}[t]
  \centering
  \includegraphics[width=\textwidth]{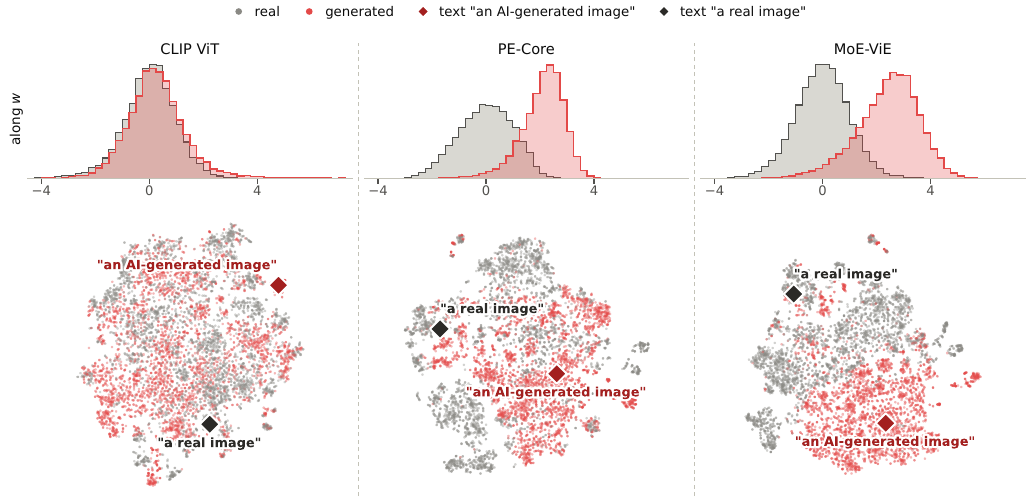}
  \caption{\textbf{Image embeddings of three backbones.} Top: score distributions of the two classes along the generation direction of the minimal pair, centered on the mean of the real images and in units of their standard deviation, with equal weight per evaluation set. Bottom: t-SNE; gray points are real images and red points generated images (sampled from the 24 evaluation sets); the two diamonds are the sentences of the minimal pair, drawn at the median position of the 50 images with the highest and the lowest score.}
  \label{fig:tsne}
\end{figure*}

\section{Preliminaries}
\label{sec:prelim}

\subsection{Zero-shot detection with the generation direction}
\label{sec:direction}

Let $f_T$ be the text encoder of a frozen vision--language model, $z(x)$ the embedding of image $x$, and $t_g$ and $t_r$ texts that describe a generated and a real image and differ only in authenticity.
The difference of their text embeddings,
\begin{equation}
  w = f_T(t_g) - f_T(t_r),
  \label{eq:direction}
\end{equation}
is called the generation direction, and the score is the projection of the image embedding onto it, $s(x)=\langle z(x), w\rangle$.

\begin{figure*}[t]
  \centering
  \includegraphics[width=\textwidth]{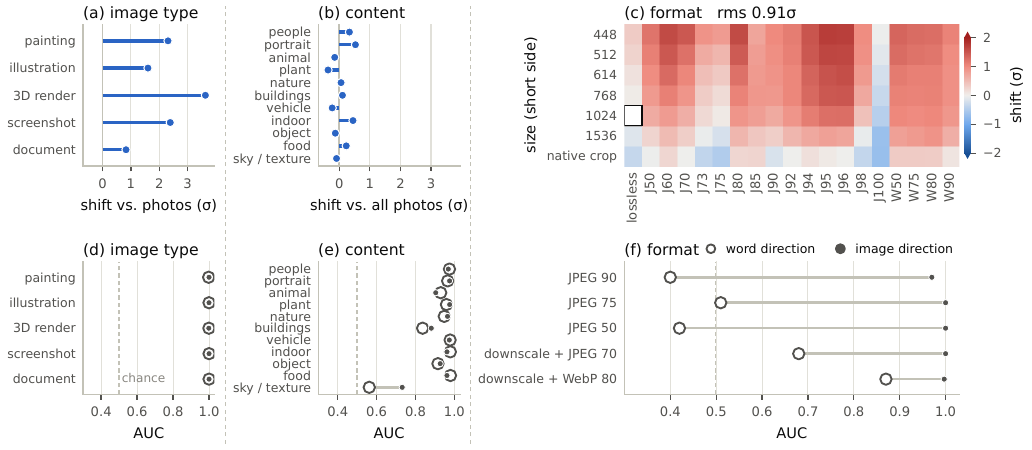}
  \caption{\textbf{Bias of the minimal pair, and whether language can express the three kinds of conditions (MoE-ViE).} Top row: shifts of the scores of real images, in units of the standard deviation $\sigma$ of the scores of real photographs; (a) five non-photographic image types relative to photographs, (b) each content theme relative to all photographs, (c) 80 photographs under 7 sizes $\times$ 19 codecs relative to the reference condition (boxed). Bottom row (d)--(f): AUC of separating two groups of real images with and without a factor along the word direction (rings) and along the image direction (dots).}
  \label{fig:bias}
\end{figure*}

With the simplest pair, $t_g=$ ``an AI-generated image'' and $t_r=$ ``a real image'' (the \emph{minimal pair} below), the mean AUC over the 24 evaluation sets already reaches 95.7\% with PE, higher than all 11 trained detectors we compare against (best 94.0\%); with MoE-ViE it reaches 93.5\%.
The same minimal pair reaches only 55.7\% with CLIP ViT-L/14, in line with the earlier conclusion that a single prompt is not sufficient for zero-shot detection with CLIP~\cite{willi2026synthclic}.
\Cref{fig:tsne} shows where the difference lies: in the CLIP feature space real and generated images are mixed, whereas in those of PE and MoE-ViE they are largely separated, and the two sentences fall in the regions of the corresponding images; along the direction of the minimal pair, the mean of the generated images lies only 0.3 standard deviations above that of the real images with CLIP, and 2.1 and 2.4 standard deviations above with PE-Core and MoE-ViE.
One possible explanation is greater exposure to AI-generated images during pretraining.

\subsection{Bias of the generation direction}
\label{sec:bias}

The two sentences of \cref{eq:direction} differ only in authenticity, yet the generation direction reflects more than authenticity.
Text embeddings are learned from image--text pairs, and if AI-generated and real images in the pretraining data are distributed differently over other conditions (for example, AI-generated images include more paintings and illustrations), these differences also enter $w$.
Let $e(s,c)$ be the text embedding that describes an image of authenticity $s$ under condition $c$, written as $e(s,c)=a\,\ind[s=\gen]+u(c)+v(c)\,\ind[s=\gen]$, with $a$ the effect of authenticity, $u(c)$ the effect of the condition and $v(c)$ their interaction.
A description without a condition amounts to an average over its own condition distribution, $p_g$ or $p_r$, so
\begin{equation}
  w \approx a + \E_{c\sim p_g}[v(c)] + \bigl(\E_{c\sim p_g}[u(c)] - \E_{c\sim p_r}[u(c)]\bigr).
  \label{eq:bias}
\end{equation}
Whenever $p_g\neq p_r$, $w$ absorbs the effect $u(c)$ of the condition itself.
If both descriptions state the same condition $c$, their difference $a+v(c)$ no longer contains $u(c)$.
In the score, this bias appears as the shift that condition $c$ causes in the scores of real images, $\beta(c)=\E[s\mid\real,c]-\E[s\mid\real,c_0]$ for a reference condition $c_0$, which can be measured with real images alone; the additional effect of a condition on generated images can only be measured with generated images, and we leave it aside.

The top row of \cref{fig:bias} measures the bias of the minimal pair with real images for three kinds of conditions.
The bias is largest for image type, with real paintings, illustrations, 3D renders, screenshots and document pages scoring 0.8 to 3.6 standard deviations higher than real photographs.
For content it is smaller, within 0.6 standard deviations for all 11 themes.
Format also shifts the scores systematically. Saving the same photographs at different sizes and with different codecs yields a root mean square offset of 0.91 standard deviations over the 132 non-reference conditions.

Image type and content can be expressed in language: when two groups of real images with and without a factor are separated, the word direction does as well as the image direction (\cref{fig:bias}, bottom row).
Format cannot.
Compressed and uncompressed images are almost perfectly separable in the image embedding, in line with the finding that vision foundation models encode compression quality linearly~\cite{invisibleshortcuts2026}, yet the word directions built from format names hardly separate them: for JPEG compression at three quality levels, their AUCs lie between 0.40 and 0.51.
This gives the design principle of \saysee: factors that language can express are fixed by pairing them in the descriptions, and factors it cannot express are measured and corrected with real images.
Neither can replace the other: real images only tell how a condition shifts the scores of real images, not where the generated images under the same condition lie (\cref{sec:ablation}).

\begin{figure*}[t]
  \centering
  \includegraphics[width=\textwidth]{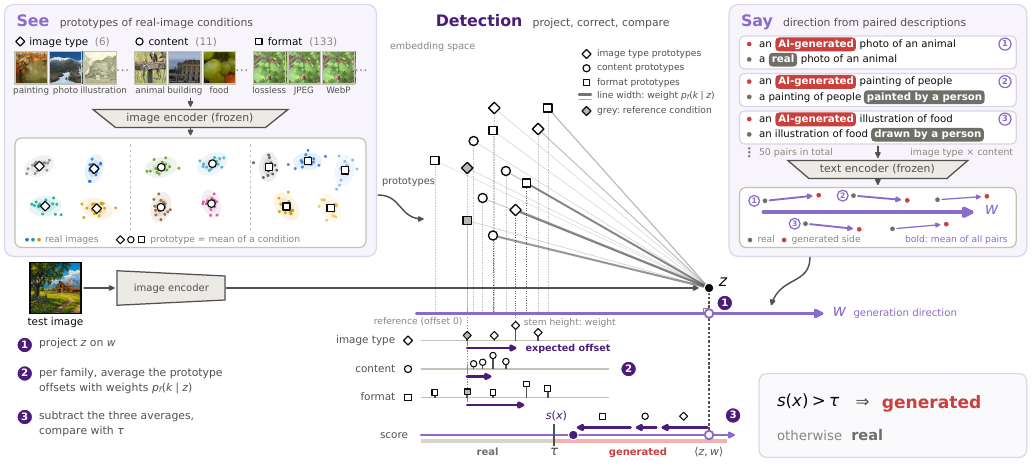}
  \caption{\textbf{Pipeline of \saysee} (schematic).
  Top left (\see): real images of three families of conditions pass the frozen image encoder; the mean of each condition is its prototype.
  Top right (\say): the two descriptions of a pair differ only in the highlighted words; each pair gives one direction, and their mean is $w$.
  Center (detection): \ding{172} $z$ is projected onto $w$; \ding{173} in each family, the offsets of the prototypes (their positions along $w$; gray: reference, zero) are averaged with the posterior $p_f(k\mid z)$ (line width); \ding{174} the three averages are subtracted and the result is compared with $\tau$.}
  \label{fig:pipeline}
\end{figure*}

\section{Method}
\label{sec:method}

\subsection{Overview}
\label{sec:overview}

The score of \saysee is
\begin{equation}
  s(x) = \langle z(x), w\rangle - b\bigl(z(x)\bigr).
  \label{eq:saysee}
\end{equation}
The first term is the projection of the image embedding onto the generation direction $w$, which \say builds from paired descriptions (\cref{sec:say}); the second term $b(z)$ comes from \see and is the expected score offset caused by the conditions of the image (\cref{sec:see}).
Both modules are computed once when the detector is built; at test time, detection takes one image encoding and a few matrix operations (\cref{fig:pipeline}), and the score is compared with a threshold $\tau$, chosen once as the value that maximizes balanced accuracy on a set of public images that does not overlap the evaluation sets (5,000 real and 5,000 generated).

\subsection{\say: generation direction from paired descriptions}
\label{sec:say}

\say writes the image type and content into both sides of every description pair, so that each pair differs only in authenticity.
The set of image types $M$ contains photograph, painting, illustration, 3D render, screenshot and document page; each of the first four is combined with 11 content themes and ``unspecified'', while screenshots and document pages take only ``unspecified'', which gives 50 combinations, and $C_m$ denotes the content options of image type $m$.
For combination $(m,c)$, the generated side is given by four templates (\eg, ``an AI-generated painting of an animal'') and by the names of 18 popular generators (\eg, ``a painting generated by Midjourney''), and the real side by two or three templates of that image type (\eg, ``a painting of an animal made by a human artist'').
Let $\bar g_{m,c}$, $\bar n_m$ and $\bar r_{m,c}$ be the normalized means of the text embeddings of these three groups of descriptions.
The combination direction and the generation direction are
\begin{align}
  d_{m,c} &= N\bigl(\tfrac12\bar g_{m,c} + \tfrac12\bar n_m\bigr) - \bar r_{m,c}, \label{eq:cell}\\
  w &= \frac{1}{|M|}\sum_{m\in M}\frac{1}{|C_m|}\sum_{c\in C_m} d_{m,c}, \label{eq:say}
\end{align}
where $N(\cdot)$ denotes normalization.
All templates and generator names are listed in the supplementary material.

\begin{figure}[t]
  \centering
  \includegraphics[width=\columnwidth]{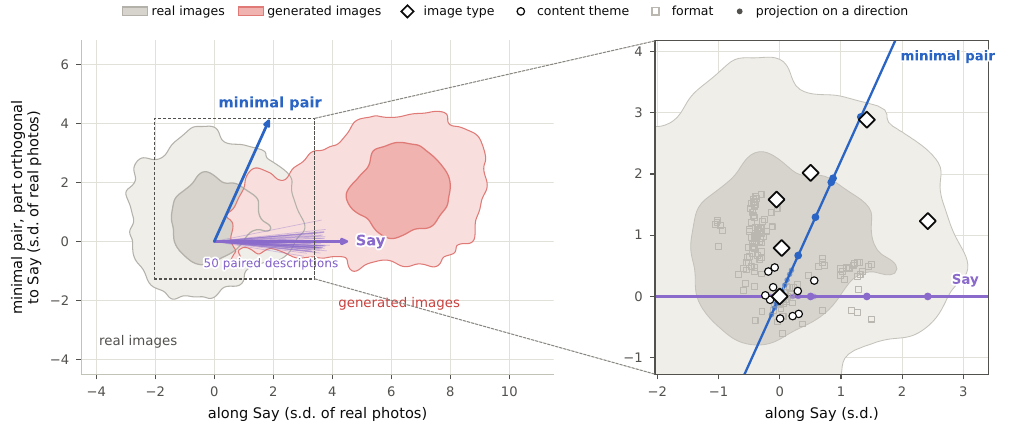}
  \caption{\textbf{What pairing changes in the generation direction (MoE-ViE).} The plane is spanned by the \say direction (horizontal) and the part of the minimal-pair direction orthogonal to it (vertical), in units of the standard deviation of real photographs along the horizontal axis. Left: the light purple arrow is the \say direction, thin lines are the 50 combination directions, and the blue arrow is the minimal-pair direction; gray and red regions show the distributions of real and generated images. Right: the dashed box enlarged; diamonds, circles and hollow squares are the means of real images of each image type, content theme and format condition, and solid dots are their projections onto the two lines.}
  \label{fig:plane}
\end{figure}

Pairing clearly changes the generation direction (\cref{fig:plane}): the direction of the minimal pair is about $66^\circ$ away from $w$.
Along the minimal pair, real 3D renders score nearly as high as generated images; along \say, paintings, illustrations and document pages return close to photographs, while generated images move further away.
The means of the format conditions spread to a similar extent along both directions; pairing has no effect on format, which is left to \see.

\subsection{\see: condition correction from real images}
\label{sec:see}

\see measures, with real images, the biases that language cannot express and subtracts them from the score.
The condition of a test image is inferred from its embedding; file format, quality and size are never read.

\paragraph{Conditions and offsets.}
We consider three families of conditions $f$: format (7 sizes $\times$ 19 codecs, 133 conditions), image type (the 6 types of \say) and content (11 themes).
Each condition $k$ corresponds to a set of real images whose mean embedding is the prototype $\mu_{f,k}$, and the offset $o_{f,k}$ is the difference between the mean score of these images along $w$ and that of the reference condition.
Format offsets are measured on paired versions of the same photographs,
\begin{equation}
  o_{\mathrm{fmt},k} = \E_i\bigl[\langle z(T_k(x_i)), w\rangle - \langle z(T_0(x_i)), w\rangle\bigr],
  \label{eq:offset}
\end{equation}
where $T_k$ converts photograph $x_i$ to condition $k$ and $T_0$ is the reference condition (short side resized to 1024, saved losslessly).
Image-type and content offsets are measured between different images, with camera photographs as reference; since the two groups may differ in more than the condition at hand, we only correct the part that pushes real images toward the generated side, $\tilde o_{f,k}=\max(o_{f,k},0)$.
For format, $\tilde o=o$.

\paragraph{Condition inference and correction.}
The posterior over the conditions of each family is given by a Gaussian discriminant model with shared covariance,
\begin{equation}
  p_f(k\mid z) = \softmax_k\Bigl[\lambda_f\bigl(\mu_{f,k}^\top\Sigma_f^{-1}z - \tfrac12\mu_{f,k}^\top\Sigma_f^{-1}\mu_{f,k}\bigr)\Bigr],
  \label{eq:posterior}
\end{equation}
where $\Sigma_f$ is the within-condition covariance of the real images of the family (Ledoit--Wolf shrinkage~\cite{ledoit2004well}) and the temperature $\lambda_f$ is chosen on held-out real images with known conditions.
The correction is the expected offset under the posterior, summed over the three families:
\begin{equation}
  b(z) = \sum_f\sum_k p_f(k\mid z)\,\tilde o_{f,k}.
  \label{eq:correction}
\end{equation}
Since the condition is inferred, the correction takes the expectation rather than a hard decision on one condition.

\section{Experiments}
\label{sec:exp}

\subsection{Experimental setup}
\label{sec:setup}

\begin{table*}[t]
\centering
\caption{\textbf{AUC / bAcc (\%) on the standard benchmarks.} The best AUC and the best bAcc are in bold. Rows above the thick rule are classic methods and methods published in 2025; rows below are methods of 2026, including arXiv-only ones.}
\label{tab:standard}
\resizebox{\textwidth}{!}{%
\begin{tabular}{lccccccc|c}
\toprule
Method & GenImage & AIGCDB & DRCT-2M & Synthbuster & GenImage++ & B-Free-NG & SynthCLIC & Mean \\
 & \cite{zhu2023genimage} & \cite{zhong2023patchcraft} & \cite{chen2024drct} & \cite{bammey2024synthbuster} & \cite{omat2025} & \cite{guillaro2025bfree} & \cite{willi2026synthclic} & \\
\midrule
UnivFD~\cite{ojha2023univfd} & 89.1 / 70.3 & 91.8 / 78.7 & 87.5 / 62.2 & 79.8 / 66.8 & 60.8 / 55.0 & 65.7 / 58.7 & 57.0 / 50.5 & 76.0 / 63.2 \\
NPR~\cite{tan2024npr} & 98.6 / 94.4 & 94.4 / 92.2 & 34.6 / 49.2 & 36.6 / 37.0 & 99.0 / 96.6 & 54.1 / 49.9 & 93.0 / 84.5 & 72.9 / 72.0 \\
AIDE~\cite{yan2025aide} & 96.3 / 86.9 & 94.1 / 81.8 & 93.5 / 58.9 & 59.9 / 54.2 & 97.8 / 86.7 & 69.7 / 63.2 & 75.5 / 67.3 & 83.8 / 71.3 \\
B-Free~\cite{guillaro2025bfree} & 96.9 / 88.8 & 97.3 / 90.2 & \textbf{100.0} / 99.3 & 99.9 / 98.0 & 94.2 / 83.9 & 99.6 / 95.2 & 83.5 / 74.4 & 95.9 / 90.0 \\
DDA~\cite{chen2025dda} & 98.0 / 89.3 & 94.7 / 86.8 & 99.8 / 97.4 & 96.9 / 89.9 & 98.6 / 93.1 & 98.9 / 92.5 & 92.2 / 84.0 & 97.0 / 90.4 \\
\thickrule
GAPL~\cite{gapl2026} & 99.5 / 96.7 & \textbf{99.7} / 97.0 & 99.8 / 97.0 & 96.8 / 90.9 & 99.1 / 96.1 & 98.8 / 93.0 & 95.3 / 89.3 & 98.4 / 94.3 \\
DEAR-c~\cite{dear2026} & 93.5 / 88.9 & 83.6 / 77.4 & 99.9 / 99.4 & 84.9 / 84.5 & \textbf{100.0} / 98.4 & \textbf{100.0} / 98.6 & \textbf{99.7} / \textbf{95.2} & 94.5 / 91.8 \\
PROBE (DINOv2)~\cite{probe2026} & \textbf{99.6} / 96.2 & 98.1 / 90.5 & \textbf{100.0} / 99.7 & 99.9 / 98.1 & 98.3 / 82.7 & \textbf{100.0} / 98.6 & 95.3 / 79.4 & 98.7 / 92.2 \\
OmniAID v1~\cite{omniaid2026} & 99.1 / \textbf{97.2} & 98.9 / 92.6 & 98.0 / 90.8 & 88.9 / 65.0 & 98.8 / 95.0 & 79.3 / 62.1 & 96.2 / 87.9 & 94.2 / 84.4 \\
SimplicityPrevails (DINOv3)~\cite{simplicityprevails2026} & 99.5 / 96.7 & 98.7 / 94.0 & 99.8 / 97.0 & 98.7 / 95.8 & 99.0 / 94.3 & 98.4 / 91.8 & 89.0 / 80.8 & 97.6 / 92.9 \\
CuRe~\cite{cure2026} & 97.0 / 88.0 & 97.5 / 89.3 & \textbf{100.0} / 99.6 & 98.7 / 93.7 & 93.5 / 89.2 & 94.1 / 82.8 & 85.8 / 80.5 & 95.2 / 89.0 \\
\textbf{\saysee (PE)} & 99.3 / 96.8 & 99.3 / 96.4 & 98.7 / 93.9 & 99.9 / 99.0 & 99.8 / 98.7 & \textbf{100.0} / 99.4 & 97.3 / 90.3 & 99.2 / 96.4 \\
\textbf{\saysee (MoE-ViE)} & \textbf{99.6} / 97.1 & \textbf{99.7} / \textbf{98.3} & \textbf{100.0} / \textbf{100.0} & \textbf{100.0} / \textbf{99.8} & 99.9 / \textbf{99.3} & \textbf{100.0} / \textbf{99.7} & 98.7 / 94.2 & \textbf{99.7} / \textbf{98.3} \\
\bottomrule
\end{tabular}}
\end{table*}

\begin{table*}[t]
\centering
\caption{\textbf{AUC / bAcc (\%) on the in-the-wild sets.} Notation as in \cref{tab:standard}.}
\label{tab:wild}
\resizebox{\textwidth}{!}{%
\begin{tabular}{lccccccccc|c}
\toprule
Method & Chameleon & WildRF & SocialRF & CommunityAI & B-Free-Online & CO-SPY-Wild & OpenFake-Reddit & RRDataset & HiRes-50K & Mean \\
 & \cite{yan2025aide} & \cite{cavia2024wildrf} & \cite{aigibench2025} & \cite{aigibench2025} & \cite{guillaro2025bfree} & \cite{cheng2025cospy} & \cite{openfake2025} & \cite{rrdataset2025} & \cite{hires50k} & \\
\midrule
UnivFD~\cite{ojha2023univfd} & 51.6 / 50.5 & 57.0 / 55.4 & 54.2 / 54.8 & 56.7 / 50.5 & 58.3 / 48.6 & 42.1 / 49.9 & 27.7 / 39.4 & 45.4 / 47.8 & 53.4 / 54.0 & 49.6 / 50.1 \\
NPR~\cite{tan2024npr} & 34.6 / 53.1 & 68.4 / 66.3 & 62.6 / 60.4 & 55.7 / 54.6 & 53.8 / 50.3 & 52.9 / 49.2 & 40.2 / 50.3 & 64.6 / 64.6 & 38.5 / 50.2 & 52.4 / 55.5 \\
AIDE~\cite{yan2025aide} & 73.2 / 58.4 & 72.8 / 65.4 & 61.9 / 58.8 & 78.6 / 59.1 & 59.2 / 50.6 & 87.0 / 58.8 & 43.8 / 48.5 & 64.9 / 60.6 & 71.5 / 55.5 & 68.1 / 57.3 \\
B-Free~\cite{guillaro2025bfree} & 81.5 / 75.3 & 97.6 / 89.8 & 92.2 / 85.8 & 89.6 / 81.7 & 95.0 / 88.8 & 85.0 / 74.2 & 65.5 / 65.7 & 73.4 / 71.5 & 89.7 / 83.4 & 85.5 / 79.6 \\
DDA~\cite{chen2025dda} & 90.8 / 81.1 & 94.9 / 88.4 & 88.9 / 81.2 & 95.0 / 84.7 & 99.1 / 94.4 & 87.6 / 77.8 & 83.0 / 77.0 & 80.5 / 74.2 & 92.9 / 84.0 & 90.3 / 82.5 \\
\thickrule
GAPL~\cite{gapl2026} & 78.6 / 70.8 & 94.2 / 84.1 & 91.3 / 80.2 & 85.5 / 76.8 & 74.1 / 65.9 & 62.3 / 55.9 & 79.8 / 70.9 & 84.0 / 77.1 & 88.4 / 79.5 & 82.0 / 73.5 \\
DEAR-c~\cite{dear2026} & 86.2 / 78.4 & 83.7 / 80.8 & 84.0 / 75.8 & 89.3 / 80.6 & 79.3 / 73.2 & 20.8 / 47.7 & 92.0 / 86.1 & 81.1 / 78.7 & 91.1 / 85.9 & 78.6 / 76.4 \\
PROBE (DINOv2)~\cite{probe2026} & 93.3 / 85.9 & 99.0 / 95.4 & 95.8 / 90.1 & 96.8 / 91.2 & 96.3 / 91.6 & 92.6 / 74.2 & 88.8 / 83.1 & 76.5 / 75.1 & 94.9 / 89.2 & 92.7 / 86.2 \\
OmniAID v1~\cite{omniaid2026} & 97.0 / 90.4 & 95.9 / 89.3 & 94.4 / 86.4 & 99.0 / 95.7 & 95.3 / 83.2 & 97.0 / 87.5 & 92.0 / 83.2 & 77.8 / 71.9 & 94.5 / 81.9 & 93.6 / 85.5 \\
SimplicityPrevails (DINOv3)~\cite{simplicityprevails2026} & 97.2 / 91.1 & 99.4 / 96.4 & 99.0 / 93.8 & 99.0 / 95.1 & 98.7 / 92.5 & 99.7 / 96.6 & 92.4 / 86.3 & 91.3 / 86.1 & \textbf{97.9} / 92.0 & 97.2 / 92.2 \\
CuRe~\cite{cure2026} & 98.0 / 94.7 & 99.2 / 96.4 & 98.0 / 93.2 & 97.9 / 94.6 & 97.0 / 89.7 & 99.6 / 97.7 & 87.8 / 81.1 & 89.6 / 82.5 & 96.3 / 91.8 & 95.9 / 91.3 \\
\textbf{\saysee (PE)} & 97.1 / 90.8 & 99.4 / 97.2 & 98.5 / 94.5 & 98.5 / 92.4 & 92.2 / 70.4 & 98.1 / 93.2 & 94.0 / 88.3 & 96.1 / 90.1 & 94.7 / 88.6 & 96.5 / 89.5 \\
\textbf{\saysee (MoE-ViE)} & \textbf{99.5} / \textbf{97.3} & \textbf{99.7} / \textbf{98.8} & \textbf{99.3} / \textbf{97.4} & \textbf{99.8} / \textbf{97.8} & \textbf{99.4} / \textbf{96.4} & \textbf{100.0} / \textbf{99.4} & \textbf{95.4} / \textbf{89.3} & \textbf{96.7} / \textbf{92.0} & 97.5 / \textbf{92.1} & \textbf{98.6} / \textbf{95.6} \\
\bottomrule
\end{tabular}}
\end{table*}

\begin{table*}[t]
\centering
\caption{\textbf{AUC / bAcc (\%) on the latest-generator sets.} Notation as in \cref{tab:standard}.}
\label{tab:latest}
\resizebox{\textwidth}{!}{%
\begin{tabular}{lcccccccc|c}
\toprule
Method & DailyBench & So-Fake-OOD & OpenFake-Test & AIGenImages2026 & AIGI-Now & MLLMGenSet & RealChain & Mirage-Test & Mean \\
 & \cite{dailybench} & \cite{sofake2025} & \cite{openfake2025} & \cite{aigenimages2026} & \cite{aiginow} & \cite{mllmgenset} & \cite{realchain} & \cite{omniaid2026} & \\
\midrule
UnivFD~\cite{ojha2023univfd} & 48.9 / 50.0 & 31.3 / 41.8 & 51.3 / 49.9 & 45.3 / 49.5 & 64.6 / 52.4 & 56.4 / 50.9 & 38.6 / 43.6 & 11.7 / 31.4 & 43.5 / 46.2 \\
NPR~\cite{tan2024npr} & 61.9 / 57.1 & 66.2 / 70.3 & 61.5 / 68.8 & 73.1 / 73.3 & 68.4 / 63.8 & 56.3 / 62.4 & 43.7 / 50.0 & 31.3 / 49.9 & 57.8 / 61.9 \\
AIDE~\cite{yan2025aide} & 65.0 / 52.2 & 65.3 / 61.4 & 83.4 / 66.3 & 68.0 / 66.0 & 90.6 / 62.7 & 73.6 / 61.1 & 71.6 / 52.9 & 52.7 / 48.6 & 71.3 / 58.9 \\
B-Free~\cite{guillaro2025bfree} & 69.4 / 62.5 & 65.9 / 63.7 & 85.4 / 77.8 & 77.9 / 72.2 & 81.1 / 61.4 & 52.3 / 52.4 & 67.2 / 65.3 & 37.7 / 53.4 & 67.1 / 63.6 \\
DDA~\cite{chen2025dda} & 76.3 / 67.0 & 86.5 / 80.7 & 91.5 / 83.1 & 92.3 / 84.9 & 86.1 / 69.7 & 69.5 / 63.5 & 72.5 / 64.4 & 58.0 / 60.4 & 79.1 / 71.7 \\
\thickrule
GAPL~\cite{gapl2026} & 79.8 / 71.3 & 81.4 / 73.7 & 88.9 / 80.8 & 86.9 / 78.3 & 92.5 / 73.0 & 72.1 / 66.5 & 71.5 / 63.2 & 62.7 / 57.6 & 79.5 / 70.5 \\
DEAR-c~\cite{dear2026} & \textbf{97.1} / \textbf{91.6} & 91.6 / 86.2 & 94.7 / 89.4 & 91.1 / 85.6 & 77.7 / 72.7 & 72.5 / \textbf{71.9} & 87.5 / 79.4 & 90.6 / 82.6 & 87.9 / 82.4 \\
PROBE (DINOv2)~\cite{probe2026} & 82.4 / 70.9 & 86.4 / 79.7 & 96.5 / 85.6 & 86.0 / 78.2 & 90.8 / 65.8 & 68.3 / 63.4 & 75.4 / 69.1 & 51.3 / 58.5 & 79.6 / 71.4 \\
OmniAID v1~\cite{omniaid2026} & 79.3 / 72.1 & 93.5 / 86.9 & 96.6 / 88.5 & 92.4 / 85.9 & 96.4 / 80.8 & 75.1 / 67.8 & 88.4 / 61.4 & \textbf{98.5} / 88.9 & 90.0 / 79.0 \\
SimplicityPrevails (DINOv3)~\cite{simplicityprevails2026} & 82.5 / 76.2 & 87.7 / 81.6 & 95.9 / 85.8 & 89.0 / 82.9 & 96.5 / 87.2 & 61.5 / 51.6 & 90.4 / 78.8 & 94.1 / 87.5 & 87.2 / 78.9 \\
CuRe~\cite{cure2026} & 86.5 / 80.1 & 85.9 / 78.6 & 92.3 / 83.4 & 81.3 / 75.3 & 92.8 / 85.1 & 59.7 / 53.1 & 90.3 / 82.6 & 92.4 / 86.5 & 85.2 / 78.1 \\
\textbf{\saysee (PE)} & 87.7 / 80.0 & 97.3 / 93.2 & \textbf{99.4} / \textbf{95.1} & 94.1 / 87.9 & \textbf{99.2} / \textbf{95.9} & 86.6 / 71.8 & \textbf{98.0} / \textbf{93.5} & 98.3 / 93.3 & 95.1 / \textbf{88.8} \\
\textbf{\saysee (MoE-ViE)} & 95.7 / 89.7 & \textbf{98.2} / \textbf{93.5} & \textbf{99.4} / 94.0 & \textbf{96.3} / \textbf{91.3} & 98.8 / 91.4 & \textbf{87.5} / 58.8 & 96.1 / 87.0 & 98.1 / \textbf{94.2} & \textbf{96.3} / 87.5 \\
\bottomrule
\end{tabular}}
\end{table*}

\begin{table}[t]
\centering
\caption{\textbf{Comparison with training-free methods (AUC, \%).} CLIDE's default reference images overlap some evaluation sets; results without them are in the supplementary material.}
\label{tab:trainingfree}
\resizebox{\columnwidth}{!}{%
\begin{tabular}{lcccc}
\toprule
Method & Standard & In-the-wild & Latest & mAUC \\
\midrule
RIGID~\cite{he2024rigid} & 65.9 & 56.0 & 49.6 & 56.8 \\
CLIDE~\cite{clide2026} & 62.7 & 70.4 & 55.8 & 63.3 \\
AEROBLADE~\cite{ricker2024aeroblade} & 77.2 & 69.9 & 66.1 & 70.8 \\
\textbf{\saysee (PE)} & 99.2 & 96.5 & 95.1 & 96.8 \\
\textbf{\saysee (MoE-ViE)} & \textbf{99.7} & \textbf{98.6} & \textbf{96.3} & \textbf{98.2} \\
\bottomrule
\end{tabular}}
\end{table}

\paragraph{Evaluation sets.}
We evaluate on 24 public evaluation sets in three groups: 7 standard benchmarks, 9 in-the-wild sets (images collected mostly from the web and social platforms) and 8 latest-generator sets (generated images from recently released generators).
Together they contain more than 90 generators, from early GANs such as ProGAN to GPT Image 2, Nano Banana 2 and FLUX.2.
To our knowledge, this is among the broadest evaluations of AI-generated image detection to date.
Their sources and preprocessing are given in the supplementary material.

\paragraph{Metrics.}
We report the mean of the per-set AUCs (mAUC) and balanced accuracy (bAcc).
\saysee uses the fixed threshold of \cref{sec:overview}, and trained detectors use the thresholds given by their authors; training-free methods come without thresholds, so for them we report AUC only.

\paragraph{Implementation.}
\saysee uses two frozen vision--language models, PE-Core-G/14~\cite{bolya2025pe} and MoE-ViE-H/14~\cite{moevie2026}, both with input size 448, with MoE-ViE as the main backbone.
Each image is cut into five $448\times448$ crops at native resolution (center and four corners), and the embeddings of the crops are averaged to give the image embedding.
The real images used by \see all come from public datasets and do not overlap the evaluation sets (supplementary material).
Experiments run on an NVIDIA RTX PRO 6000 GPU; we also ran \saysee on a Huawei Ascend NPU (supplementary material).

\subsection{Comparing with the state-of-the-art methods}
\label{sec:comparison}

\paragraph{Compared methods.}
We compare with 11 trained detectors (\cref{tab:standard,tab:wild,tab:latest}) and 3 training-free methods (\cref{tab:trainingfree}).
All results are obtained by us with the official code and weights on the same images (details in the supplementary material).

On the standard benchmarks, most recent detectors are close to saturation (\cref{tab:standard}).
On the in-the-wild sets, whose images have been compressed and resized by platforms, most detectors degrade clearly, while \saysee (MoE-ViE) has the best AUC on 8 of the 9 sets (\cref{tab:wild}).
The latest-generator sets differ most from common training data, and the lead of \saysee is largest there (96.3\% with MoE-ViE, at most 90.0\% for existing detectors; \cref{tab:latest}).
All methods do poorly on MLLMGenSet, for reasons given in the supplementary material.
Over all 24 sets, \saysee reaches an mAUC of 98.2\% (MoE-ViE) and 96.8\% (PE), 4.2 and 2.8 points above the best existing detector, SimplicityPrevails, and a bAcc of 93.7\% and 91.3\% at its fixed threshold, against at most 88.0\% for existing detectors.
In a paired bootstrap over evaluation sets, the 95\% confidence intervals of the differences between \saysee and SimplicityPrevails, OmniAID v1 and CuRe all exclude zero.
The three training-free methods fall well behind (\cref{tab:trainingfree}).

\subsection{Comparison with detectors trained on the same backbone}
\label{sec:samebackbone}

\begin{table}[t]
\centering
\caption{\textbf{Comparison with detectors trained on the same backbone (\%).} bAcc uses threshold 0.5 for the linear classifier and LoRA, and the fixed threshold for \saysee.}
\label{tab:samebackbone}
\resizebox{\columnwidth}{!}{%
\begin{tabular}{llccccc}
\toprule
Backbone & Method & Standard & In-the-wild & Latest & mAUC & bAcc \\
\midrule
\multirow{3}{*}{MoE-ViE} & Linear classifier & 98.9 & 96.4 & 90.6 & 95.2 & 92.3 \\
 & LoRA fine-tuning & 99.2 & 96.7 & 90.7 & 95.4 & 83.0 \\
 & \textbf{SaySee} & \textbf{99.7} & \textbf{98.6} & \textbf{96.3} & \textbf{98.2} & \textbf{93.7} \\
\midrule
\multirow{3}{*}{PE} & Linear classifier & 98.9 & 89.7 & 91.2 & 92.9 & 86.0 \\
 & LoRA fine-tuning & 98.1 & 94.8 & 87.3 & 93.3 & 77.2 \\
 & \textbf{SaySee} & \textbf{99.2} & \textbf{96.5} & \textbf{95.1} & \textbf{96.8} & \textbf{91.3} \\
\bottomrule
\end{tabular}}
\end{table}

\begin{table}[t]
\centering
\caption{\textbf{Evaluation with format decoupled from authenticity (\%).} A lower format gap is better; DEAR was not run. Per-pairing results are in the supplementary material.}
\label{tab:format}
\resizebox{\columnwidth}{!}{%
\begin{tabular}{lccc}
\toprule
Method & mAUC (16 pairings) & Format gap & Worst pairing \\
\midrule
UnivFD & 49.7 & 14.9 & 34.6 \\
NPR & 48.6 & 35.7 & 13.2 \\
AIDE & 69.3 & 13.3 & 57.1 \\
B-Free & 80.4 & 7.6 & 70.9 \\
DDA & 84.1 & 6.9 & 75.7 \\
\thickrule
GAPL & 77.6 & 26.4 & 45.7 \\
PROBE (DINOv2) & 87.4 & 5.7 & 81.5 \\
OmniAID v1 & 89.0 & 12.7 & 70.3 \\
SimplicityPrevails (DINOv3) & 92.1 & 3.0 & 88.6 \\
CuRe & 91.9 & \textbf{0.6} & 91.1 \\
\midrule
\multicolumn{4}{l}{\textit{Training-free methods}}\\
RIGID & 52.0 & 2.4 & 48.8 \\
CLIDE & 62.6 & 6.7 & 56.1 \\
AEROBLADE & 66.5 & 14.6 & 48.9 \\
\textbf{\saysee (PE)} & 95.7 & 2.8 & 93.6 \\
\textbf{\saysee (MoE-ViE)} & \textbf{97.2} & 2.5 & \textbf{94.6} \\
\bottomrule
\end{tabular}}
\end{table}

\begin{table}[t]
\centering
\caption{\textbf{Effect of the two modules (\%).} mAUC is over the 24 evaluation sets; Gap and Worst are the format gap and worst-pairing mAUC (\cref{tab:format}). Per-group results are in the supplementary material.}
\label{tab:modules}
\begingroup
\setlength{\tabcolsep}{3pt}
\resizebox{\columnwidth}{!}{%
\begin{tabular}{l|ccc|ccc}
\toprule
& \multicolumn{3}{c|}{MoE-ViE} & \multicolumn{3}{c}{PE} \\
Method & mAUC & Gap & Worst & mAUC & Gap & Worst \\
\midrule
Minimal pair & 93.48 & 4.05 & 88.52 & 95.72 & 6.56 & 84.78 \\
Minimal pair + See & 93.09 & 4.52 & 87.56 & 94.89 & 4.84 & 88.29 \\
Say & 97.99 & 4.43 & 90.00 & 96.34 & 5.81 & 87.58 \\
Say + See (SaySee) & \textbf{98.15} & \textbf{2.48} & \textbf{94.57} & \textbf{96.80} & \textbf{2.80} & \textbf{93.56} \\
\bottomrule
\end{tabular}}
\endgroup
\end{table}

To compare the direction given by language with directions learned from generated images on the same features, we follow the protocol of SimplicityPrevails, train a linear classifier on the frozen features of each backbone with the SD1.4 training set of GenImage (about 320k images), and fine-tune the backbone with LoRA under the same settings (\cref{tab:samebackbone}).
Both the linear classifier and LoRA fine-tuning have a lower mAUC than \saysee, and the 95\% confidence intervals of the differences all exclude zero; LoRA fine-tuning and the linear classifier differ by under 0.5 points on both backbones.

\subsection{Robustness analysis}
\label{sec:robust}

A detector that relies on format obtains inflated scores on existing evaluation sets (\cref{sec:intro}).
To test directly whether a detector takes format as evidence of authenticity, we take 150 real and 150 generated images from each evaluation set and make four versions of every image: unprocessed, JPEG (quality 75), WebP (quality 80), and a simulated triple re-post on social platforms, 7,200 images in total.
Choosing one version for the real images and one for the generated images gives 16 pairings.
We report the mean mAUC over the 16 pairings, the \emph{worst pairing} (the lowest mAUC among the 16), and the \emph{format gap}: the 12 pairings with different versions on the two sides form 6 pairs in which the versions of real and generated images are swapped, and the format gap is the mean absolute difference in mAUC within these 6 pairs.

Once format is decoupled from authenticity, 8 of the 10 trained detectors have a format gap between 5.7 and 35.7 points, and the direction of the gap varies from detector to detector, which shows that they take the correspondence between format and authenticity in their training sets as evidence (\cref{tab:format}).
On the mean mAUC over the 16 pairings and on the worst pairing, \saysee is the best of all methods (97.2\% and 94.6\% with MoE-ViE, against at most 92.1\% and 91.1\%); its format gap is close to that of SimplicityPrevails, and only CuRe has a lower one.
The remaining gap arises because compression shifts the scores of generated images less than those of real ones, while \see measures offsets on real images only (supplementary material).

\subsection{Ablation study}
\label{sec:ablation}

\paragraph{The two modules.}
Paired descriptions raise the mAUC from 93.48\% to 97.99\% with MoE-ViE and from 95.72\% to 96.34\% with PE (\cref{tab:modules}); with MoE-ViE the largest gain is on the latest generators, from 89.52\% to 95.55\%.
The real-image correction cannot take the place of this step: when \see is added directly to the minimal pair, the mAUC improves on neither backbone.
\say changes the direction but does not make it invariant to format; adding \see raises the worst pairing from 90.00\% to 94.57\% with MoE-ViE and from 87.58\% to 93.56\% with PE, lowers the format gap, and raises the mAUC further to 98.15\% and 96.80\%.
Both modules also reduce the bias measured in \cref{sec:bias} (supplementary material): pairing brings paintings, illustrations and document pages within $0.51\sigma$ and 3D renders from $3.63\sigma$ to $1.42\sigma$, and \see lowers the RMS format offset from $0.85\sigma$ to $0.13\sigma$.

\paragraph{Design of \see.}
When one family of conditions is corrected at a time, the mAUC gain with MoE-ViE comes from the image-type family, while the format family mainly lowers the format gap; with PE, the gain comes mainly from the format family (supplementary material).
The supplement also reports the effects of the generator names, the number of reference images and the temperature, size-matched prompt controls, a held-out codec experiment, real-only threshold calibration, and comparisons with linear probing and LoRA fine-tuning under the same five-crop feature aggregation protocol.

\section{Conclusion}
\label{sec:conclusion}

We presented \saysee, a training-free detector of AI-generated images.
Instead of learning from generated images, it reads out the concept of ``AI-generated'' that vision--language models acquire during pretraining: factors that language can express are paired in the descriptions, and factors it cannot express are measured and corrected with real images.
On 24 evaluation sets, \saysee outperforms existing trained detectors, and it performs best in the evaluation that decouples format from authenticity.

\paragraph{Limitations.}
\saysee is bounded by what the vision foundation model learned in pretraining, and its results on the latest generators remain below those on the standard benchmarks.
\see measures offsets on real images only and cannot correct the additional effect of a condition on generated images; \eg, WebP shifts the scores of generated images less than those of real ones (\cref{sec:robust}).

Beyond routine image processing, jailbreak studies in generative vision--language models~\cite{zhou2025jpro,zhou2026weakood} motivate testing how image authenticity detectors respond to deliberate input manipulation.

\section*{LLM Usage Statement}

We used Claude (Anthropic) via Claude Code to copyedit the text and translate some paragraphs into English.
Claude also helped write code for evaluation and for reproducing the compared methods.
The authors formulated the research question and designed the method and experiments.
They wrote the core implementation and the main text and take responsibility for the entire paper.

{
    \small
    \bibliographystyle{ieeenat_fullname}
    \bibliography{main}

\begin{thebibliography}{57}
\providecommand{\natexlab}[1]{#1}
\providecommand{\url}[1]{\texttt{#1}}
\expandafter\ifx\csname urlstyle\endcsname\relax
  \providecommand{\doi}[1]{doi: #1}\else
  \providecommand{\doi}{doi: \begingroup \urlstyle{rm}\Url}\fi

\bibitem[Bammey(2024)]{bammey2024synthbuster}
Quentin Bammey.
\newblock Synthbuster: Towards detection of diffusion model generated images.
\newblock \emph{IEEE Open Journal of Signal Processing}, 5:\penalty0 1--9,
  2024.

\bibitem[Betker et~al.(2023)Betker, Goh, Jing, Brooks, Wang, Li, Ouyang,
  Zhuang, Lee, Guo, Manassra, Dhariwal, Chu, Jiao, and
  Ramesh]{betker2023dalle3}
James Betker, Gabriel Goh, Li Jing, Tim Brooks, Jianfeng Wang, Linjie Li, Long
  Ouyang, Juntang Zhuang, Joyce Lee, Yufei Guo, Wesam Manassra, Prafulla
  Dhariwal, Casey Chu, Yunxin Jiao, and Aditya Ramesh.
\newblock Improving image generation with better captions.
\newblock \url{https://cdn.openai.com/papers/dall-e-3.pdf}, 2023.

\bibitem[Betser et~al.(2026)Betser, Hofman, Vainshtein, and Gilboa]{clide2026}
Roy Betser, Omer Hofman, Roman Vainshtein, and Guy Gilboa.
\newblock General and domain-specific zero-shot detection of generated images
  via conditional likelihood.
\newblock In \emph{Proceedings of the IEEE/CVF Winter Conference on
  Applications of Computer Vision (WACV)}, pages 7809--7820, 2026.

\bibitem[Bolya et~al.(2025)Bolya, Huang, Sun, Cho, Madotto, Wei, Ma, Zhi,
  Rajasegaran, Bangalath, Wang, Monteiro, Xu, Dong, Ravi, Li, Doll{\'a}r, and
  Feichtenhofer]{bolya2025pe}
Daniel Bolya, Po-Yao Huang, Peize Sun, Jang~Hyun Cho, Andrea Madotto, Chen Wei,
  Tengyu Ma, Jiale Zhi, Jathushan Rajasegaran, Hanoona Bangalath, Junke Wang,
  Marco Monteiro, Hu Xu, Shiyu Dong, Nikhila Ravi, Shang-Wen Li, Piotr
  Doll{\'a}r, and Christoph Feichtenhofer.
\newblock Perception encoder: The best visual embeddings are not at the output
  of the network.
\newblock In \emph{Advances in Neural Information Processing Systems
  (NeurIPS)}, pages 67743--67796, 2025.

\bibitem[Cao et~al.(2026)Cao, Tu, Xiao, Deng, Chen, Lin, and Wei]{probe2026}
Zijie Cao, Weijie Tu, Yao Xiao, Weijian Deng, Weiyan Chen, Liang Lin, and
  Pengxu Wei.
\newblock Where detectors fail: Probing generative space for generalizable
  {AI}-generated image detection.
\newblock In \emph{Proceedings of the 43rd International Conference on Machine
  Learning (ICML)}, pages 11477--11495. PMLR, 2026.

\bibitem[Cavia et~al.(2024)Cavia, Horwitz, Reiss, and Hoshen]{cavia2024wildrf}
Bar Cavia, Eliahu Horwitz, Tal Reiss, and Yedid Hoshen.
\newblock Real-time deepfake detection in the real-world.
\newblock \emph{arXiv preprint arXiv:2406.09398}, 2024.

\bibitem[Chen et~al.(2024)Chen, Zeng, Yang, and Yang]{chen2024drct}
Baoying Chen, Jishen Zeng, Jianquan Yang, and Rui Yang.
\newblock {DRCT}: Diffusion reconstruction contrastive training towards
  universal detection of diffusion generated images.
\newblock In \emph{Proceedings of the 41st International Conference on Machine
  Learning (ICML)}, pages 7621--7639. PMLR, 2024.

\bibitem[Chen et~al.(2025)Chen, Xi, Yan, Zhang, Wu, Xie, Chen, Xu, Guan, Yao,
  and Ding]{chen2025dda}
Ruoxin Chen, Junwei Xi, Zhiyuan Yan, Ke-Yue Zhang, Shuang Wu, Jingyi Xie, Xu
  Chen, Lei Xu, Isabel Guan, Taiping Yao, and Shouhong Ding.
\newblock Dual data alignment makes {AI}-generated image detector easier
  generalizable.
\newblock In \emph{Advances in Neural Information Processing Systems
  (NeurIPS)}, pages 117894--117919, 2025.

\bibitem[Chen et~al.(2026)Chen, Gao, Lin, Zhang, Zhao, Guan, Yao, and
  Ding]{aiginow}
Ruoxin Chen, Jiahui Gao, Kaiqing Lin, Keyue Zhang, Yandan Zhao, Isabel Guan,
  Taiping Yao, and Shouhong Ding.
\newblock Align, don't blend: Orthogonal supervision disentanglement as the key
  to generalizable {AI}-generated image detection.
\newblock In \emph{Proceedings of the 34th ACM International Conference on
  Multimedia (MM)}, pages 640--649. ACM, 2026.

\bibitem[Cheng et~al.(2025)Cheng, Lyu, Wang, Zhang, and Sehwag]{cheng2025cospy}
Siyuan Cheng, Lingjuan Lyu, Zhenting Wang, Xiangyu Zhang, and Vikash Sehwag.
\newblock {CO-SPY}: Combining semantic and pixel features to detect synthetic
  images by {AI}.
\newblock In \emph{Proceedings of the IEEE/CVF Conference on Computer Vision
  and Pattern Recognition (CVPR)}, pages 13455--13465, 2025.

\bibitem[Corvi et~al.(2023)Corvi, Cozzolino, Zingarini, Poggi, Nagano, and
  Verdoliva]{corvi2023detection}
Riccardo Corvi, Davide Cozzolino, Giada Zingarini, Giovanni Poggi, Koki Nagano,
  and Luisa Verdoliva.
\newblock On the detection of synthetic images generated by diffusion models.
\newblock In \emph{Proceedings of the IEEE International Conference on
  Acoustics, Speech and Signal Processing (ICASSP)}, pages 1--5, 2023.

\bibitem[Cozzolino et~al.(2024)Cozzolino, Poggi, Corvi, Nie{\ss}ner, and
  Verdoliva]{cozzolino2024raising}
Davide Cozzolino, Giovanni Poggi, Riccardo Corvi, Matthias Nie{\ss}ner, and
  Luisa Verdoliva.
\newblock Raising the bar of {AI}-generated image detection with {CLIP}.
\newblock In \emph{Proceedings of the IEEE/CVF Conference on Computer Vision
  and Pattern Recognition (CVPR) Workshops}, pages 4356--4366, 2024.

\bibitem[Cui et~al.(2026)Cui, Liu, Yu, Tan, Wei, Wang, Qin, Zhu, Wang, Lan, and
  Wu]{cure2026}
Manni Cui, Ruiqi Liu, Zijian Yu, Hao Tan, Zibo Wei, Zian Wang, Ziheng Qin,
  Huijia Zhu, Weiqiang Wang, Jun Lan, and Shu Wu.
\newblock Learning continuous source responses for generalizable {AI}-generated
  image detection.
\newblock \emph{arXiv preprint arXiv:2609.14316}, 2026.

\bibitem[Epstein et~al.(2023)Epstein, Hertzmann, {the Investigators of Human
  Creativity}, Akten, Farid, Fjeld, Frank, Groh, Herman, Leach, Mahari,
  Pentland, Russakovsky, Schroeder, and Smith]{epstein2023art}
Ziv Epstein, Aaron Hertzmann, {the Investigators of Human Creativity}, Memo
  Akten, Hany Farid, Jessica Fjeld, Morgan~R. Frank, Matthew Groh, Laura
  Herman, Neil Leach, Robert Mahari, Alex~{``Sandy''} Pentland, Olga
  Russakovsky, Hope Schroeder, and Amy Smith.
\newblock Art and the science of generative {AI}.
\newblock \emph{Science}, 380\penalty0 (6650):\penalty0 1110--1111, 2023.

\bibitem[Esser et~al.(2024)Esser, Kulal, Blattmann, Entezari, M{\"u}ller,
  Saini, Levi, Lorenz, Sauer, Boesel, Podell, Dockhorn, English, and
  Rombach]{esser2024sd3}
Patrick Esser, Sumith Kulal, Andreas Blattmann, Rahim Entezari, Jonas
  M{\"u}ller, Harry Saini, Yam Levi, Dominik Lorenz, Axel Sauer, Frederic
  Boesel, Dustin Podell, Tim Dockhorn, Zion English, and Robin Rombach.
\newblock Scaling rectified flow transformers for high-resolution image
  synthesis.
\newblock In \emph{Proceedings of the 41st International Conference on Machine
  Learning (ICML)}, pages 12606--12633. PMLR, 2024.

\bibitem[Frank et~al.(2020)Frank, Eisenhofer, Sch{\"o}nherr, Fischer, Kolossa,
  and Holz]{frank2020leveraging}
Joel Frank, Thorsten Eisenhofer, Lea Sch{\"o}nherr, Asja Fischer, Dorothea
  Kolossa, and Thorsten Holz.
\newblock Leveraging frequency analysis for deep fake image recognition.
\newblock In \emph{Proceedings of the 37th International Conference on Machine
  Learning (ICML)}, pages 3247--3258. PMLR, 2020.

\bibitem[Grommelt et~al.(2024)Grommelt, Weiss, Pfreundt, and
  Keuper]{grommelt2024fake}
Patrick Grommelt, Louis Weiss, Franz-Josef Pfreundt, and Janis Keuper.
\newblock Fake or {JPEG}? {R}evealing common biases in generated image
  detection datasets.
\newblock In \emph{Proceedings of the European Conference on Computer Vision
  (ECCV) Workshops}, pages 80--95. Springer, 2024.

\bibitem[Guillaro et~al.(2025)Guillaro, Zingarini, Usman, Sud, Cozzolino, and
  Verdoliva]{guillaro2025bfree}
Fabrizio Guillaro, Giada Zingarini, Ben Usman, Avneesh Sud, Davide Cozzolino,
  and Luisa Verdoliva.
\newblock A bias-free training paradigm for more general {AI}-generated image
  detection.
\newblock In \emph{Proceedings of the IEEE/CVF Conference on Computer Vision
  and Pattern Recognition (CVPR)}, pages 18685--18694, 2025.

\bibitem[Guo et~al.(2026)Guo, Ye, Zhang, Kang, Fu, He, and Li]{omniaid2026}
Yuncheng Guo, Junyan Ye, Chenjue Zhang, Hengrui Kang, Haohuan Fu, Conghui He,
  and Weijia Li.
\newblock {OmniAID}: Decoupling semantics and artifacts for universal
  {AI}-generated image detection in the wild.
\newblock In \emph{Proceedings of the 43rd International Conference on Machine
  Learning (ICML)}, pages 38614--38641. PMLR, 2026.

\bibitem[He et~al.(2024)He, Chen, and Ho]{he2024rigid}
Zhiyuan He, Pin-Yu Chen, and Tsung-Yi Ho.
\newblock {RIGID}: A training-free and model-agnostic framework for robust
  {AI}-generated image detection.
\newblock \emph{arXiv preprint arXiv:2405.20112}, 2024.

\bibitem[Huang et~al.(2025)Huang, Li, Yang, Peng, Huang, Wu, Tao, Yang, and
  Cheng]{sofake2025}
Zhenglin Huang, Xiangtai Li, Xi Yang, Bei Peng, Xiaowei Huang, Baoyuan Wu,
  Dacheng Tao, Ming-Hsuan Yang, and Guangliang Cheng.
\newblock {So-Fake}: Benchmarking and explaining social media image forgery
  detection.
\newblock \emph{arXiv preprint arXiv:2505.18660}, 2025.

\bibitem[Jiang et~al.(2026)Jiang, Tang, Gao, Yang, Cao, Shen, Zhang, Li, and
  Zhang]{dailybench}
Xin Jiang, Hao Tang, Junyao Gao, Zijie Yang, Meiqi Cao, Fei Shen, Dongming
  Zhang, Zechao Li, and Yongdong Zhang.
\newblock {DailyBench}: A unified benchmark for {AI}-generated and manipulated
  images from modern generative models.
\newblock \emph{arXiv preprint arXiv:2607.24016}, 2026.

\bibitem[Kim et~al.(2026)Kim, Choi, Seong, Kim, Lee, Yi, and Choi]{dear2026}
Dahye Kim, Jaehyun Choi, Hyun~Seok Seong, Seongho Kim, Donghun Lee, Sungwon Yi,
  and Jang-Ho Choi.
\newblock Dissect and prune: Enhancing robustness in {AI}-generated image
  detection.
\newblock In \emph{Proceedings of the 43rd International Conference on Machine
  Learning (ICML)}, pages 57385--57410. PMLR, 2026.

\bibitem[Ledoit and Wolf(2004)]{ledoit2004well}
Olivier Ledoit and Michael Wolf.
\newblock A well-conditioned estimator for large-dimensional covariance
  matrices.
\newblock \emph{Journal of Multivariate Analysis}, 88\penalty0 (2):\penalty0
  365--411, 2004.

\bibitem[Li et~al.(2025{\natexlab{a}})Li, Wang, Li, Miao, Sun, Zhang, Ji, and
  Zhu]{rrdataset2025}
Chunxiao Li, Xiaoxiao Wang, Meiling Li, Boming Miao, Peng Sun, Yunjian Zhang,
  Xiangyang Ji, and Yao Zhu.
\newblock Bridging the gap between ideal and real-world evaluation:
  Benchmarking {AI}-generated image detection in challenging scenarios.
\newblock In \emph{Proceedings of the IEEE/CVF International Conference on
  Computer Vision (ICCV)}, pages 20379--20389, 2025{\natexlab{a}}.

\bibitem[Li et~al.(2025{\natexlab{b}})Li, Yan, He, Zeng, Jiang, Xiong, and
  Fu]{aigibench2025}
Ziqiang Li, Jiazhen Yan, Ziwen He, Kai Zeng, Weiwei Jiang, Lizhi Xiong, and
  Zhangjie Fu.
\newblock Is artificial intelligence generated image detection a solved
  problem?
\newblock In \emph{Advances in Neural Information Processing Systems (NeurIPS),
  Datasets and Benchmarks Track}, pages 191125--191159, 2025{\natexlab{b}}.

\bibitem[Lin et~al.(2024)Lin, Gupta, Zhang, Ren, Liu, Ding, Wang, Li,
  Verdoliva, and Hu]{lin2024detecting}
Li Lin, Neeraj Gupta, Yue Zhang, Hainan Ren, Chun-Hao Liu, Feng Ding, Xin Wang,
  Xin Li, Luisa Verdoliva, and Shu Hu.
\newblock Detecting multimedia generated by large {AI} models: A survey.
\newblock \emph{arXiv preprint arXiv:2402.00045}, 2024.

\bibitem[Liu et~al.(2026)Liu, Han, Zhang, Yao, Yan, Shen, Chen, Cui, Sun, Weng,
  Dong, Wang, and Wu]{realchain}
Ruiqi Liu, Yi Han, Zhengbo Zhang, Liwei Yao, Zhiyuan Yan, Jialiang Shen, ZhiJin
  Chen, Manni Cui, Boyi Sun, Lubin Weng, Jing Dong, Yan Wang, and Shu Wu.
\newblock Beyond artifacts: Real-centric envelope modeling for reliable
  {AI}-generated image detection.
\newblock In \emph{Proceedings of the European Conference on Computer Vision
  (ECCV)}, pages 225--243. Springer, 2026.

\bibitem[Livernoche et~al.(2025)Livernoche, Arodi, Musulan, Yang, Salvail,
  Marceau~Caron, Godbout, and Rabbany]{openfake2025}
Victor Livernoche, Akshatha Arodi, Andreea Musulan, Zachary Yang, Adam Salvail,
  Ga{\'e}tan Marceau~Caron, Jean-Fran{\c{c}}ois Godbout, and Reihaneh Rabbany.
\newblock {OpenFake}: An open dataset and platform toward real-world deepfake
  detection.
\newblock \emph{arXiv preprint arXiv:2509.09495}, 2025.

\bibitem[Marra et~al.(2019)Marra, Gragnaniello, Verdoliva, and
  Poggi]{marra2019gans}
Francesco Marra, Diego Gragnaniello, Luisa Verdoliva, and Giovanni Poggi.
\newblock Do {GANs} leave artificial fingerprints?
\newblock In \emph{Proceedings of the IEEE Conference on Multimedia Information
  Processing and Retrieval (MIPR)}, pages 506--511, 2019.

\bibitem[Mu et~al.(2026)Mu, Hu, Zou, Bai, and Hu]{hires50k}
Lianrui Mu, Haoji Hu, Xingze Zou, Jianhong Bai, and Jiaqi Hu.
\newblock No pixel left behind: A detail-preserving architecture for robust
  high-resolution {AI}-generated image detection.
\newblock In \emph{International Conference on Learning Representations
  (ICLR)}, 2026.

\bibitem[Ojha et~al.(2023)Ojha, Li, and Lee]{ojha2023univfd}
Utkarsh Ojha, Yuheng Li, and Yong~Jae Lee.
\newblock Towards universal fake image detectors that generalize across
  generative models.
\newblock In \emph{Proceedings of the IEEE/CVF Conference on Computer Vision
  and Pattern Recognition (CVPR)}, pages 24480--24489, 2023.

\bibitem[Pantsios et~al.(2026)Pantsios, Karageorgiou, Koutlis, Karantaidis,
  Papadopoulou, and Papadopoulos]{aigenimages2026}
Thanasis Pantsios, Dimitrios Karageorgiou, Christos Koutlis, George
  Karantaidis, Olga Papadopoulou, and Symeon Papadopoulos.
\newblock Automated in-the-wild data collection for continual {AI} generated
  image detection.
\newblock In \emph{Proceedings of the 5th ACM International Workshop on
  Multimedia AI against Disinformation (MAD)}, pages 39--48. ACM, 2026.

\bibitem[Park and Owens(2025)]{park2025cf}
Jeongsoo Park and Andrew Owens.
\newblock Community forensics: Using thousands of generators to train fake
  image detectors.
\newblock In \emph{Proceedings of the IEEE/CVF Conference on Computer Vision
  and Pattern Recognition (CVPR)}, pages 8245--8257, 2025.

\bibitem[Qin et~al.(2026)Qin, Ji, Tao, Tian, Liu, Wang, and Zheng]{gapl2026}
Ziheng Qin, Yuheng Ji, Renshuai Tao, Yuxuan Tian, Yuyang Liu, Yipu Wang, and
  Xiaolong Zheng.
\newblock Scaling up {AI}-generated image detection with generator-aware
  prototypes.
\newblock In \emph{Proceedings of the IEEE/CVF Conference on Computer Vision
  and Pattern Recognition (CVPR)}, pages 43008--43017, 2026.

\bibitem[Radford et~al.(2021)Radford, Kim, Hallacy, Ramesh, Goh, Agarwal,
  Sastry, Askell, Mishkin, Clark, Krueger, and Sutskever]{radford2021clip}
Alec Radford, Jong~Wook Kim, Chris Hallacy, Aditya Ramesh, Gabriel Goh,
  Sandhini Agarwal, Girish Sastry, Amanda Askell, Pamela Mishkin, Jack Clark,
  Gretchen Krueger, and Ilya Sutskever.
\newblock Learning transferable visual models from natural language
  supervision.
\newblock In \emph{Proceedings of the 38th International Conference on Machine
  Learning (ICML)}, pages 8748--8763. PMLR, 2021.

\bibitem[Rajan et~al.(2025)Rajan, Ojha, Schloesser, and Lee]{rajan2025aligned}
Anirudh~Sundara Rajan, Utkarsh Ojha, Jedidiah Schloesser, and Yong~Jae Lee.
\newblock Aligned datasets improve detection of latent diffusion-generated
  images.
\newblock In \emph{International Conference on Learning Representations
  (ICLR)}, 2025.

\bibitem[Ricker et~al.(2024{\natexlab{a}})Ricker, Damm, Holz, and
  Fischer]{ricker2024towards}
Jonas Ricker, Simon Damm, Thorsten Holz, and Asja Fischer.
\newblock Towards the detection of diffusion model deepfakes.
\newblock In \emph{Proceedings of the 19th International Joint Conference on
  Computer Vision, Imaging and Computer Graphics Theory and Applications
  (VISAPP)}, pages 446--457. SCITEPRESS, 2024{\natexlab{a}}.

\bibitem[Ricker et~al.(2024{\natexlab{b}})Ricker, Lukovnikov, and
  Fischer]{ricker2024aeroblade}
Jonas Ricker, Denis Lukovnikov, and Asja Fischer.
\newblock {AEROBLADE}: Training-free detection of latent diffusion images using
  autoencoder reconstruction error.
\newblock In \emph{Proceedings of the IEEE/CVF Conference on Computer Vision
  and Pattern Recognition (CVPR)}, pages 9130--9140, 2024{\natexlab{b}}.

\bibitem[Rombach et~al.(2022)Rombach, Blattmann, Lorenz, Esser, and
  Ommer]{rombach2022ldm}
Robin Rombach, Andreas Blattmann, Dominik Lorenz, Patrick Esser, and Bj{\"o}rn
  Ommer.
\newblock High-resolution image synthesis with latent diffusion models.
\newblock In \emph{Proceedings of the IEEE/CVF Conference on Computer Vision
  and Pattern Recognition (CVPR)}, pages 10684--10695, 2022.

\bibitem[Stojni{\'c} et~al.(2026)Stojni{\'c}, Ramos, Kordopatis-Zilos, Garcia,
  and Tolias]{invisibleshortcuts2026}
Vladan Stojni{\'c}, Ryan Ramos, Giorgos Kordopatis-Zilos, Noa Garcia, and
  Giorgos Tolias.
\newblock Invisible shortcuts: Why vision encoders know your camera.
\newblock In \emph{Proceedings of the European Conference on Computer Vision
  (ECCV)}, pages 112--131. Springer, 2026.

\bibitem[Tan et~al.(2024)Tan, Zhao, Wei, Gu, Liu, and Wei]{tan2024npr}
Chuangchuang Tan, Yao Zhao, Shikui Wei, Guanghua Gu, Ping Liu, and Yunchao Wei.
\newblock Rethinking the up-sampling operations in {CNN}-based generative
  network for generalizable deepfake detection.
\newblock In \emph{Proceedings of the IEEE/CVF Conference on Computer Vision
  and Pattern Recognition (CVPR)}, pages 28130--28139, 2024.

\bibitem[Wang et~al.(2020)Wang, Wang, Zhang, Owens, and Efros]{wang2020cnnspot}
Sheng-Yu Wang, Oliver Wang, Richard Zhang, Andrew Owens, and Alexei~A. Efros.
\newblock {CNN}-generated images are surprisingly easy to spot... for now.
\newblock In \emph{Proceedings of the IEEE/CVF Conference on Computer Vision
  and Pattern Recognition (CVPR)}, pages 8692--8701, 2020.

\bibitem[Wang et~al.(2023)Wang, Bao, Zhou, Wang, Hu, Chen, and
  Li]{wang2023dire}
Zhendong Wang, Jianmin Bao, Wengang Zhou, Weilun Wang, Hezhen Hu, Hong Chen,
  and Houqiang Li.
\newblock {DIRE} for diffusion-generated image detection.
\newblock In \emph{Proceedings of the IEEE/CVF International Conference on
  Computer Vision (ICCV)}, pages 22445--22455, 2023.

\bibitem[Willi et~al.(2026)Willi, Mathys, and Graber]{willi2026synthclic}
Marco Willi, Melanie Mathys, and Michael Graber.
\newblock Synthetic image detection with {CLIP}: Understanding and assessing
  predictive cues.
\newblock \emph{Array}, 32:\penalty0 101222, 2026.

\bibitem[Yan et~al.(2025)Yan, Li, Cai, Hao, Jiang, Hu, and Xie]{yan2025aide}
Shilin Yan, Ouxiang Li, Jiayin Cai, Yanbin Hao, Xiaolong Jiang, Yao Hu, and
  Weidi Xie.
\newblock A sanity check for {AI}-generated image detection.
\newblock In \emph{International Conference on Learning Representations
  (ICLR)}, 2025.

\bibitem[Yu et~al.(2019)Yu, Davis, and Fritz]{yu2019attributing}
Ning Yu, Larry~S. Davis, and Mario Fritz.
\newblock Attributing fake images to {GANs}: Learning and analyzing {GAN}
  fingerprints.
\newblock In \emph{Proceedings of the IEEE/CVF International Conference on
  Computer Vision (ICCV)}, pages 7555--7565, 2019.

\bibitem[Zhang et~al.(2026{\natexlab{a}})Zhang, Dong, Tran, Gschwind, Yang,
  Chen, Ahmadyan, Moon, Zhang, Kirmani, Damavandi, and Kumar]{moevie2026}
Bonan Zhang, Shiyu Dong, Quan~Hung Tran, Katharina Gschwind, Shuqi Yang, Sijia
  Chen, Adel Ahmadyan, Seungwhan Moon, Lu Zhang, Ahmed Kirmani, Babak
  Damavandi, and Anuj Kumar.
\newblock {MoE-ViE}: Mixture of experts vision encoder for efficient image and
  video understanding.
\newblock In \emph{Proceedings of the European Conference on Computer Vision
  (ECCV)}, pages 297--316. Springer, 2026{\natexlab{a}}.

\bibitem[Zhang et~al.(2026{\natexlab{b}})Zhang, Yu, Guan, Tian, Zhang, and
  Zhu]{mllmgenset}
Zirui Zhang, Yinbo Yu, Donghai Guan, Chunwei Tian, Daoqiang Zhang, and Qi Zhu.
\newblock A benchmark dataset for {MLLM}-generated image detection: {GPT}
  {Image2} \& {Nano} {Banana2}.
\newblock \emph{arXiv preprint arXiv:2608.01258}, 2026{\natexlab{b}}.

\bibitem[Zheng et~al.(2024)Zheng, Lin, Zhao, Wang, Guo, Liu, and
  Shen]{zheng2024breaking}
Chende Zheng, Chenhao Lin, Zhengyu Zhao, Hang Wang, Xu Guo, Shuai Liu, and Chao
  Shen.
\newblock Breaking semantic artifacts for generalized {AI}-generated image
  detection.
\newblock In \emph{Advances in Neural Information Processing Systems
  (NeurIPS)}, pages 59570--59596, 2024.

\bibitem[Zhong et~al.(2023)Zhong, Xu, Li, Qian, and Zhang]{zhong2023patchcraft}
Nan Zhong, Yiran Xu, Sheng Li, Zhenxing Qian, and Xinpeng Zhang.
\newblock {PatchCraft}: Exploring texture patch for efficient {AI}-generated
  image detection.
\newblock \emph{arXiv preprint arXiv:2311.12397}, 2023.

\bibitem[Zhou et~al.(2025{\natexlab{a}})Zhou, Bai, Gao, Dai, and
  Xia]{zhou2025jpro}
Yuxuan Zhou, Yang Bai, Kuofeng Gao, Tao Dai, and Shu-Tao Xia.
\newblock {JPRO}: Automated multimodal jailbreaking via multi-agent
  collaboration framework.
\newblock \emph{arXiv preprint arXiv:2511.07315}, 2025{\natexlab{a}}.

\bibitem[Zhou et~al.(2025{\natexlab{b}})Zhou, He, Lin, Fan, Ding, and
  Li]{omat2025}
Yue Zhou, Xinan He, Kaiqing Lin, Bing Fan, Feng Ding, and Bin Li.
\newblock Breaking latent prior bias in detectors for generalizable {AIGC}
  image detection.
\newblock In \emph{Advances in Neural Information Processing Systems
  (NeurIPS)}, pages 34606--34636, 2025{\natexlab{b}}.

\bibitem[Zhou et~al.(2026{\natexlab{a}})Zhou, He, Lin, Fan, Ding, and
  Li]{simplicityprevails2026}
Yue Zhou, Xinan He, Kaiqing Lin, Bing Fan, Feng Ding, and Bin Li.
\newblock Simplicity prevails: The emergence of generalizable {AIGI} detection
  in visual foundation models.
\newblock In \emph{Proceedings of the 34th ACM International Conference on
  Multimedia (MM)}, pages 16095--16103. ACM, 2026{\natexlab{a}}.

\bibitem[Zhou et~al.(2026{\natexlab{b}})Zhou, Peng, Bai, Gao, Zhang, Zhang,
  Chen, Yu, Dai, and Xia]{zhou2026weakood}
Yuxuan Zhou, Yuzhao Peng, Yang Bai, Kuofeng Gao, Yihao Zhang, Yechao Zhang, Xun
  Chen, Tao Yu, Tao Dai, and Shu-Tao Xia.
\newblock Why does {Weak-OOD} help? a further step towards understanding
  jailbreaking {VLMs}.
\newblock In \emph{Proceedings of the 2026 Conference on Empirical Methods in
  Natural Language Processing}, pages 45038--45067, 2026{\natexlab{b}}.

\bibitem[Zhou et~al.(2026{\natexlab{c}})Zhou, Yu, Huang, Zhang, Dai, and
  Xia]{zhou2026crda}
Yuxuan Zhou, Tao Yu, Wen Huang, Yuheng Zhang, Tao Dai, and Shu-Tao Xia.
\newblock Improving deepfake detection with reinforcement learning-based
  adaptive data augmentation.
\newblock \emph{Proceedings of the AAAI Conference on Artificial Intelligence},
  40\penalty0 (5):\penalty0 3381--3389, 2026{\natexlab{c}}.

\bibitem[Zhu et~al.(2023)Zhu, Chen, Yan, Huang, Lin, Li, Tu, Hu, Hu, and
  Wang]{zhu2023genimage}
Mingjian Zhu, Hanting Chen, Qiangyu Yan, Xudong Huang, Guanyu Lin, Wei Li,
  Zhijun Tu, Hailin Hu, Jie Hu, and Yunhe Wang.
\newblock {GenImage}: A million-scale benchmark for detecting {AI}-generated
  image.
\newblock In \emph{Advances in Neural Information Processing Systems (NeurIPS),
  Datasets and Benchmarks Track}, pages 77771--77782, 2023.

\end{thebibliography}
}
\clearpage
\appendix
\setcounter{table}{0}
\setcounter{figure}{0}
\setcounter{equation}{0}
\renewcommand{\thetable}{S\arabic{table}}
\renewcommand{\thefigure}{S\arabic{figure}}
\renewcommand{\theequation}{S\arabic{equation}}
\renewcommand{\theHtable}{supp.\arabic{table}}
\renewcommand{\theHfigure}{supp.\arabic{figure}}
\renewcommand{\theHequation}{supp.\arabic{equation}}

\hyphenation{ImageNet GenImage ArtBench SaySee AEROBLADE}
\setcounter{topnumber}{4}
\setcounter{dbltopnumber}{3}
\setcounter{totalnumber}{6}
\renewcommand{\topfraction}{0.95}
\renewcommand{\dbltopfraction}{0.95}
\renewcommand{\textfraction}{0.05}
\renewcommand{\floatpagefraction}{0.85}
\renewcommand{\dblfloatpagefraction}{0.85}
\raggedbottom

\twocolumn[\centering{\Large\bfseries Appendix}\par\vspace{1em}]
\pdfbookmark[0]{Appendix}{appendix}
\noindent Two groups of results answer the question in the title most directly.
The first is the fair same-backbone comparison (\cref{sec:supp-probe}, \cref{tab:s13,tab:s15}): on the same backbone and the same image embedding, a linear probe trained on 320 thousand images, even with its generated training images compressed with the JPEG settings of real images, has a lower mAUC over the 24 evaluation sets and a lower mean mAUC over the 16 pairings with format decoupled from authenticity than \saysee with both backbones, and a larger format gap; with MoE-ViE, the two mAUCs are 96.95\% \vs 98.15\% and 95.23\% \vs 97.19\%.
The second is a threshold set from real images only (\cref{sec:supp-threshold}, \cref{tab:s10,tab:s11}), with which the whole detector, from the generation direction and the correction to the threshold, uses no generated image: with MoE-ViE and the reference set that includes social-platform photographs, the bAcc over the 24 evaluation sets is 93.6\%, compared with 93.7\% for the fixed threshold of the main paper.

\noindent This appendix provides the details and results that did not fit in the main paper.
\cref{sec:supp-templates} lists all descriptions of \say, and \cref{sec:supp-real} the real images used by \see.
\cref{sec:supp-setup,sec:supp-comparisons,sec:supp-probe,sec:supp-format} accompany Secs.~5.1 to 5.4 of the main paper: the experimental setup, the comparison with existing detectors (including results by generator release date and thresholds set from real images only), the comparison with same-backbone trained methods, and the evaluation with format decoupled from authenticity.
\cref{sec:supp-ablations,sec:supp-bias} accompany Sec.~5.5 with further ablations and with how bias is measured and removed, and \cref{sec:supp-npu} reports results on a Huawei Ascend NPU.

Tables and figures in this appendix carry the prefix S.
``Tab.~1--7'', ``Fig.~1--5'', ``Eq.~(1)--(8)'' and section numbers without a letter refer to the main paper.

\section{Description templates and generator names of \say}
\label{sec:supp-templates}

\paragraph{Combinations.}
\say uses six image types: photo, painting, illustration, 3D render, screenshot and document page.
For content, there are 11 themes, written as people, a person's face, an animal, plants, a landscape, a building, a vehicle, an indoor room, an object, food and the sky, plus the option ``unspecified''.
The noun phrase $N$ of a description is ``\{image type\}'' if the content is unspecified and ``\{image type\} of \{content\}'' otherwise, \eg, ``painting of an animal''.
The first four image types are each combined with all 12 content options, and screenshot and document page only with ``unspecified'', which gives 50 combinations.

\paragraph{Templates.}
\cref{tab:s01} lists all templates.
The generated side uses the same four templates for every image type, while the real side has its own two or three templates for each image type.

\begin{table*}[t]
\centering
\small
\caption{\textbf{Description templates of \say.} $N$ is the noun phrase; the article is ``a'' or ``an'' according to its first letter.}
\label{tab:s01}
\begin{tabular}{ll>{\raggedright\arraybackslash}p{0.67\textwidth}}
\toprule
Side & Image type & Templates \\
\midrule
Generated & all & an AI-generated N; a N generated by AI; a N made with a text-to-image model; a N created by artificial intelligence \\
Generated (names) & all & a \{image type\} generated by \{name\} \\
Real & photo & a real N; a N taken with a camera; a N taken by a photographer \\
Real & painting & a N made by a human artist; a N painted by a person \\
Real & illustration & a N drawn by a human artist; a N drawn by a person \\
Real & 3D render & a N made by a human artist; a N created by a 3D artist \\
Real & screenshot & a N captured on a computer; a N taken on a phone \\
Real & document page & a N scanned from paper; a N written and laid out by a person \\
\bottomrule
\end{tabular}

\end{table*}

\paragraph{Generator names.}
We fix 18 names in advance: Midjourney, Stable Diffusion, DALL-E, FLUX, GPT-4o, GPT-Image, Nano Banana, Imagen, Seedream, Ideogram, Adobe Firefly, Leonardo AI, Kolors, Hunyuan Image, Qwen-Image, Recraft, Playground and Krea.
A name sentence states only the image type, not the content, so all combinations of an image type share the same 18 sentences.

\paragraph{Aggregation.}
All text embeddings are normalized.
For combination $(m,c)$, $\bar g_{m,c}$ is the normalized mean of the embeddings of the four generated-side templates, $\bar n_m$ that of the 18 name sentences, and $\bar r_{m,c}$ that of the real-side templates.
The combination directions and the generation direction then follow Eqs.~(4) and~(5).
The minimal pair consists of ``an AI-generated image'' and ``a real image''.

\section{Real images used by \see}
\label{sec:supp-real}

Each of the three families of conditions of \see estimates its prototypes and offsets from a set of real images.
All of these images come from public datasets, and both backbones use the same images.

The images of each family are split into an estimation part, which gives the prototypes and offsets, and a held-out part, which takes no part in the estimation (\cref{tab:s02}).
Below we describe how each family uses the two parts; \cref{tab:s03} lists the sources of each part.

\begin{table}[t]
\centering
\small
\caption{\textbf{Numbers of real images used by the three families of \see.} The images under ``Estimation'' give the prototypes and offsets; those under ``Held-out'' take no part in the estimation. The 4,862 photos of the image-type family are the JPEG photographs among the 9,724 of the content family, and the 2,510 held-out photographs of the two families are the same.}
\label{tab:s02}
\fittab{
\begin{tabular}{llrr}
\toprule
Family & Conditions & Estimation & Held-out \\
\midrule
Format & 133 format conditions & 211 & 80 \\
\midrule
Image type & Photos & 4,862 & 2,510 \\
 & Paintings & 800 & 756 \\
 & Illustrations & 168 & 192 \\
 & 3D renders & 327 & 340 \\
 & Screenshots & 295 & 297 \\
 & Document pages & 321 & 317 \\
\midrule
Content & 11 content themes & 9,724 & 2,510 \\
\bottomrule
\end{tabular}
}
\end{table}

\paragraph{Format.}
We use 291 photographs, each converted to 133 conditions: 7 sizes (the whole image resized to a short side of 448, 512, 614, 768, 1024 or 1536 pixels, plus local crops at native resolution) times 19 codecs (lossless; JPEG quality 50, 60, 70, 73, 75, 80, 85, 90, 92, 94, 95, 96, 98 and 100; WebP quality 50, 75, 80 and 90).
The reference condition is a short side of 1024 pixels saved losslessly.
Of these photographs, 211 form the estimation part.
To select the temperature, we estimate prototypes and offsets on 140 of them and compare the candidate values on the other 71; after the selection, prototypes and offsets are re-estimated on all 211.
The remaining 80 form the held-out part, which enters no estimation and is used only to measure format offsets in figures such as Fig.~3 and \cref{fig:s06}.

\paragraph{Photographs.}
The image-type and content families share one set of photographs, 9,724 in the estimation part and 2,510 in the held-out part.
Each part is half JPEG files and half losslessly saved files: 4,862 of each in the estimation part and 1,255 of each in the held-out part.

\paragraph{Image type.}
The photo prototype is estimated from the 4,862 JPEG photographs of the estimation part (Wikimedia Commons 2,624, HDR+ 1,812, FODB 426), and the held-out photos are the 2,510 photographs above.
Each of the other five image types has its own set of images, split at random into an estimation part and a held-out part.
The held-out images of all six image types are pooled and split at random into two halves, one for selecting the temperature and one for evaluation.

The offset of an image type is the mean score of its estimation part along the generation direction minus that of the photographs.
Both sides consist of JPEG files, so the offset contains no difference due to JPEG coding and reflects the image type alone.
For this reason, the photo prototype uses only JPEG photographs, and the 3D renders (Hypersim), screenshots (ScreenSpot) and document pages (DocLayNet), whose original files are lossless, are converted as whole images to JPEG at quality 90 before being read.
The paintings and illustrations (NGA, Met, ArtBench, Wikimedia Commons) are JPEG files with few exceptions.
The offset of JPEG relative to lossless saving is corrected separately by the format family.

\paragraph{Content.}
The 9,724 photographs of the estimation part fall into 11 content themes: buildings and streets 1,588, natural landscapes 1,389, plants 1,088, vehicles 1,034, indoor scenes 805, groups of people 783, sky and textures 731, portraits 617, animals 604, objects 590, and food 495.
The held-out part consists of the same 2,510 photographs, each labeled with its theme and split at random into two halves, one for selecting the temperature and one for evaluation.

Unless stated otherwise, the unit $\sigma$ in the main paper and in this document is the standard deviation of the scores of the 2,510 held-out photographs.

\begin{table*}[t]
\centering
\small
\caption{\textbf{Numbers of images from each source, and licenses.} The two ``Format'' columns and the two ``Photos'' columns correspond to the photographs of the format family and to the photographs shared by the image-type and content families in \cref{tab:s02}; the last row gives the column totals.}
\label{tab:s03}
\fittab[\textwidth]{
\begin{tabular}{lrrrr>{\raggedright\arraybackslash}p{0.22\textwidth}>{\raggedright\arraybackslash}p{0.25\textwidth}}
\toprule
Source & \shortstack{Format\\estimation} & \shortstack{Format\\held-out} & \shortstack{Photos\\estimation} & \shortstack{Photos\\held-out} & Non-photo image types & License \\
\midrule
HDR+ & 59 & 19 & 2,793 & 263 &  & CC BY-SA 4.0 \\
MIT-Adobe FiveK & 43 & 16 & 1,468 & 275 &  & research use only \\
Wikimedia Commons & 40 & 16 & 2,624 & 521 & paintings 46, illustrations 360, 3D renders 67, document pages 38 & per image; mostly CC BY-SA 4.0, also CC0, CC BY and public domain \\
Cube++ & 22 & 14 & 865 & 167 &  & CC BY 4.0 \\
PASCALRAW & 18 & 8 & 649 & 121 &  & CC BY-NC-SA 3.0 \\
LSDIR & 11 & 4 & 637 & 98 &  & research use only \\
DIV2K & 10 &  & 262 & 42 &  & research use only \\
FODB & 8 & 3 & 426 & 21 &  & research use only \\
VISION &  &  &  & 631 &  & CC BY-SA 4.0 \\
raw.pixls.us &  &  &  & 371 &  & CC0 \\
NGA &  &  &  &  & paintings 896 & CC0 (open access) \\
The Met &  &  &  &  & paintings 365 & CC0 (open access) \\
ArtBench-10 &  &  &  &  & paintings 249 & research use only \\
Hypersim &  &  &  &  & 3D renders 600 & CC BY-SA 3.0 \\
ScreenSpot &  &  &  &  & screenshots 592 & Apache-2.0 \\
DocLayNet &  &  &  &  & document pages 600 & CDLA-Permissive-1.0 \\
Total & 211 & 80 & 9,724 & 2,510 & 3,813 &  \\
\bottomrule
\end{tabular}
}
\end{table*}

From MIT-Adobe FiveK we use only the RAW files, not the retouched versions.
We inspected the ScreenSpot screenshots one by one and removed 12 that may contain AI-generated content.

\paragraph{No overlap with the evaluation sets.}
Where the source of the real images of an evaluation set is known, it is ImageNet, MSCOCO, RAISE-1k, CLIC 2020 or CC3M, none of which appears in \cref{tab:s03}; the other evaluation sets use real images collected by their authors or taken from other public datasets.
Using the image embeddings of both backbones, we compared every real image used by \see with all images of the 24 evaluation sets.
No pair has a cosine similarity above 0.98, and inspecting the pairs above 0.95 one by one, we found no identical images.

\section{Experimental setup}
\label{sec:supp-setup}

\subsection{Composition of the evaluation sets}

\cref{tab:s04} lists the 24 evaluation sets as we use them.
We made all design choices for \saysee{} using only validation data disjoint from these 24 evaluation sets.
These choices included how to word the templates and pair descriptions by image type and content in \say{}, and how to define the conditions and construct their prototypes in \see{}.

\paragraph{Use of the images.}
We use the released files as they are, with no resizing, cropping or re-encoding beforehand.
\saysee reads five $448\times448$ crops of an image at native resolution.
An image whose short side is below 448 pixels is first upscaled to a short side of 448 pixels (bicubic interpolation, aspect ratio kept).
The other methods use their own preprocessing.

\paragraph{Real images.}
Each evaluation set uses the real images specified by its authors.
GenImage++ provides no real images, so we draw them from the ImageNet validation set: we first sample images whose short side is at least 448 pixels and then add smaller images in the proportion they have in the validation set.
For CO-SPY-Wild we use the CC3M real images released by its authors.
The authors of Synthbuster and B-Free-NG both specify RAISE-1k as the real images, so the two sets have identical real images.
SocialRF and WildRF, Chameleon and CommunityAI, and AIGI-Now and DRCT-2M also share a small number of images; we keep them in each set as released.

\paragraph{MLLMGenSet.}
Half of the generated images of this set are the original files and half are versions preprocessed by the dataset authors.
We count both halves; results on the original files alone are given in \cref{sec:supp-mllm}.

\begin{table*}[t]
\centering
\footnotesize
\caption{\textbf{Composition of the 24 evaluation sets.}}
\label{tab:s04}
\begin{tabular}{ll>{\raggedright\arraybackslash}p{0.15\textwidth}>{\raggedright\arraybackslash}p{0.555\textwidth}}
\toprule
Group & Set & Source of real images & Generators or subsets \\
\midrule
Standard & GenImage & ImageNet & ADM, BigGAN, GLIDE, Midjourney, SD v1.4, SD v1.5, VQDM, Wukong \\
 & AIGCDB & original set & 17: ProGAN, StyleGAN, StyleGAN2, BigGAN, CycleGAN, StarGAN, GauGAN, whichfaceisreal, ADM, GLIDE, VQDM, DALL\textperiodcentered{}E 2, Midjourney, SD v1.4, SD v1.5, SDXL, Wukong \\
 & DRCT-2M & MSCOCO & 13: LDM, SD v1.4, v1.5, v2.1, SDXL and its Refiner, SD-Turbo, SDXL-Turbo, LCM-LoRA (2), ControlNet (3) \\
 & Synthbuster & RAISE-1k & DALL\textperiodcentered{}E 2, DALL\textperiodcentered{}E 3, Firefly, GLIDE, Midjourney v5, SD v1.3, SD v1.4, SD v2, SDXL \\
 & GenImage++ & ImageNet validation set & 11 subsets: FLUX.1 (6), SD 1.5, SDXL, SD3 (3) \\
 & B-Free-NG & RAISE-1k & FLUX, SD3 \\
 & SynthCLIC & CLIC 2020 & FLUX.1-dev, FLUX.1-schnell, SD3-medium, Imagen 3 \\
\midrule
In-the-wild & Chameleon & original set & no generator labels \\
 & WildRF & original set & by platform: Reddit, Twitter, Facebook \\
 & SocialRF & original set & no generator labels \\
 & CommunityAI & original set & no generator labels \\
 & B-Free-Online & original set & no generator labels \\
 & CO-SPY-Wild & CC3M & Lexica \\
 & OpenFake-Reddit & original set & no generator labels \\
 & RRDataset & original set & by scenario: original, recaptured, transmitted through platforms \\
 & HiRes-50K & original set & 8 resolution bins \\
\midrule
Latest & DailyBench & original set & SD 3.5, FLUX, FLUX.2, Qwen-Image, Z-Image, Nano Banana 2, GPT Image 2 \\
 & So-Fake-OOD & original set & 15: FLUX.2, FLUX.1 pro, GPT Image 1.5, GPT Image 2, GPT-4o, HiDream, Ideogram 2, Ideogram 3, Imagen 3, Imagen 4, Recraft v3, Seedream 3.0, Seedream 4.5, Nano Banana, Nano Banana 2 \\
 & OpenFake-Test & original set & 18, including GPT Image 2, Nano Banana Pro, Midjourney v7, FLUX.2 klein, Seedream 5.0, Sora 2, Z-Image Turbo, Recraft v3, Ideogram 2.0, ERNIE-Image \\
 & AIGenImages2026 & original set & 19, including FLUX.2 (3), Gemini 2.5 Flash Image, Gemini 3 Pro Image, GPT Image 1.5, Firefly Image 5, Midjourney v7, Imagen 4, Seedream 4.5, Z-Image Turbo \\
 & AIGI-Now & original set & 9: Nano Banana, FLUX dev, FLUX Krea, FLUX Kontext, FLUX pro, GPT-4o, Jimeng, Kling, MiniMax \\
 & MLLMGenSet & original set & GPT Image 2, Nano Banana 2; half original files, half preprocessed versions \\
 & RealChain & original set & FLUX.1, Hunyuan 3, Nano Banana, Qwen-Image, SD 3.5, Seedream 4 \\
 & Mirage-Test & original set & 6 generators or fine-tuned models, 5 content categories \\
\bottomrule
\end{tabular}

\end{table*}

\subsection{Reproduction of the compared methods}
\label{sec:supp-repro}

We run all compared methods with the code, weights and preprocessing released by their authors (\cref{tab:s05}).
Trained detectors use the thresholds given by their authors.
The three training-free methods come without a fixed threshold, so we report only their AUC.
\saysee uses a fixed threshold of 0.0272 with PE and 0.0267 with MoE-ViE.
All scores are oriented so that a higher value means more likely generated.
DEAR and \mbox{AEROBLADE} process the whole image at its original resolution.
For these two methods, the AUC uses all images of at most 4 megapixels and, for larger images, a sample stratified by evaluation set and authenticity with the corresponding weights.

\begin{table*}[t]
\centering
\footnotesize
\caption{\textbf{Compared methods, with their training data and weights.}}
\label{tab:s05}
\begin{tabular}{ll>{\raggedright\arraybackslash}p{0.36\textwidth}>{\raggedright\arraybackslash}p{0.26\textwidth}}
\toprule
Method & Venue & Training data & Weights \\
\midrule
UnivFD~\cite{ojha2023univfd} & CVPR 2023 & ProGAN & released by the authors \\
NPR~\cite{tan2024npr} & CVPR 2024 & ProGAN & released by the authors \\
DRCT~\cite{chen2024drct} & ICML 2024 & DRCT-2M (SD v1.4) & released by the authors; ConvNeXt-B and CLIP ViT-L versions \\
AIDE~\cite{yan2025aide} & ICLR 2025 & GenImage (SD v1.4) & released by the authors \\
B-Free~\cite{guillaro2025bfree} & CVPR 2025 & COCO with SD 2.1 reconstructions and inpainting & released by the authors \\
DDA~\cite{chen2025dda} & NeurIPS 2025 & COCO with SD 2.1 VAE reconstructions & released by the authors \\
GAPL~\cite{gapl2026} & CVPR 2026 & subsets of GenImage, ProGAN and Community Forensics & released by the authors \\
DEAR~\cite{dear2026} & ICML 2026 & real images of COCO and LSUN; LDM-generated images for DEAR-c, paired reconstructions for DEAR-r & released by the authors; DEAR-c and DEAR-r versions \\
PROBE~\cite{probe2026} & ICML 2026 & reconstruction-based training set (DINOv2 version), GenImage SD v1.4 (ResNet-50 version); both plus the authors' synthesized samples & released by the authors; DINOv2 and ResNet-50 versions \\
OmniAID~\cite{omniaid2026} & ICML 2026 & Mirage-Train & released by the authors; v1 and v2 \\
SimplicityPrevails~\cite{simplicityprevails2026} & ACM MM 2026 & GenImage (SD v1.4) & the authors' DINOv3 linear classifier \\
CuRe~\cite{cure2026} & arXiv 2609.14316 & GenImage (SD v1.4) & released by the authors \\
RIGID~\cite{he2024rigid} & arXiv 2405.20112 & none & DINOv2 ViT-L/14 \\
CLIDE~\cite{clide2026} & WACV 2026 & none & CLIP ViT-L/14 and the reference embeddings released by the authors \\
AEROBLADE~\cite{ricker2024aeroblade} & CVPR 2024 & none & autoencoders of SD1, SD2 and Kandinsky 2.1 \\
\bottomrule
\end{tabular}

\end{table*}

\paragraph{Reference images of CLIDE.}
CLIDE estimates whitening statistics from the embeddings of a reference set, by default 5,000 images of the MSCOCO validation set.
These images include all real images of DRCT-2M and half of those of AIGI-Now.
If we remove them from the evaluation sets, DRCT-2M can no longer be evaluated and the AUC on AIGI-Now drops from 85.9\% to 73.2\%.
The mAUC over the remaining 23 evaluation sets is then 61.1\%, compared with 63.3\% over the 24 sets with the default reference images.

\section{Comparison with existing detectors}
\label{sec:supp-comparisons}

\subsection{Method versions not listed in the main paper}

DEAR, PROBE and OmniAID each released two versions, and the main paper reports the one with the higher mAUC over the 24 evaluation sets.
\cref{tab:s06} gives the other version of each method, together with the two versions of DRCT.
The other version of each of the three methods has a lower mAUC than the version reported in the main paper, and both versions of DRCT, which the main paper does not list, fall below the best existing detector.

\begin{table}[t]
\centering
\small
\caption{\textbf{Versions of methods not listed in the main paper (\%).} The first three columns are the mean AUC of the standard, in-the-wild and latest-generator groups.}
\label{tab:s06}
\fittab{
\begin{tabular}{lccccc}
\toprule
Method & Standard & In-the-wild & Latest & mAUC & bAcc \\
\midrule
DRCT (CLIP ViT-L) & 82.7 & 67.7 & 57.5 & 68.7 & 61.5 \\
DRCT (ConvNeXt-B) & 88.3 & 77.6 & 70.8 & 78.5 & 70.5 \\
DEAR-r & 91.5 & 76.6 & 84.8 & 83.7 & 79.2 \\
PROBE (ResNet-50) & 87.4 & 77.6 & 75.9 & 79.9 & 71.4 \\
OmniAID v2 & 91.9 & 92.7 & 83.2 & 89.3 & 79.2 \\
\bottomrule
\end{tabular}
}
\end{table}

\subsection{Training-free methods on each evaluation set}

\begin{table}[t]
\centering
\small
\caption{\textbf{AUC (\%) of the training-free methods on each evaluation set.}}
\label{tab:s07}
\fittab{
\begin{tabular}{lccccc}
\toprule
 &  &  &  & \multicolumn{2}{c}{SaySee} \\
\cmidrule(lr){5-6}
Set & RIGID & CLIDE & AEROBLADE & PE & MoE-ViE \\
\midrule
GenImage & 81.0 & 55.6 & 85.3 & 99.3 & 99.6 \\
AIGCDB & 80.1 & 58.9 & 69.4 & 99.3 & 99.7 \\
DRCT-2M & 74.1 & 99.4 & 90.3 & 98.7 & 100.0 \\
Synthbuster & 61.9 & 62.5 & 74.2 & 99.9 & 100.0 \\
GenImage++ & 80.7 & 55.2 & 87.7 & 99.8 & 99.9 \\
B-Free-NG & 42.1 & 57.3 & 68.5 & 100.0 & 100.0 \\
SynthCLIC & 41.1 & 49.6 & 64.8 & 97.3 & 98.7 \\
Chameleon & 57.0 & 72.3 & 67.7 & 97.1 & 99.5 \\
WildRF & 58.9 & 65.2 & 75.9 & 99.4 & 99.7 \\
SocialRF & 60.7 & 72.1 & 61.7 & 98.5 & 99.3 \\
CommunityAI & 65.6 & 77.8 & 83.0 & 98.5 & 99.8 \\
B-Free-Online & 67.2 & 53.8 & 80.8 & 92.2 & 99.4 \\
CO-SPY-Wild & 58.9 & 67.8 & 70.7 & 98.1 & 100.0 \\
OpenFake-Reddit & 50.9 & 84.5 & 61.0 & 94.0 & 95.4 \\
RRDataset & 34.9 & 70.1 & 60.6 & 96.1 & 96.7 \\
HiRes-50K & 50.2 & 69.7 & 68.0 & 94.7 & 97.5 \\
DailyBench & 43.9 & 48.4 & 67.8 & 87.7 & 95.7 \\
So-Fake-OOD & 44.5 & 47.2 & 66.3 & 97.3 & 98.2 \\
OpenFake-Test & 75.7 & 48.8 & 82.0 & 99.4 & 99.4 \\
AIGenImages2026 & 50.0 & 50.7 & 51.8 & 94.1 & 96.3 \\
AIGI-Now & 55.2 & 85.9 & 73.7 & 99.2 & 98.8 \\
MLLMGenSet & 46.6 & 47.6 & 54.3 & 86.6 & 87.5 \\
RealChain & 44.6 & 42.9 & 70.7 & 98.0 & 96.1 \\
Mirage-Test & 36.5 & 74.4 & 62.5 & 98.3 & 98.1 \\
mAUC & 56.8 & 63.3 & 70.8 & 96.8 & 98.2 \\
\bottomrule
\end{tabular}
}
\end{table}

\cref{tab:s07} gives the per-set results behind Tab.~4.
With MoE-ViE, \saysee outperforms all three training-free methods on every one of the 24 evaluation sets, and its mAUC is 27.4 points above that of AEROBLADE, the strongest of the three.
With PE, it is slightly below CLIDE on a single set, DRCT-2M (98.7\% \vs 99.4\%), whose real images are among the reference images of CLIDE (\cref{sec:supp-repro}).

\subsection{Generators by release date}
\label{sec:supp-release}

\saysee reads a concept that the backbone acquired in pretraining (Sec.~3.1).
Here we test whether it still works for generators released after the backbone.
Of the 24 evaluation sets, 15 label each generated image with its generator, which gives 67 generators after versions of the same model are merged.
We treat the generated images of one generator in one evaluation set as a group, compute their AUC against all real images of that set, and average the groups by the release date of the generator (\cref{tab:s08}).
SimplicityPrevails, the existing detector with the highest mAUC, serves as the reference.

\begin{table}[t]
\centering
\small
\caption{\textbf{Mean AUC (\%) by release date of the generator.} The generated images of one generator in one evaluation set form a group; each group is scored against all real images of that set, and the table gives the mean over groups.}
\label{tab:s08}
\fittab{
\begin{tabular}{lcccc}
\toprule
 &  &  & \multicolumn{2}{c}{SaySee} \\
\cmidrule(lr){4-5}
Release & Generators & SimplicityPrevails (DINOv3) & PE & MoE-ViE \\
\midrule
2022 or earlier & 18 & 99.18 & 99.30 & 99.75 \\
2023 & 6 & 98.93 & 99.35 & 99.92 \\
2024 & 11 & 94.41 & 97.99 & 99.16 \\
First half of 2025 & 13 & 93.66 & 98.19 & 98.59 \\
Second half of 2025 & 14 & 88.15 & 94.80 & 96.35 \\
2026 & 5 & 78.78 & 92.39 & 95.01 \\
\bottomrule
\end{tabular}
}
\end{table}

The AUC of all three methods falls with the release date, and that of \saysee falls the least.
PE-Core was released in April 2025~\cite{bolya2025pe}, so the 19 generators released in the second half of 2025 or in 2026 cannot be in its pretraining data.
The pretraining data cutoff of MoE-ViE~\cite{moevie2026} is not known, so the analysis below centers on PE.
On these generators, the mean AUC of \saysee is 94.13\% (PE) and 95.98\% (MoE-ViE), compared with 85.54\% for SimplicityPrevails; on the generators released earlier, the three values are 98.68\%, 99.34\% and 96.54\%.
The lead of \saysee thus grows with newer generators, from 2.14 and 2.80 points to 8.59 and 10.44 points.

Generators whose names are among the 18 generator names of \say are not detected better than the others.
Among the generators released in 2024 or later, the 11 whose names are not among them have a mean AUC of 98.27\% (PE) and 98.74\% (MoE-ViE), and the 32 whose names are among them 95.91\% and 97.37\%.

These results also show what the pretraining of the backbone contributes.
No part of \saysee, the backbone included, has seen the generators released after PE-Core, yet \saysee (PE) reaches a mean AUC of 94.13\% on them, above the 85.54\% of SimplicityPrevails, which is trained on generated images.
The AUC of \saysee also falls less with the release date: from 2022 or earlier to 2026, \saysee (PE) falls from 99.30\% to 92.39\%, and SimplicityPrevails from 99.18\% to 78.78\%.
What the backbone provides is the concept of ``AI-generated'', not a memory of particular generators: the earlier CLIP has not acquired the concept, and both the minimal pair and \say are close to chance with it (\cref{sec:supp-backbone}); PE has acquired it, and it carries over to generators PE has never seen.
Building \saysee uses no generated image: the direction is given by text, and the correction and the threshold are estimated from real images (\cref{sec:supp-threshold}); what it requires of the backbone is that the concept was acquired in generic pretraining.

\subsection{Correct decisions on real and on generated images}

\cref{tab:s09} gives the shares of real and of generated images that \saysee classifies correctly at its fixed threshold, which is the same for all 24 evaluation sets.
With MoE-ViE, 98.5\% of the real images and 88.9\% of the generated images are classified correctly; on the standard group, the two shares are 99.8\% and 96.9\%.

\begin{table}[!htb]
\centering
\small
\caption{\textbf{Shares of correctly classified real and generated images for \saysee (\%).} Computed per evaluation set and then averaged.}
\label{tab:s09}
\fittab{
\begin{tabular}{llcc}
\toprule
Backbone & Scope & Real images & Generated images \\
\midrule
MoE-ViE & 24 sets & 98.5 & 88.9 \\
 & Standard & 99.8 & 96.9 \\
 & In-the-wild & 97.5 & 93.8 \\
 & Latest & 98.5 & 76.5 \\
\midrule
PE & 24 sets & 91.9 & 90.6 \\
 & Standard & 96.2 & 96.5 \\
 & In-the-wild & 86.7 & 92.3 \\
 & Latest & 94.1 & 83.5 \\
\bottomrule
\end{tabular}
}
\end{table}

\subsection{Thresholds set from real images only}
\label{sec:supp-threshold}

\begin{table*}[t]
\centering
\small
\caption{\textbf{bAcc (\%) with thresholds set from real images only.} The last two columns are the shares of real and of generated images classified correctly.}
\label{tab:s10}
\fittab[\textwidth]{
\begin{tabular}{llccccccc}
\toprule
Backbone & Threshold from & Threshold & 24 sets & Standard & In-the-wild & Latest & Real & Generated \\
\midrule
MoE-ViE & Fixed (main paper) & 0.0267 & 93.7 & 98.3 & 95.6 & 87.5 & 98.5 & 88.9 \\
 & Diverse real images & 0.0185 & 92.0 & 98.7 & 89.3 & 89.2 & 90.1 & 93.9 \\
 & Photos only & 0.0201 & 92.9 & 98.7 & 91.6 & 89.4 & 92.8 & 93.1 \\
 & + social-platform photos & 0.0228 & 93.6 & 98.6 & 93.9 & 88.8 & 95.9 & 91.3 \\
\midrule
PE & Fixed (main paper) & 0.0272 & 91.3 & 96.4 & 89.5 & 88.8 & 91.9 & 90.6 \\
 & Diverse real images & 0.0293 & 91.1 & 96.2 & 90.2 & 87.5 & 94.7 & 87.4 \\
 & Photos only & 0.0308 & 90.2 & 95.7 & 89.6 & 86.0 & 95.9 & 84.4 \\
 & + social-platform photos & 0.0275 & 91.3 & 96.3 & 89.7 & 88.7 & 92.4 & 90.2 \\
\bottomrule
\end{tabular}
}
\end{table*}

\begin{table*}[t]
\centering
\small
\caption{\textbf{False-positive rate and recall (\%) per group at the fixed threshold and at the thresholds set from real images only.} Each cell gives the false-positive rate / the recall, \ie, the shares of real and of generated images above the threshold, computed per evaluation set and then averaged within the group.}
\label{tab:s11}
\fittab[\textwidth]{
\begin{tabular}{llcccc}
\toprule
Backbone & Threshold from & Standard & In-the-wild & Latest & 24 sets \\
\midrule
MoE-ViE & Fixed (main paper) & 0.2 / 96.9 & 2.5 / 93.8 & 1.5 / 76.5 & 1.5 / 88.9 \\
 & Diverse real images & 1.2 / 98.6 & 18.6 / 97.2 & 7.9 / 86.2 & 9.9 / 93.9 \\
 & Photos only & 0.8 / 98.2 & 13.5 / 96.6 & 5.6 / 84.5 & 7.2 / 93.1 \\
 & + social-platform photos & 0.5 / 97.7 & 7.6 / 95.5 & 3.4 / 81.0 & 4.1 / 91.3 \\
\midrule
PE & Fixed (main paper) & 3.8 / 96.5 & 13.3 / 92.3 & 5.9 / 83.5 & 8.1 / 90.6 \\
 & Diverse real images & 2.3 / 94.8 & 9.0 / 89.3 & 3.8 / 78.8 & 5.3 / 87.4 \\
 & Photos only & 1.5 / 93.0 & 7.2 / 86.4 & 2.8 / 74.7 & 4.1 / 84.4 \\
 & + social-platform photos & 3.6 / 96.2 & 12.6 / 92.0 & 5.6 / 82.9 & 7.6 / 90.2 \\
\bottomrule
\end{tabular}
}
\end{table*}

The fixed threshold of the main paper is chosen on a set of labeled public images (Sec.~4.1).
Here we evaluate a fully real-only alternative: after backbone pretraining, the generation direction is constructed from text, and both the correction and the threshold are estimated from real images alone, the threshold being the 95th percentile of the scores of \saysee on a reference set of real images.
The whole detector, from the generation direction and the correction to the threshold, then needs no generated image.
With three reference sets, the bAcc is 92.0\% to 93.6\% with MoE-ViE and 90.2\% to 91.3\% with PE, compared with 93.7\% and 91.3\% with the fixed threshold; the ranking and all AUC values are unchanged.

The three reference sets are: (1) diverse real images, consisting of 30\% clean unprocessed photographs, 40\% lightly post-processed photographs, 20\% photographs with several post-processing steps, and 10\% paintings, illustrations, 3D renders, screenshots and document pages; (2) only the photographs of set (1); (3) set (1) extended with real photographs that have circulated on social platforms, where clean photographs, post-processed photographs and social-platform photographs make up 30\% each and the other 10\% is as in (1).
None of the three uses generated images or any image of the 24 evaluation sets.

A threshold only moves the decision point and does not change the ranking.
The bAcc of the three thresholds set from real images only is at most 1.7 and 1.1 points from that of the fixed threshold (\cref{tab:s10}), and the closest is the reference set with social-platform photographs added, at 93.6\% and 91.3\%.
With MoE-ViE, these thresholds lie below the fixed one; the bAcc is then lower on the in-the-wild group (89.3\% to 93.9\%, \vs 95.6\%) and higher on the standard and latest-generator groups.

\cref{tab:s11} gives the false-positive rate (the share of real images above the threshold) and the recall (the share of generated images above it) per group.
With MoE-ViE, the lower thresholds raise both: with the reference set with social-platform photographs, the recall over the 24 evaluation sets rises from 88.9\% to 91.3\% and the false-positive rate from 1.5\% to 4.1\%, mostly on the in-the-wild group (from 2.5\% to 7.6\%).
On MLLMGenSet, the fixed threshold detects only 17.8\% of the generated images, while the three thresholds set from real images only detect 25.6\% to 39.0\%, with false-positive rates of 0.8\% to 1.9\%.
With PE, all three thresholds are higher than the fixed one; the false-positive rate falls from 8.1\% to between 4.1\% and 7.6\%, and the recall from 90.6\% to between 84.4\% and 90.2\%.

\subsection{Score distributions}

\begin{figure*}[t]
\centering
\includegraphics[width=\textwidth]{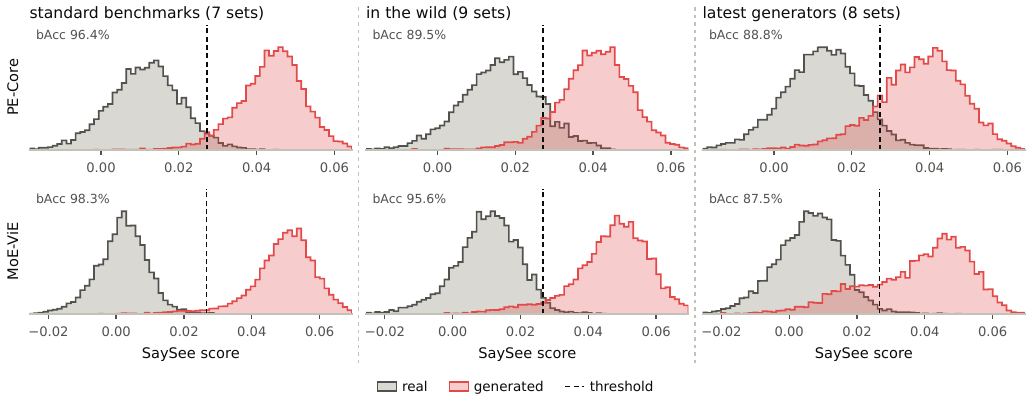}
\caption{\textbf{Score distributions of \saysee.} Distributions of the scores of real and generated images on the three groups of evaluation sets; the dashed line is the fixed threshold, and the number in the top-left corner is the mean bAcc of the group. The evaluation sets of a group have equal weight, and the horizontal axis covers 99.8\% of the images.}
\label{fig:s01}
\end{figure*}

\cref{fig:s01} shows the score distributions of \saysee for real and generated images on the three groups of evaluation sets.
On the standard and in-the-wild groups, the two distributions barely overlap.
On the latest-generator group, part of the generated images fall below the threshold, mostly from MLLMGenSet.

\subsection{MLLMGenSet}
\label{sec:supp-mllm}

MLLMGenSet is difficult for every method.
Its generated images come from GPT Image 2 and Nano Banana 2 and consist mostly of text layouts, and half of them were preprocessed and upscaled by the dataset authors.
On all images, \saysee reaches an AUC of 86.6\% (PE) and 87.5\% (MoE-ViE), more than 11 points above OmniAID v1, the best of the other methods (\cref{tab:s12}).
On the original files alone, most methods improve markedly, and \saysee exceeds 94\% with both backbones.

\begin{table}[t]
\centering
\small
\caption{\textbf{AUC (\%) on MLLMGenSet: all images, and original files only.}}
\label{tab:s12}
\fittab{
\begin{tabular}{lcc}
\toprule
Method & All images & Original files only \\
\midrule
UnivFD & 56.4 & 63.6 \\
NPR & 56.3 & 68.7 \\
AIDE & 73.6 & 89.3 \\
B-Free & 52.3 & 58.9 \\
DDA & 69.5 & 88.6 \\
GAPL & 72.1 & 88.6 \\
DEAR-c & 72.5 & 97.0 \\
PROBE (DINOv2) & 68.3 & 84.0 \\
OmniAID v1 & 75.1 & 92.9 \\
CuRe & 59.7 & 62.8 \\
SimplicityPrevails (DINOv3) & 61.5 & 64.8 \\
SaySee (PE) & 86.6 & 94.3 \\
SaySee (MoE-ViE) & 87.5 & 95.2 \\
\bottomrule
\end{tabular}
}
\end{table}

\begin{figure*}[t]
\centering
\includegraphics[width=0.94\textwidth]{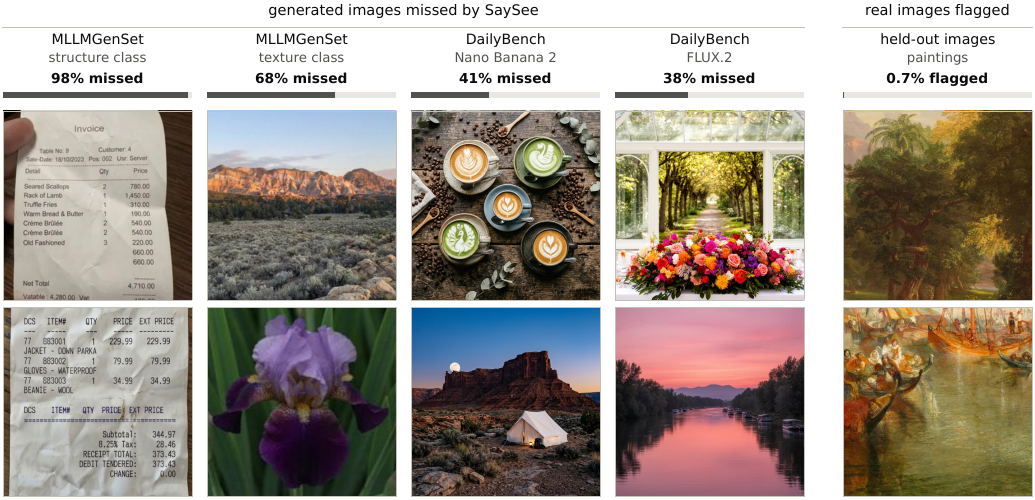}
\caption{\textbf{Failure cases of \saysee (MoE-ViE).} The first four columns are missed generated images: the structural and texture categories of MLLMGenSet, and the images of Nano Banana 2 and FLUX.2 in DailyBench; the last column shows held-out paintings of \see (\cref{tab:s02}) that are classified as generated. The error rate of each group is given above its column. The two images of a column are those at the 25\% and 75\% positions when the misclassified images of the group are sorted by score; images whose main subject is a face, that involve real persons or organizations, or whose license does not allow reprinting are not printed and are replaced by the nearest one in rank.}
\label{fig:s02}
\end{figure*}

With MoE-ViE, almost all errors of \saysee on this set are misses: at the fixed threshold, 99.7\% of the real images are classified correctly and 17.8\% of the generated images are detected.
The miss rate is 92\% on the preprocessed half and 72\% on the original files.
Of the three artifact categories defined by the dataset authors, the text-dominated structural category is missed most often, while the mixed and texture categories have miss rates of 81\% and 68\%.
Images generated by GPT Image 2 and Nano Banana 2 are detected at clearly higher rates in other evaluation sets such as DailyBench.
The misses on this set therefore come mainly from the text-dominated content and the preprocessing, and much less from the generators themselves.
On the original files alone, the mAUC of \saysee over the 24 evaluation sets is 98.5\% (MoE-ViE) and 97.1\% (PE).

\subsection{Failure cases}
\label{sec:supp-failures}

\cref{fig:s02} shows the groups of generated images where misses concentrate, together with real images classified as generated.
On MLLMGenSet, almost all errors are misses, and they concentrate in text-dominated generated images (\cref{sec:supp-mllm}).
The AUC between these images and the real images of the set is 71.2\% with \say alone and 80.4\% with \saysee, so the misses originate in the generation direction itself, which separates this kind of image poorly.
On DailyBench, the misses concentrate in the images of Nano Banana 2 and FLUX.2, while the miss rates of GPT Image 2, FLUX and SD~3.5 do not exceed 4\%.
Among the held-out real images of \see (\cref{tab:s02}), those above the threshold with MoE-ViE are mostly 3D renders: 22 of 340, compared with 11 of the 4,072 images of the other five image types.
3D renders are an image type that is not orthogonal to the generation direction, and since \see corrects each image type by its mean offset, an individual image can still exceed the threshold.

\section{Comparison with same-backbone trained methods}
\label{sec:supp-probe}

\subsection{Training settings}
\label{sec:supp-training}

\paragraph{Whole-image resize protocol:} the protocol used in Tab.~5 of the main paper, which resizes the whole image to $448\times448$. It is also the prevailing input scheme and is used by most linear probing and fine-tuning methods.
We train both the linear classifier and the LoRA fine-tuning on the SD~v1.4 training set of GenImage (323,997 images) without data augmentation, with a batch size of 128 for 2 epochs (5,064 steps), and use the weights at the end of training.
The linear classifier has a learning rate of $10^{-3}$.
LoRA, with rank 8, alpha 8 and a learning rate of $10^{-4}$, is trained jointly with the linear classifier.
With PE, it is applied to the q, k, v and o projections of the 50 blocks and of the attention pooling; with MoE-ViE, also to the gate, up and down projections of the shared experts.

\paragraph{Five-crop protocol:} the protocol used by \saysee, which reads five crops at native resolution and averages their embeddings.
To rule out the effect of how images are read, we retrain the linear classifier and the LoRA fine-tuning on the same image embedding as \saysee, keeping the training set, the optimization settings and the number of steps of the whole-image resize protocol.
The linear probe takes the mean of the five crop embeddings as input, \ie, the image embedding of \saysee.
The linear head of the LoRA fine-tuning takes normalized crop embeddings as input: during training, one of the five crops of each image is drawn at random at every step, and at test time the five crop embeddings are averaged as in \saysee.

\subsection{Five-crop protocol: results on the 24 evaluation sets}

\begin{table*}[t]
\centering
\small
\caption{\textbf{Same-backbone trained methods and \saysee on the 24 evaluation sets (\%).} The first three columns are the mean AUC of the standard, in-the-wild and latest-generator groups; the bAcc threshold is 0.5 for the linear classifier, the linear probe and LoRA, and the fixed threshold for \saysee; no bAcc is reported for the linear probe with \see (\cref{sec:supp-probe-see}). The protocol of each trained method is given in parentheses (\cref{sec:supp-training}).}
\label{tab:s13}
\fittab[0.7\textwidth]{
\begin{tabular}{llcccll}
\toprule
Backbone & Method & Standard & In-the-wild & Latest & mAUC & bAcc \\
\midrule
MoE-ViE & Linear classifier (whole-image resize) & 98.88 & 96.39 & 90.63 & 95.20 & 92.32 \\
 & LoRA fine-tuning (whole-image resize) & 99.25 & 96.73 & 90.65 & 95.44 & 83.04 \\
 & Linear probe (five-crop) & 99.21 & 98.10 & 93.62 & 96.93 & 91.86 \\
 & Linear probe (five-crop + JPEG) & 99.12 & 98.26 & 93.58 & 96.95 & 91.89 \\
 & Linear probe + See (five-crop) & 99.42 & 97.91 & 93.36 & 96.83 & -- \\
 & LoRA fine-tuning (five-crop) & 90.07 & 77.78 & 72.51 & 79.61 & 64.59 \\
 & LoRA fine-tuning (five-crop + JPEG) & 91.21 & 83.23 & 74.13 & 82.52 & 68.32 \\
 & SaySee & 99.71 & 98.60 & 96.28 & \textbf{98.15} & 93.69 \\
\midrule
PE & Linear classifier (whole-image resize) & 98.87 & 89.69 & 91.20 & 92.87 & 85.97 \\
 & LoRA fine-tuning (whole-image resize) & 98.13 & 94.78 & 87.28 & 93.26 & 77.18 \\
 & Linear probe (five-crop) & 98.43 & 94.12 & 94.33 & 95.45 & 91.20 \\
 & Linear probe (five-crop + JPEG) & 98.45 & 94.21 & 94.29 & 95.47 & 91.24 \\
 & Linear probe + See (five-crop) & 98.90 & 93.52 & 93.75 & 95.17 & -- \\
 & LoRA fine-tuning (five-crop) & 92.16 & 88.91 & 78.06 & 86.24 & 65.71 \\
 & LoRA fine-tuning (five-crop + JPEG) & 95.62 & 92.33 & 77.86 & 88.47 & 67.83 \\
 & SaySee & 99.18 & 96.49 & 95.07 & \textbf{96.80} & 91.29 \\
\bottomrule
\end{tabular}
}
\end{table*}

\cref{tab:s13} also lists the results under the whole-image resize protocol.
The linear probe under the five-crop protocol has a higher mAUC than the linear classifier under the whole-image resize protocol (96.93\% \vs 95.20\% with MoE-ViE, 95.45\% \vs 92.87\% with PE); part of this gain comes from compression traces (\cref{sec:supp-probe-format}).
Even so, \saysee, which uses no generated image except to set its threshold, has a higher mAUC and a higher bAcc than the linear probe trained on 320 thousand images, with both backbones.
With MoE-ViE, the mAUC is 98.15\% \vs 96.93\% and the bAcc 93.69\% \vs 91.86\%; with PE, the mAUC is 96.80\% \vs 95.45\% and the bAcc 91.29\% \vs 91.20\%.
It is also higher on all three groups of evaluation sets.

LoRA fine-tuning degrades sharply under this protocol: its mAUC drops from 95.44\% and 93.26\% under the whole-image resize protocol to 79.61\% and 86.24\%.
It remains very good on its training distribution (AUC of 98.5\% and 99.5\% on GenImage) but falls sharply on the other evaluation sets.
The reason lies in the training set.
In the SD~v1.4 training set of GenImage, all real images are JPEG files and all generated images are $512\times512$ PNG files~\cite{grommelt2024fake}.
Crops at native resolution preserve this difference in compression format, and once the backbone is fine-tuned, the model bases its decision on the file format.
At test time, the five-crop LoRA fine-tuning detects only 1.2\% (MoE-ViE) and 7.8\% (PE) of the generated JPEG images at threshold 0.5, compared with 51.9\% and 39.8\% for LoRA fine-tuning under the whole-image resize protocol.
On the 4 evaluation sets whose real images are all JPEG and whose generated images are all in other formats (GenImage, GenImage++, CommunityAI and CO-SPY-Wild), the AUC stays above 96\%.
On the 8 sets whose real and generated images share one format, the AUC drops markedly on most of them, down to 43.3\% and 37.9\% on Mirage-Test; only DRCT-2M, whose generators belong to the same Stable Diffusion family as the training set, is essentially unchanged.

File formats also benefit the linear probe on the 24 evaluation sets.
In 8 of the 24 sets, the real images are mostly JPEG and the generated images mostly PNG or WebP (the share of JPEG files is more than 50 percentage points higher among the real images than among the generated ones), the same format composition as the training set: GenImage, GenImage++, AIGCDB, CommunityAI, CO-SPY-Wild, OpenFake-Test, So-Fake-OOD and \mbox{AIGenImages2026}.
\cref{tab:s14} gives the mean AUC for this split.
The linear probe has a higher mean AUC than \saysee only on these 8 sets.
On the other 16 evaluation sets, the mean AUC of \saysee is higher with both backbones, by 1.85 points with MoE-ViE and 2.48 points with PE.
The linear classifier and LoRA fine-tuning under the whole-image resize protocol are below \saysee on both groups, by 3.5 to 5.3 points on the other 16 evaluation sets.
The five-crop LoRA fine-tuning shows the largest difference between the two groups.

\begin{table*}[t]
\centering
\small
\caption{\textbf{Mean AUC (\%) with the evaluation sets grouped by their file formats.} Format-separated: the 8 sets whose real images are mostly JPEG and whose generated images are mostly not. The protocol of each trained method is given in parentheses.}
\label{tab:s14}
\fittab[0.7\textwidth]{
\begin{tabular}{llcc}
\toprule
Backbone & Method & Format-separated (8 sets) & Other 16 sets \\
\midrule
MoE-ViE & Linear classifier (whole-image resize) & 97.15 & 94.22 \\
 & LoRA fine-tuning (whole-image resize) & 98.02 & 94.15 \\
 & Linear probe (five-crop) & 99.16 & 95.82 \\
 & LoRA fine-tuning (five-crop) & 91.76 & 73.53 \\
 & SaySee & 99.12 & 97.67 \\
\midrule
PE & Linear classifier (whole-image resize) & 97.09 & 90.76 \\
 & LoRA fine-tuning (whole-image resize) & 96.47 & 91.65 \\
 & Linear probe (five-crop) & 99.12 & 93.61 \\
 & LoRA fine-tuning (five-crop) & 95.91 & 81.41 \\
 & SaySee & 98.22 & 96.09 \\
\bottomrule
\end{tabular}
}
\end{table*}

\begin{table*}[t]
\centering
\small
\caption{\textbf{Evaluation with format decoupled from authenticity (\%).} Evaluation as in Tab.~6. mAUC: mean over the 16 pairings; Same and Diff.: mean mAUC over the 4 pairings with the same version on both sides and over the 12 pairings with different versions. The protocol of each trained method is given in parentheses.}
\label{tab:s15}
\fittab[0.7\textwidth]{
\begin{tabular}{llccccc}
\toprule
Backbone & Method & mAUC & Format gap & Worst pairing & Same & Diff. \\
\midrule
MoE-ViE & Linear classifier (whole-image resize) & 93.85 & 5.74 & 86.97 & 94.19 & 93.74 \\
 & LoRA fine-tuning (whole-image resize) & 93.37 & 5.36 & 86.79 & 93.85 & 93.21 \\
 & Linear probe (five-crop) & 93.64 & 5.94 & 89.74 & 94.85 & 93.24 \\
 & Linear probe (five-crop + JPEG) & 95.23 & 2.92 & 92.74 & 94.65 & 95.42 \\
 & Linear probe + See (five-crop) & 94.66 & 3.04 & 92.52 & 95.03 & 94.54 \\
 & LoRA fine-tuning (five-crop) & 68.90 & 44.61 & 11.96 & 76.51 & 66.37 \\
 & LoRA fine-tuning (five-crop + JPEG) & 76.33 & 35.22 & 34.52 & 78.31 & 75.67 \\
 & SaySee & 97.19 & 2.48 & 94.57 & 97.38 & 97.12 \\
\midrule
PE & Linear classifier (whole-image resize) & 89.90 & 13.64 & 72.17 & 91.58 & 89.34 \\
 & LoRA fine-tuning (whole-image resize) & 91.25 & 8.82 & 81.11 & 91.98 & 91.01 \\
 & Linear probe (five-crop) & 91.45 & 18.23 & 57.53 & 96.36 & 89.82 \\
 & Linear probe (five-crop + JPEG) & 93.92 & 8.92 & 77.14 & 95.20 & 93.49 \\
 & Linear probe + See (five-crop) & 94.65 & 5.32 & 89.41 & 95.00 & 94.53 \\
 & LoRA fine-tuning (five-crop) & 78.20 & 33.26 & 34.63 & 85.49 & 75.76 \\
 & LoRA fine-tuning (five-crop + JPEG) & 86.20 & 22.16 & 59.12 & 89.66 & 85.05 \\
 & SaySee & 95.66 & 2.80 & 93.56 & 95.69 & 95.65 \\
\bottomrule
\end{tabular}
}
\end{table*}

\subsection{Five-crop protocol: evaluation with format decoupled from authenticity}
\label{sec:supp-probe-format}

The trained methods of both protocols share one training set, so the evaluation of Sec.~5.4 directly tests how much each of them relies on format (\cref{tab:s15}).

With format decoupled from authenticity, all four trained methods of \cref{sec:supp-training} fall behind \saysee: their mean mAUC over the 16 pairings and their worst pairing are lower, and their format gap is more than twice that of \saysee.
With PE, the linear probe has a format gap of 18.23 points and a worst pairing of only 57.53\%, compared with 2.80 points and 93.56\% for \saysee.
With MoE-ViE, the format gap and the worst pairing are 5.94 points and 89.74\% for the linear probe, and 2.48 points and 94.57\% for \saysee.
When both sides are in the same version, format offers no cue; the mean mAUC is then 94.85\% for the linear probe and 97.38\% for \saysee with MoE-ViE, and 96.36\% and 95.69\% with PE.
When the versions differ, the mean mAUC of the linear probe with PE drops to 89.82\%, compared with 95.65\% for \saysee.

LoRA fine-tuning under the five-crop protocol relies on format the most.
Its format gap is 44.61 points with MoE-ViE and 33.26 points with PE, and its worst pairing reaches only 11.96\% and 34.63\%.
These values are below 50\%, which means that under that pairing, real images mostly score higher than generated ones.
Under the whole-image resize protocol, the format gap of LoRA fine-tuning is much smaller, 5.36 and 8.82 points.

Part of the gain of the linear probe also comes from compression traces.
Over the 24 evaluation sets, the five-crop linear probe is 1.73 (MoE-ViE) and 2.58 (PE) points above the linear classifier under the whole-image resize protocol.
With format decoupled from authenticity, its mean mAUC over the 16 pairings is instead 0.21 points lower with MoE-ViE and only 1.55 points higher with PE.
On the pairings with different versions, it falls below the whole-image linear classifier with MoE-ViE (93.24\% \vs 93.74\%) and is only 0.48 points higher with PE (89.82\% \vs 89.34\%); with PE, its format gap also grows from 13.64 to 18.23 points and its worst pairing drops from 72.17\% to 57.53\%.
Resizing the whole image destroys compression traces, so they cannot be learned, whereas crops at native resolution preserve them and turn them into a shortcut.
\cref{tab:s14} is consistent with this: the linear probe has a higher mean AUC than \saysee only on the 8 sets whose format composition matches the training set.

\paragraph{JPEG-aligned training.}
To test this shortcut directly, we compress every generated training image as JPEG with the quantization tables and chroma subsampling of a randomly paired real training image, so that the JPEG settings of the two classes follow the same distribution, and retrain the linear probe and the LoRA fine-tuning with all other settings of the five-crop protocol (rows ``five-crop + JPEG'' in \cref{tab:s13,tab:s15}).
After the alignment, the probe depends much less on format: its format gap falls from 5.94 and 18.23 points to 2.92 and 8.92 points, and its worst pairing rises from 89.74\% and 57.53\% to 92.74\% and 77.14\%, while its mAUC over the 24 evaluation sets is essentially unchanged (96.95\% and 95.47\%).
The format gap of the LoRA fine-tuning also falls, from 44.61 and 33.26 points to 35.22 and 22.16 points, but remains larger than that of every other method except the unaligned five-crop LoRA fine-tuning.
The aligned probe still falls behind \saysee: its mAUC over the 24 evaluation sets is lower by 1.20 (MoE-ViE) and 1.33 (PE) points and it is lower on all three groups; its mean mAUC over the 16 pairings is lower by 1.96 and 1.74 points, its worst pairing by 1.83 and 16.42 points, and its format gap is larger.

\subsection{Five-crop protocol: linear probe with \see}
\label{sec:supp-probe-see}

To separate the roles of the generation direction and of the real-image correction, we add \see to the linear probe of the five-crop protocol: the prototypes are unchanged, the offsets of all conditions are recomputed along the weight direction of the probe, and the temperatures are re-selected by the original rule, exactly as \see is rebuilt for each description set in \cref{sec:supp-condfree}; no generated image is used (\cref{tab:s13,tab:s15}).

\see also reduces the dependence of the probe on format.
The worst pairing rises from 89.74\% to 92.52\% with MoE-ViE and from 57.53\% to 89.41\% with PE, the format gap falls from 5.94 and 18.23 points to 3.04 and 5.32 points, and the mean mAUC over the 16 pairings rises from 93.64\% and 91.45\% to 94.66\% and 94.65\%.
The mAUC over the 24 evaluation sets, however, does not rise (96.93\% \vs 96.83\% with MoE-ViE, 95.45\% \vs 95.17\% with PE), and the in-the-wild and latest-generator groups drop slightly.
The same \see added to \say raises both the mAUC over the 24 evaluation sets and the latest-generator group (Tab.~7, \cref{tab:s20}).

With the same correction, \saysee is ahead of the probe with \see on the 24 evaluation sets, on all three groups, on the mean mAUC over the 16 pairings, on the format gap and on the worst pairing.
Its mAUC over the 24 evaluation sets is higher by 1.32 (MoE-ViE) and 1.63 (PE) points, the latest-generator group by 2.92 and 1.32 points, and the worst pairing by 2.05 and 4.15 points.
The real-image correction thus also reduces the dependence of a trained detector on format, while under the same correction the generation direction given by language generalizes better.
No bAcc is reported for the probe with \see: the correction shifts the scores, so the original threshold of the probe no longer applies.

\FloatBarrier
\section{Evaluation with format decoupled from authenticity}
\label{sec:supp-format}

\subsection{Per-pairing results}
\label{sec:supp-pairings}

\begin{figure*}[t]
\centering
\includegraphics[width=\textwidth]{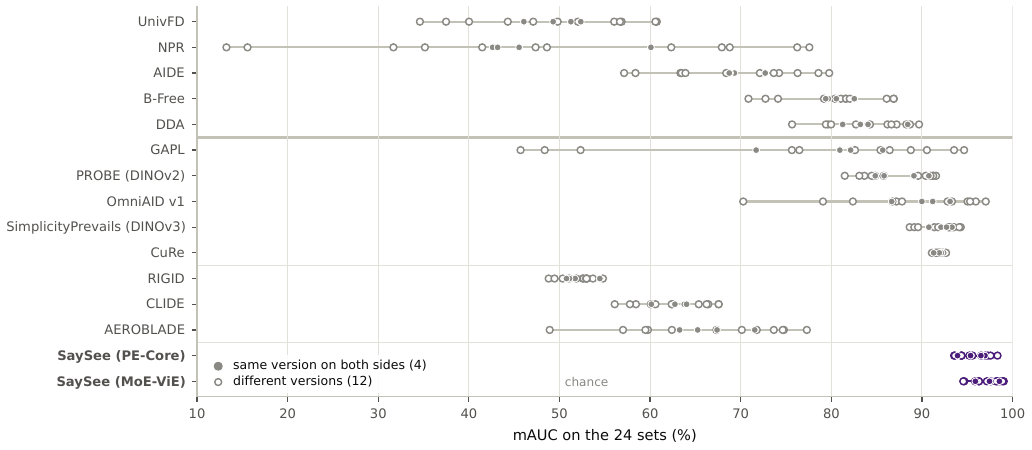}
\caption{\textbf{mAUC of every method under all 16 pairings.} One row per method, in the order of Tab.~6. Filled circles are the 4 pairings with the same version on both sides, and open circles the 12 pairings with different versions; the line joins the lowest and the highest value.}
\label{fig:s03}
\end{figure*}

\cref{fig:s03} shows the mAUC of every method of Tab.~6 under all 16 pairings, and \cref{tab:s16} lists the values of \saysee.
The 16 values of \saysee range from 94.6\% to 99.0\% with MoE-ViE and from 93.6\% to 98.3\% with PE, so its worst pairing is above the mean over the 16 pairings of every other method (the highest being 92.1\%, Tab.~6).
Most trained detectors vary widely across the pairings with different versions on the two sides, and in directions that differ from detector to detector.

\begin{table}[t]
\centering
\small
\caption{\textbf{mAUC (\%) of \saysee under the 16 pairings.} Rows: version of the real images; columns: version of the generated images.}
\label{tab:s16}
\fittab{
\begin{tabular}{llcccc}
\toprule
Backbone & Real $\backslash$ Generated & Unprocessed & JPEG 75 & WebP 80 & Triple re-post \\
\midrule
MoE-ViE & Unprocessed & 98.55 & 97.62 & 96.29 & 94.67 \\
 & JPEG 75 & 98.57 & 97.62 & 96.29 & 94.57 \\
 & WebP 80 & 99.03 & 98.32 & 97.46 & 95.86 \\
 & Triple re-post & 98.90 & 98.19 & 97.16 & 95.90 \\
\midrule
PE & Unprocessed & 96.96 & 95.23 & 93.56 & 94.36 \\
 & JPEG 75 & 97.00 & 95.37 & 93.62 & 94.36 \\
 & WebP 80 & 97.27 & 95.57 & 93.93 & 94.35 \\
 & Triple re-post & 98.34 & 97.57 & 96.53 & 96.51 \\
\bottomrule
\end{tabular}
}
\end{table}

\subsection{Effect of compression on the scores of real and generated images}
\label{sec:supp-compression}

\begin{figure}[t]
\centering
\includegraphics[width=0.78\linewidth]{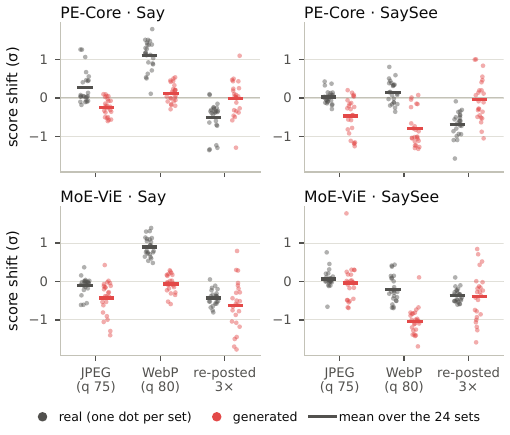}
\caption{\textbf{Score changes after compression and re-posting.} Each dot is the mean change in one evaluation set, and the line the mean over the 24 sets; unit: within-set standard deviation of unprocessed real images.}
\label{fig:s04}
\end{figure}

\see measures offsets on real images only and applies the same correction to generated images.
\cref{fig:s04} compares the mean score change of the two kinds of images after compression and after re-posting.
In this section, $\sigma$ denotes the within-set standard deviation of the scores of unprocessed real images.
With \say alone, WebP compression raises the scores of real images by $0.90\sigma$ (MoE-ViE) and $1.10\sigma$ (PE), while the scores of generated images hardly change.
\see restores the scores of real images: after the correction, their mean change under WebP compression shrinks to $0.20\sigma$ and $0.15\sigma$ in magnitude, and under JPEG compression it does not exceed $0.06\sigma$.
The scores of generated images drop by $1.03\sigma$ and $0.80\sigma$ under WebP compression, and this is where the remaining format gap of \saysee in Tab.~6 comes from: of two pairings with swapped versions, the one whose generated images are processed more heavily has the lower mAUC.

\subsection{Generalization of \see to unseen formats}
\label{sec:supp-unseen}

The format family of \see covers only three codec families: lossless, JPEG and WebP.
This section reports the held-out codec experiment mentioned in Sec.~5.5.
It shows that the correction of \see generalizes to formats that never appear among its prototypes.

\paragraph{Setting.}
We save the unprocessed versions of all images of the evaluation with format decoupled from authenticity as AVIF, JPEG XL, HEIC and JPEG 2000.
None of these four formats is among the 133 conditions of \see, and none of the reference photographs is stored in them.
We encode with the default settings of each library; the JPEG 2000 library has no default lossy quality, so we set its bitrate equal to that of AVIF (\cref{tab:s17}).
\see is left unchanged.
Together with the 4 existing versions, each new format forms 9 pairings that contain it.
We report the mean mAUC over these 9 pairings, the worst pairing, and the format gap when the new format is swapped with the other 4 versions.

\begin{table}[H]
\centering
\small
\caption{\textbf{Encoder libraries, settings and mean bitrates of the versions.} Bitrates are in bits per pixel.}
\label{tab:s17}
\fittab{
\begin{tabular}{lllc}
\toprule
Version & Encoder library & Settings & Mean bitrate \\
\midrule
JPEG 75 & Pillow 11.3.0 & quality 75 & 1.10 \\
WebP 80 & Pillow 11.3.0 & quality 80 & 0.91 \\
AVIF & libavif 1.3.0 & default (quality 75) & 0.99 \\
JPEG XL & libjxl 0.12.0 & default (quality 90) & 1.28 \\
HEIC & libheif 1.23.6, x265 4.3 & default (quality 50) & 0.88 \\
JPEG 2000 & OpenJPEG 2.5.3 & same bitrate as AVIF & 0.99 \\
\bottomrule
\end{tabular}
}
\end{table}

\paragraph{Results.}
Although \see has seen none of these four formats, it lowers the format gap and raises the worst pairing for all of them with both backbones (\cref{tab:s18}).
For AVIF with MoE-ViE, the format gap falls from 6.40 to 2.29 points and the worst pairing rises from 87.91\% to 93.65\%; with PE, the format gap falls from 5.33 to 2.93 points and the worst pairing rises from 89.18\% to 91.27\%.
Over the four formats together, the format gap falls from 4.54 to 1.90 points with MoE-ViE and from 4.52 to 2.78 points with PE, reductions of 58\% and 38\%.
The resulting format gap is no larger than the one among the four versions that \see has seen (2.48 and 2.80 points, Tab.~6).
The mean mAUC of \saysee is 96.97\% and 94.82\% on the pairings that contain a new format, and 97.19\% and 95.66\% on the seen versions.
In terms of the format gap, unseen formats behave like seen ones.

\begin{table}[!htb]
\centering
\small
\caption{\textbf{Results on the pairings that contain a new format (\%).} Each cell gives \say $\rightarrow$ \saysee.}
\label{tab:s18}
\fittab{
\begin{tabular}{lllll}
\toprule
Backbone & Format & Mean mAUC & Format gap & Worst pairing \\
\midrule
MoE-ViE & AVIF & 95.98 $\rightarrow$ 97.24 & 6.40 $\rightarrow$ 2.29 & 87.91 $\rightarrow$ 93.65 \\
 & JPEG XL & 96.89 $\rightarrow$ 97.39 & 3.16 $\rightarrow$ 1.61 & 92.00 $\rightarrow$ 94.77 \\
 & HEIC & 96.27 $\rightarrow$ 97.20 & 4.80 $\rightarrow$ 1.72 & 89.46 $\rightarrow$ 94.37 \\
 & JPEG 2000 & 95.70 $\rightarrow$ 96.04 & 3.78 $\rightarrow$ 1.97 & 90.52 $\rightarrow$ 94.45 \\
 & All four & 96.21 $\rightarrow$ 96.97 & 4.54 $\rightarrow$ 1.90 & 87.91 $\rightarrow$ 93.65 \\
\midrule
PE & AVIF & 94.50 $\rightarrow$ 94.66 & 5.33 $\rightarrow$ 2.93 & 89.18 $\rightarrow$ 91.27 \\
 & JPEG XL & 95.75 $\rightarrow$ 95.71 & 3.07 $\rightarrow$ 2.01 & 90.87 $\rightarrow$ 93.39 \\
 & HEIC & 94.89 $\rightarrow$ 95.02 & 4.04 $\rightarrow$ 2.65 & 91.06 $\rightarrow$ 93.17 \\
 & JPEG 2000 & 93.76 $\rightarrow$ 93.90 & 5.64 $\rightarrow$ 3.55 & 88.56 $\rightarrow$ 89.65 \\
 & All four & 94.73 $\rightarrow$ 94.82 & 4.52 $\rightarrow$ 2.78 & 88.56 $\rightarrow$ 89.65 \\
\bottomrule
\end{tabular}
}
\end{table}

\paragraph{Where the generalization comes from.}
\see infers the conditions of an image from how close its embedding is to each prototype, and never needs to know the actual format.
A new format lies close to formats that have been measured, and \see corrects it with their offsets.
\cref{tab:s19} gives the posterior of the format family for the images of each version, summed by codec family.
AVIF, HEIC and JPEG 2000 images are assigned mainly to WebP prototypes, and JPEG XL images are split between the JPEG and WebP families, whereas WebP 80 and JPEG 75 images go almost entirely to their own family.

\begin{table}[!htb]
\centering
\small
\caption{\textbf{Mean posterior of the format family for the images of each version, summed by codec family.}}
\label{tab:s19}
\fittab{
\begin{tabular}{llccc}
\toprule
Backbone & Version & Lossless & JPEG & WebP \\
\midrule
MoE-ViE & AVIF & 0.001 & 0.146 & 0.853 \\
 & JPEG XL & 0.022 & 0.439 & 0.538 \\
 & HEIC & 0.004 & 0.140 & 0.856 \\
 & JPEG 2000 & 0.023 & 0.309 & 0.668 \\
 & WebP 80 & 0.001 & 0.032 & 0.966 \\
 & JPEG 75 & 0.000 & 0.992 & 0.008 \\
\midrule
PE & AVIF & 0.004 & 0.278 & 0.718 \\
 & JPEG XL & 0.048 & 0.593 & 0.358 \\
 & HEIC & 0.013 & 0.169 & 0.818 \\
 & JPEG 2000 & 0.083 & 0.225 & 0.693 \\
 & WebP 80 & 0.000 & 0.006 & 0.994 \\
 & JPEG 75 & 0.000 & 0.994 & 0.006 \\
\bottomrule
\end{tabular}
}
\end{table}

\paragraph{Adding a format.}
Adding a new format to the prototypes of \see only requires encoding the reference photographs of the format family in that format and estimating its prototypes and offsets.
No generated image is needed.

\FloatBarrier
\section{Further ablations}
\label{sec:supp-ablations}

\subsection{The two modules on the three groups of evaluation sets}

\begin{table}[H]
\centering
\small
\caption{\textbf{Mean AUC (\%) of the rows of Tab.~7 on the three groups of evaluation sets.}}
\label{tab:s20}
\fittab{
\begin{tabular}{llcccc}
\toprule
Backbone & Method & Standard & In-the-wild & Latest & mAUC \\
\midrule
MoE-ViE & Minimal pair & 97.48 & 93.90 & 89.52 & 93.48 \\
 & Minimal pair + See & 96.83 & 91.49 & 91.61 & 93.09 \\
 & Say & 99.76 & 98.77 & 95.55 & 97.99 \\
 & Say + See (SaySee) & 99.71 & 98.60 & 96.28 & 98.15 \\
\midrule
PE & Minimal pair & 98.60 & 95.71 & 93.21 & 95.72 \\
 & Minimal pair + See & 98.59 & 94.34 & 92.26 & 94.89 \\
 & Say & 99.32 & 96.59 & 93.44 & 96.34 \\
 & Say + See (SaySee) & 99.18 & 96.49 & 95.07 & 96.80 \\
\bottomrule
\end{tabular}
}
\end{table}

\cref{tab:s20} breaks down the results of Tab.~7 by group.
\say is above the minimal pair on all three groups.
With MoE-ViE, the gain is largest on the hardest group, the latest generators, which rises from 89.52\% to 95.55\%; with PE, this group rises from 93.21\% to 93.44\%.
\see raises it further, by 0.73 points with MoE-ViE and 1.63 points with PE.

\subsection{Generator names}
\label{sec:supp-names}

\cref{tab:s21} tests whether the effect of the generator names comes from what the names mean.
Adding the 18 generator names raises both the mAUC over the 24 evaluation sets and the mean mAUC over the 16 pairings, with both backbones.
Replacing them with 18 fictional names, 18 names of image software unrelated to AI, or 6 generic terms for AI tools gives lower values than \say on both metrics, and an mAUC over the 24 sets below that obtained without any names.
The effect therefore comes from the meaning of the names; extra sentences on the generated side bring no gain by themselves.

\begin{table}[H]
\centering
\small
\caption{\textbf{Ablation of the generator names (\%).} \see is off in all rows. mAUC is over the 24 evaluation sets; ``16 pairings'' is the mean mAUC over the 16 pairings.}
\label{tab:s21}
\fittab{
\begin{tabular}{lcccc}
\toprule
 & \multicolumn{2}{c}{MoE-ViE} & \multicolumn{2}{c}{PE} \\
\cmidrule(lr){2-3}\cmidrule(lr){4-5}
Added descriptions & mAUC & 16 pairings & mAUC & 16 pairings \\
\midrule
None & 97.66 & 96.24 & 96.13 & 94.48 \\
18 generator names (Say) & 97.99 & 96.51 & 96.34 & 94.92 \\
18 fictional names & 97.35 & 96.20 & 94.88 & 93.59 \\
18 names of image software & 97.40 & 96.22 & 95.26 & 94.12 \\
6 generic terms for AI tools & 97.42 & 96.07 & 95.54 & 93.91 \\
\bottomrule
\end{tabular}
}
\end{table}

Examples of the fictional names are Zorvex and Lumibra, and the image software includes Photoshop, Blender and Procreate.
The generic terms are ``a diffusion model'', ``a neural network'', ``a generative model'', ``an AI image generator'', ``a machine learning model'' and ``a generative adversarial network''.
All three replacements use the same sentence pattern and the same aggregation as the generator names.

\subsection{Condition-free paired descriptions}
\label{sec:supp-condfree}

In the main paper, ``paired descriptions'' refers to the complete description set of \say.
Going from the minimal pair to \say changes four things: each side goes from one description to a set of templates, the generated side gains the generator names, the descriptions go from one combination to 50, and both sides state the image type and the content.
\cref{sec:supp-names} examines the generator names.
To separate the effect of adding combinations from that of stating the conditions, we construct paired descriptions that state neither the image type nor the content, hereafter \emph{condition-free}.
These are the size-matched prompt controls mentioned in Sec.~5.5.

\paragraph{Construction.}
We construct two condition-free sets of paired descriptions.
The first has a single combination whose noun is always ``image'': the generated side uses the 4 templates and the 18 generator names of \say, and the real side uses 4 templates that name no image type (``a real image'', ``an authentic image'', ``an image made by a person'', ``an image created by a human'').
The second has the same size as \say, 50 combinations.
In it, six nouns that name no image type take the place of the image types: ``image'', ``picture'', ``digital image'' and ``digital picture'' with 12 combinations each, and ``visual'' and ``still image'' with one each, matching the four image types with 12 combinations and the screenshot and document page with one combination in \say.
Twelve sentence frames that name no content take the place of the contents: no frame, and ``this is \{\}'', ``it looks like \{\}'', ``here is \{\}'', ``this appears to be \{\}'', ``an example of \{\}'', ``\{\}, shown here'', ``this seems to be \{\}'', ``it is \{\}'', ``we see \{\}'', ``a typical example of \{\}'' and ``look at \{\}''.
Both sides of a combination use the same frame, with the templates of the first set.
We also construct two sets that state only the image type or only the content, again with 50 combinations, by replacing the nouns of the second set with the 6 image types or its frames with the 12 content options.
Every set uses the generator names of \say, except for the last two rows of \cref{tab:s22,tab:s23}, which are two sets without generator names, both with 50 combinations.
All sets are aggregated by Eqs.~(4) and~(5) as in \say, and \see is re-estimated on each direction with the original procedure.

\cref{tab:s22} gives the results of the complete method (with \see), and \cref{tab:s23} those of the generation direction alone (without \see).
From top to bottom, the first six rows of both tables add the components of \say in three steps: rows 1 to 2 add the templates and the generator names, rows 2 to 3 add combinations, and rows 3 to 6 state the image type and the content, with rows 4 and 5 stating only one of the two.

\begin{table*}[t]
\centering
\small
\caption{\textbf{Ablation of the design of \say, with \see (\%).} mAUC is over the 24 evaluation sets; the next three columns are the mean mAUC over the 16 pairings of the evaluation with format decoupled from authenticity, the format gap and the worst pairing (defined in Sec.~5.4). Comb.: number of combinations. Names: with or without the 18 generator names.}
\label{tab:s22}
\fittab[\textwidth]{
\begin{tabular}{lllcccccccc}
\toprule
 &  &  & \multicolumn{4}{c}{MoE-ViE} & \multicolumn{4}{c}{PE} \\
\cmidrule(lr){4-7}\cmidrule(lr){8-11}
Paired descriptions & Comb. & Names & mAUC & 16 pairings & Format gap & Worst pairing & mAUC & 16 pairings & Format gap & Worst pairing \\
\midrule
Minimal pair & -- & -- & 93.09 & 92.32 & 4.52 & 87.56 & 94.89 & 93.78 & 4.84 & 88.29 \\
Condition-free & 1 & yes & 97.01 & 95.79 & 3.25 & 92.68 & 95.53 & 94.02 & 5.03 & 89.03 \\
Condition-free & 50 & yes & 97.16 & 96.39 & 2.94 & 93.53 & 96.02 & 94.84 & 4.05 & 91.54 \\
Image type only & 50 & yes & 98.01 & 97.11 & 2.64 & 94.52 & 96.53 & 95.53 & 3.27 & 93.01 \\
Content only & 50 & yes & 98.13 & 97.19 & 2.61 & 94.47 & 96.36 & 95.52 & 3.39 & 93.32 \\
Image type and content (Say) & 50 & yes & 98.15 & 97.19 & 2.48 & 94.57 & 96.80 & 95.66 & 2.80 & 93.56 \\
Condition-free & 50 & no & 97.33 & 96.33 & 2.69 & 93.56 & 95.50 & 94.24 & 3.74 & 89.85 \\
Image type and content & 50 & no & 97.93 & 96.87 & 2.41 & 94.40 & 96.13 & 94.76 & 2.76 & 91.29 \\
\bottomrule
\end{tabular}
}
\end{table*}

\begin{table*}[t]
\centering
\small
\caption{\textbf{Ablation of the design of \say, without \see (\%).} Columns as in \cref{tab:s22}.}
\label{tab:s23}
\fittab[\textwidth]{
\begin{tabular}{lllcccccccc}
\toprule
 &  &  & \multicolumn{4}{c}{MoE-ViE} & \multicolumn{4}{c}{PE} \\
\cmidrule(lr){4-7}\cmidrule(lr){8-11}
Paired descriptions & Comb. & Names & mAUC & 16 pairings & Format gap & Worst pairing & mAUC & 16 pairings & Format gap & Worst pairing \\
\midrule
Minimal pair & -- & -- & 93.48 & 93.07 & 4.05 & 88.52 & 95.72 & 94.08 & 6.56 & 84.78 \\
Condition-free & 1 & yes & 96.63 & 95.36 & 4.38 & 89.77 & 96.10 & 94.03 & 8.70 & 82.11 \\
Condition-free & 50 & yes & 96.93 & 95.75 & 5.40 & 88.28 & 96.19 & 94.53 & 7.68 & 84.39 \\
Image type only & 50 & yes & 97.87 & 96.30 & 5.01 & 89.01 & 96.46 & 94.80 & 6.81 & 86.12 \\
Content only & 50 & yes & 98.06 & 96.54 & 4.57 & 89.90 & 96.38 & 95.02 & 6.67 & 85.67 \\
Image type and content (Say) & 50 & yes & 97.99 & 96.51 & 4.43 & 90.00 & 96.34 & 94.92 & 5.81 & 87.58 \\
Condition-free & 50 & no & 96.93 & 95.45 & 3.41 & 91.38 & 95.71 & 94.40 & 6.17 & 86.34 \\
Image type and content & 50 & no & 97.66 & 96.24 & 3.66 & 91.14 & 96.13 & 94.48 & 6.05 & 87.28 \\
\bottomrule
\end{tabular}
}
\end{table*}

\paragraph{First step: templates and generator names.}
This step goes from the minimal pair to the condition-free set with one combination: each side moves from one description to a set of templates, and the generated side gains the generator names.
Without \see, it raises the mAUC over the 24 evaluation sets from 93.48\% to 96.63\% with MoE-ViE and from 95.72\% to 96.10\% with PE.
The mean mAUC over the 16 pairings rises from 93.07\% to 95.36\% with MoE-ViE and is essentially unchanged with PE (94.08\% and 94.03\%).

\paragraph{Second step: adding combinations.}
This step goes from one combination to 50, and the descriptions still state no condition.
In the complete method (\cref{tab:s22}), it improves all four metrics with both backbones.
With MoE-ViE, the mAUC over the 24 evaluation sets rises from 97.01\% to 97.16\%, the mean mAUC over the 16 pairings from 95.79\% to 96.39\% and the worst pairing from 92.68\% to 93.53\%, while the format gap falls from 3.25 to 2.94 points.
With PE, the same three metrics rise from 95.53\% to 96.02\%, from 94.02\% to 94.84\% and from 89.03\% to 91.54\%, and the format gap falls from 5.03 to 4.05 points.

\paragraph{Third step: stating the image type and the content.}
This step goes from the condition-free set with 50 combinations to \say: the number of combinations stays the same, and the descriptions of both sides now state the image type and the content.
In the complete method (\cref{tab:s22}), all four metrics improve further with both backbones.
With MoE-ViE, the mAUC over the 24 evaluation sets rises from 97.16\% to 98.15\%, the mean mAUC over the 16 pairings from 96.39\% to 97.19\% and the worst pairing from 93.53\% to 94.57\%, while the format gap falls from 2.94 to 2.48 points.
With PE, the mAUC over the 24 evaluation sets rises from 96.02\% to 96.80\%, the mean mAUC over the 16 pairings from 94.84\% to 95.66\% and the worst pairing from 91.54\% to 93.56\%, while the format gap falls from 4.05 to 2.80 points.
Stating only the image type or only the content already beats the condition-free set on all four metrics with both backbones.
Stating both gives the highest mAUC over the 24 evaluation sets, the highest worst pairing and the smallest format gap, again with both backbones.

Without \see (\cref{tab:s23}), the third step also improves all four metrics with both backbones.
With MoE-ViE, the mAUC over the 24 evaluation sets rises from 96.93\% to 97.99\%, the mean mAUC over the 16 pairings from 95.75\% to 96.51\% and the worst pairing from 88.28\% to 90.00\%, while the format gap falls from 5.40 to 4.43 points.
With PE, the same three metrics rise from 96.19\% to 96.34\%, from 94.53\% to 94.92\% and from 84.39\% to 87.58\%, and the format gap falls from 7.68 to 5.81 points.
The second step also raises the mAUC over the 24 evaluation sets and the mean mAUC over the 16 pairings with both backbones: from 96.63\% to 96.93\% and from 95.36\% to 95.75\% with MoE-ViE, and from 96.10\% to 96.19\% and from 94.03\% to 94.53\% with PE.
With PE, the condition-free set with one combination has the largest format gap (8.70 points) and the lowest worst pairing (82.11\%).
Adding combinations and then stating the conditions lower the format gap to 7.68 and 5.81 points, and \see brings it further down to 2.80 points.

Without generator names (the last two rows of both tables, both with 50 combinations), the third step with \see also improves all four metrics with both backbones.
With MoE-ViE, the mAUC over the 24 evaluation sets rises from 97.33\% to 97.93\%, the mean mAUC over the 16 pairings from 96.33\% to 96.87\% and the worst pairing from 93.56\% to 94.40\%, while the format gap falls from 2.69 to 2.41 points.
With PE, the same three metrics rise from 95.50\% to 96.13\%, from 94.24\% to 94.76\% and from 89.85\% to 91.29\%, and the format gap falls from 3.74 to 2.76 points.
Without \see, all four metrics improve with PE as well.
With MoE-ViE, the mAUC over the 24 evaluation sets and the mean mAUC over the 16 pairings rise (from 96.93\% to 97.66\% and from 95.45\% to 96.24\%), while the format gap and the worst pairing are slightly worse (from 3.41 to 3.66 points and from 91.38\% to 91.14\%).

\subsection{Design of \see}
\label{sec:supp-see}

\paragraph{Correcting one family of conditions.}
\cref{tab:s24} gives the results when only one family of conditions is corrected.
The format family alone removes most of the dependence on format: the format gap falls from 4.43 to 2.35 points with MoE-ViE and from 5.81 to 2.80 points with PE, and the worst pairing rises from 90.00\% to 94.45\% and from 87.58\% to 93.36\%.
With MoE-ViE, the gain in mAUC over the 24 evaluation sets comes from the image-type family (from 97.99\% to 98.21\%).
With PE, it comes mainly from the format family (from 96.34\% to 96.69\%), with a contribution from the content family.
The image-type family makes no correction with PE: the scores of non-photo image types are never above those of photographs, and only positive offsets are corrected (Sec.~4.3, $\tilde o=\max(o,0)$).
With all three families corrected, the mean mAUC over the 16 pairings is the highest with both backbones.

\begin{table*}[t]
\centering
\small
\caption{\textbf{Results when a single family of conditions is corrected (\%).} Columns as in \cref{tab:s22}.}
\label{tab:s24}
\fittab[\textwidth]{
\begin{tabular}{lcccccccc}
\toprule
 & \multicolumn{4}{c}{MoE-ViE} & \multicolumn{4}{c}{PE} \\
\cmidrule(lr){2-5}\cmidrule(lr){6-9}
Corrected conditions & mAUC & 16 pairings & Format gap & Worst pairing & mAUC & 16 pairings & Format gap & Worst pairing \\
\midrule
None (Say) & 97.99 & 96.51 & 4.43 & 90.00 & 96.34 & 94.92 & 5.81 & 87.58 \\
Format only & 97.96 & 96.89 & 2.35 & 94.45 & 96.69 & 95.49 & 2.80 & 93.36 \\
Image type only & 98.21 & 96.79 & 4.49 & 89.87 & 96.34 & 94.92 & 5.81 & 87.58 \\
Content only & 97.97 & 96.48 & 4.49 & 89.89 & 96.48 & 95.12 & 5.64 & 87.87 \\
All three (SaySee) & 98.15 & 97.19 & 2.48 & 94.57 & 96.80 & 95.66 & 2.80 & 93.56 \\
\bottomrule
\end{tabular}
}
\end{table*}

\paragraph{Number of reference photographs.}
\cref{tab:s25} varies the number of reference photographs of the format family while keeping the image-type and content families unchanged; for each number, the prototypes and offsets are re-estimated and the temperature is re-selected from the drawn photographs.
\see needs very few reference photographs.
With only 25, the format gap already falls from 4.43 to 2.50 points with MoE-ViE and from 5.81 to 2.70 points with PE, and the worst pairing rises from 90.00\% to 94.42\% and from 87.58\% to 92.96\%.
Beyond 100 photographs, none of the four metrics changes by more than 0.05 points with MoE-ViE.
With PE, the mAUC and the mean mAUC over the 16 pairings keep rising slightly with more photographs, to 96.89\% and 95.76\% at 300.
We further extend the reference photographs from 211 to 1,011, adding 800 photographs from the same sources and in the same proportions as the original 211.
The four metrics then change by at most 0.05 points with MoE-ViE and 0.23 points with PE, and with PE the mean mAUC over the 16 pairings rises from 95.66\% to between 95.76\% and 95.82\%.
With more photographs, the offsets are estimated more accurately: the root mean square of the corrected offsets over the 132 non-reference conditions (\cref{sec:supp-bias}) falls from $0.13\sigma$ to $0.08\sigma$ with MoE-ViE and from $0.11\sigma$ to $0.10\sigma$ with PE, while the detection metrics stay level.
The 211 photographs used by \saysee already lie on this plateau.

\begin{table*}[t]
\centering
\small
\caption{\textbf{Number of reference photographs of the format family (\%).} Columns as in \cref{tab:s22}. 0 photographs is \say alone; 211 is the number used by \saysee, \ie, all 291 photographs of the format family except the 80 held out (\cref{sec:supp-real}); 1,011 is these 211 plus the 800 new photographs; the other rows are means over 5 random draws: below 211, the photographs are drawn from these 211; above 211, these 211 are kept and new photographs are added.}
\label{tab:s25}
\fittab[\textwidth]{
\begin{tabular}{lcccccccc}
\toprule
 & \multicolumn{4}{c}{MoE-ViE} & \multicolumn{4}{c}{PE} \\
\cmidrule(lr){2-5}\cmidrule(lr){6-9}
Photos & mAUC & 16 pairings & Format gap & Worst pairing & mAUC & 16 pairings & Format gap & Worst pairing \\
\midrule
0 (Say) & 97.99 & 96.51 & 4.43 & 90.00 & 96.34 & 94.92 & 5.81 & 87.58 \\
25 & 98.05 & 97.09 & 2.50 & 94.42 & 96.73 & 95.62 & 2.70 & 92.96 \\
50 & 98.05 & 97.11 & 2.46 & 94.48 & 96.65 & 95.52 & 2.78 & 93.41 \\
100 & 98.11 & 97.15 & 2.49 & 94.54 & 96.73 & 95.59 & 2.82 & 93.58 \\
150 & 98.12 & 97.16 & 2.49 & 94.55 & 96.82 & 95.65 & 2.97 & 93.59 \\
211 (SaySee) & 98.15 & 97.19 & 2.48 & 94.57 & 96.80 & 95.66 & 2.80 & 93.56 \\
300 & 98.13 & 97.17 & 2.49 & 94.56 & 96.89 & 95.76 & 2.87 & 93.76 \\
400 & 98.13 & 97.15 & 2.49 & 94.54 & 96.85 & 95.77 & 2.86 & 93.74 \\
600 & 98.12 & 97.14 & 2.50 & 94.55 & 96.80 & 95.76 & 2.91 & 93.69 \\
800 & 98.12 & 97.14 & 2.52 & 94.54 & 96.84 & 95.82 & 2.93 & 93.74 \\
1,011 & 98.12 & 97.15 & 2.52 & 94.56 & 96.83 & 95.82 & 2.93 & 93.79 \\
\bottomrule
\end{tabular}
}
\end{table*}

\paragraph{Temperature.}
The temperature of each family is selected from 15 fixed candidate values, using only real images that are not used for the prototypes being evaluated (\cref{sec:supp-real}).
\cref{tab:s26} varies the temperature of one family at a time, with the other two at their selected values.
\saysee is not sensitive to the temperature.
For $\lambda\ge0.02$, varying the temperature of the format family changes the mAUC by at most 0.02 points with MoE-ViE and keeps it between 96.54\% and 96.95\% with PE, and the format gap does not exceed 2.48 and 2.97 points.
For very small $\lambda$, the posterior tends to uniform, the correction of the format family stops working, and the format gap and the worst pairing return to levels close to those of \say.
The temperature of the image-type family moves the mAUC with MoE-ViE between 97.95\% and 98.15\%; with PE, the offsets of this family are all zero and the temperature has no effect.
The temperature of the content family changes the mAUC by at most 0.02 points with MoE-ViE and 0.12 points with PE.

\begin{table*}[t]
\centering
\small
\caption{\textbf{Sensitivity to the temperature (\%).} For the format family, each cell gives the mAUC over the 24 evaluation sets, the mean mAUC over the 16 pairings, the format gap and the worst pairing; for the image-type and content families, the mAUC over the 24 evaluation sets. Selected values are in bold.}
\label{tab:s26}
\fittab[\textwidth]{
\begin{tabular}{cllllll}
\toprule
 & \multicolumn{3}{c}{MoE-ViE} & \multicolumn{3}{c}{PE} \\
\cmidrule(lr){2-4}\cmidrule(lr){5-7}
$\lambda$ & Format & Image type & Content & Format & Image type & Content \\
\midrule
5 & 98.13 / 97.19 / 2.44 / 94.62 & 98.12 & 98.14 & 96.54 / 95.41 / 2.64 / 93.31 & \textbf{96.80} & 96.81 \\
2 & 98.13 / 97.19 / 2.44 / 94.62 & 98.12 & 98.14 & 96.56 / 95.43 / 2.63 / 93.34 & 96.80 & 96.81 \\
1 & 98.13 / 97.20 / 2.44 / 94.62 & 98.12 & 98.15 & 96.58 / 95.44 / 2.63 / 93.37 & 96.80 & 96.81 \\
0.5 & 98.14 / 97.20 / 2.45 / 94.62 & 98.12 & 98.15 & 96.62 / 95.49 / 2.64 / 93.41 & 96.80 & 96.81 \\
0.2 & 98.15 / 97.21 / 2.45 / 94.62 & 98.12 & \textbf{98.15} & 96.70 / 95.58 / 2.68 / 93.49 & 96.80 & \textbf{96.80} \\
0.1 & 98.15 / 97.20 / 2.46 / 94.60 & 98.13 & 98.16 & \textbf{96.80 / 95.66 / 2.80 / 93.56} & 96.80 & 96.77 \\
0.05 & \textbf{98.15 / 97.19 / 2.48 / 94.57} & 98.15 & 98.16 & 96.92 / 95.74 / 2.97 / 93.72 & 96.80 & 96.73 \\
0.02 & 98.14 / 97.19 / 2.39 / 94.54 & \textbf{98.15} & 98.16 & 96.95 / 95.80 / 2.85 / 92.63 & 96.80 & 96.70 \\
0.01 & 98.15 / 97.10 / 2.87 / 93.02 & 98.11 & 98.16 & 96.78 / 95.63 / 3.68 / 90.63 & 96.80 & 96.69 \\
0.005 & 98.17 / 96.94 / 3.73 / 91.33 & 98.04 & 98.16 & 96.63 / 95.40 / 4.73 / 89.16 & 96.80 & 96.69 \\
0.002 & 98.19 / 96.83 / 4.26 / 90.32 & 97.98 & 98.16 & 96.54 / 95.23 / 5.32 / 88.33 & 96.80 & 96.69 \\
0.001 & 98.19 / 96.80 / 4.42 / 90.02 & 97.96 & 98.16 & 96.51 / 95.18 / 5.48 / 88.10 & 96.80 & 96.69 \\
0.0005 & 98.19 / 96.78 / 4.49 / 89.89 & 97.95 & 98.16 & 96.50 / 95.15 / 5.56 / 87.98 & 96.80 & 96.69 \\
0.0002 & 98.20 / 96.77 / 4.53 / 89.81 & 97.95 & 98.16 & 96.49 / 95.13 / 5.61 / 87.91 & 96.80 & 96.69 \\
0.0001 & 98.20 / 96.77 / 4.54 / 89.79 & 97.95 & 98.16 & 96.49 / 95.13 / 5.63 / 87.89 & 96.80 & 96.69 \\
\bottomrule
\end{tabular}
}
\end{table*}

\begin{table*}[t]
\centering
\small
\caption{\textbf{AUC (\%) of the minimal pair and \say on each evaluation set with CLIP ViT-L/14.} Without \see.}
\label{tab:s27}
\fittab[\textwidth]{
\begin{tabular}{lcclcc}
\toprule
Set & Minimal pair & Say & Set & Minimal pair & Say \\
\midrule
GenImage & 74.7 & 76.1 & CO-SPY-Wild & 69.8 & 21.0 \\
AIGCDB & 81.2 & 81.8 & OpenFake-Reddit & 35.2 & 34.2 \\
DRCT-2M & 65.8 & 68.4 & RRDataset & 51.2 & 56.4 \\
Synthbuster & 81.4 & 56.0 & HiRes-50K & 46.8 & 46.3 \\
GenImage++ & 66.2 & 59.0 & DailyBench & 60.5 & 70.3 \\
B-Free-NG & 71.5 & 50.2 & So-Fake-OOD & 56.3 & 64.9 \\
SynthCLIC & 48.5 & 57.5 & OpenFake-Test & 41.2 & 49.9 \\
Chameleon & 53.7 & 43.8 & AIGenImages2026 & 63.1 & 60.6 \\
WildRF & 47.8 & 70.5 & AIGI-Now & 51.7 & 48.1 \\
SocialRF & 46.8 & 68.2 & MLLMGenSet & 67.9 & 70.4 \\
CommunityAI & 56.5 & 61.3 & RealChain & 45.8 & 39.2 \\
B-Free-Online & 22.2 & 21.5 & Mirage-Test & 31.9 & 36.6 \\
mAUC & 55.7 & 54.7 &  &  &  \\
\bottomrule
\end{tabular}
}
\end{table*}

\subsection{An earlier backbone}
\label{sec:supp-backbone}

\saysee reads a separation that the backbone already has (Sec.~3.1).
CLIP ViT-L/14 is an earlier backbone that lacks it: its mAUC over the 24 evaluation sets is close to chance for both the minimal pair and \say, 55.7\% and 54.7\% (\cref{tab:s27}), while the same descriptions reach 96.34\% with PE and 97.99\% with MoE-ViE.
This is in line with the observation of Sec.~3.1.

\section{Measuring and removing bias}
\label{sec:supp-bias}

\paragraph{Source and size of the bias.}
The bias of a trained detector comes from its training set: within one training set, real and generated images differ in format and content in a fixed way, and the detector learns this difference as evidence (\cref{sec:supp-probe-format}).
The bias of the generation direction comes instead from the image--text pairs used to pretrain the backbone (Sec.~3.2).
These pairs cover far more generators, formats, image types and contents than any detection training set, and the bias left in the direction is smaller.
On the same backbone and the same image embedding, the format gap of \say is 4.43 (MoE-ViE) and 5.81 (PE) points, compared with 5.36 to 44.61 and 8.82 to 33.26 points for the four trained methods of \cref{sec:supp-training} (\cref{tab:s15}).
This bias appears as a shift in the scores of real images, so real images alone suffice to measure and remove it: 25 reference photographs already lower the format gap to 2.50 and 2.70 points (\cref{tab:s25}), and the correction carries over to codecs that \see has never seen (\cref{sec:supp-unseen}).
After the correction, \saysee has a smaller format gap and a higher worst pairing than the linear probe with the same correction (\cref{sec:supp-probe-see}).

\begin{figure*}[t]
\centering
\includegraphics[width=\textwidth]{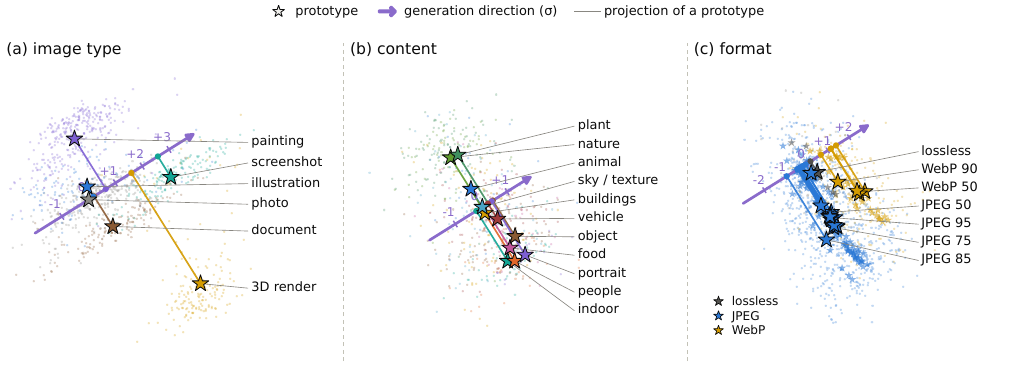}
\caption{\textbf{Prototypes of \see along the generation direction (MoE-ViE).} Each panel is a plane that contains the generation direction of \say (light purple line); the other axis is the direction orthogonal to it along which the prototypes differ most, on a compressed scale. Ticks on the line are in units of $\sigma$, with 0 at the reference condition. Stars are prototypes, and thin lines are their perpendiculars to the generation direction. Small light dots are the real images of each condition, in the color of their prototype. (a) The 6 prototypes of image type. (b) The 11 prototypes of content; 0 is the mean of all photographs. (c) Format: small stars are the 133 prototypes, and large stars the mean over the 7 sizes of one codec.}
\label{fig:s05}
\end{figure*}

\paragraph{Prototypes of \see.}
\cref{fig:s05} shows where the prototypes of the three families lie along the generation direction.
The foot of the perpendicular from a prototype to the generation direction gives its offset from the reference condition, \ie, the quantity that \see corrects.
With MoE-ViE, the screenshot and 3D render prototypes lie farthest among the image types ($+2.6\sigma$ and $+1.6\sigma$), paintings lie at $+0.6\sigma$, and illustrations and document pages are close to photos.
The prototypes of the content themes all lie between $-0.19\sigma$ and $+0.45\sigma$.
Among formats, the WebP prototypes fall on the positive side, farther for lower quality, and most JPEG prototypes fall on the negative side.

\begin{figure*}[t]
\centering
\includegraphics[width=\textwidth]{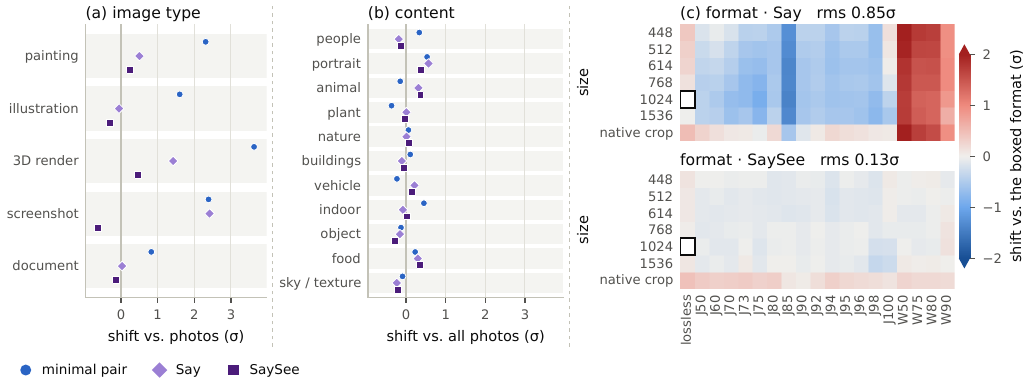}
\caption{\textbf{Shifts of the scores of real images over the three families of conditions, after \say and after \see (MoE-ViE).} The panels correspond to the top row of Fig.~3 and use the same unit. (a) Image type and (b) content: each row gives the shifts of the three scores, minimal pair, \say and \saysee. (c) Format: offsets under the 133 conditions for \say (top) and \saysee (bottom).}
\label{fig:s06}
\end{figure*}

\paragraph{Removing the bias.}
\cref{fig:s06} shows the score shifts of real images after \say and after \see for the three families of Fig.~3.
For image type, \say removes most of the shift (\cref{tab:s28}).
Relative to photos, paintings go from $2.31\sigma$ to $0.51\sigma$, illustrations from $1.60\sigma$ to $-0.05\sigma$ and 3D renders from $3.63\sigma$ to $1.42\sigma$.
Screenshots are still shifted by $2.42\sigma$ after \say, and \see corrects this to $-0.62\sigma$.
The content shifts are small to begin with, within $0.6\sigma$ under all three scores.
For format, the root mean square of the offsets over the 132 non-reference conditions is $0.85\sigma$ with \say alone and $0.13\sigma$ after \see, and the share of conditions with an absolute offset above $0.2\sigma$ drops from 84\% to 12\%.

\begin{table}[tb]
\centering
\small
\caption{\textbf{Shifts of the scores of held-out real images relative to real photographs along the generation directions of the minimal pair and of \say ($\sigma$).} Procedure and unit as in the top row of Fig.~3; without \see.}
\label{tab:s28}
\fittab{
\begin{tabular}{lcccc}
\toprule
 & \multicolumn{2}{c}{MoE-ViE} & \multicolumn{2}{c}{PE} \\
\cmidrule(lr){2-3}\cmidrule(lr){4-5}
Image type & Minimal pair & Say & Minimal pair & Say \\
\midrule
Paintings & +2.31 & +0.51 & $-$0.23 & $-$1.10 \\
Illustrations & +1.60 & $-$0.05 & $-$0.67 & $-$1.93 \\
3D renders & +3.63 & +1.42 & $-$0.29 & $-$0.09 \\
Screenshots & +2.39 & +2.42 & $-$0.43 & $-$0.15 \\
Document pages & +0.83 & +0.03 & $-$3.09 & $-$1.41 \\
\bottomrule
\end{tabular}
}
\end{table}

\begin{figure*}[t]
\centering
\includegraphics[width=\textwidth]{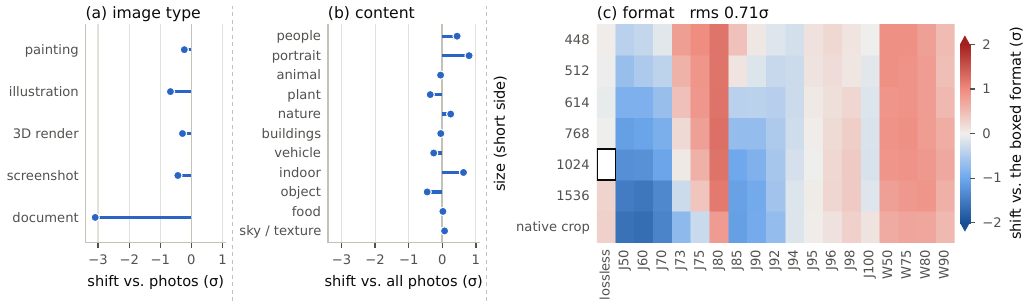}
\caption{\textbf{Bias of the generation direction of the minimal pair with PE.} Procedure and unit as in the top row of Fig.~3.}
\label{fig:s07}
\end{figure*}

\begin{figure*}[t]
\centering
\includegraphics[width=\textwidth]{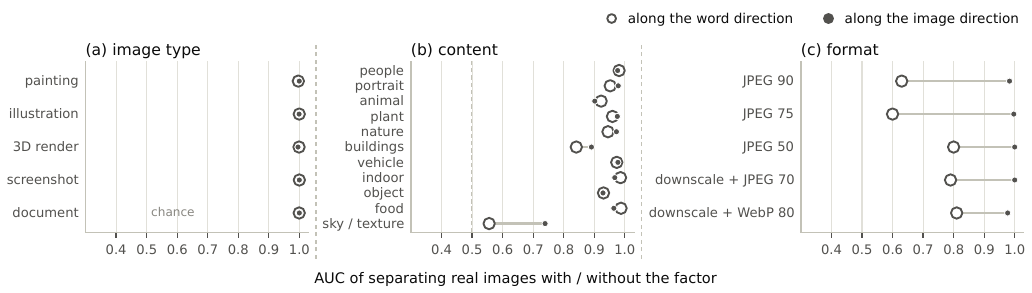}
\caption{\textbf{Whether language can express the three kinds of conditions (PE).} As in the bottom row of Fig.~3.}
\label{fig:s08}
\end{figure*}

\begin{figure*}[t]
\centering
\includegraphics[width=\textwidth]{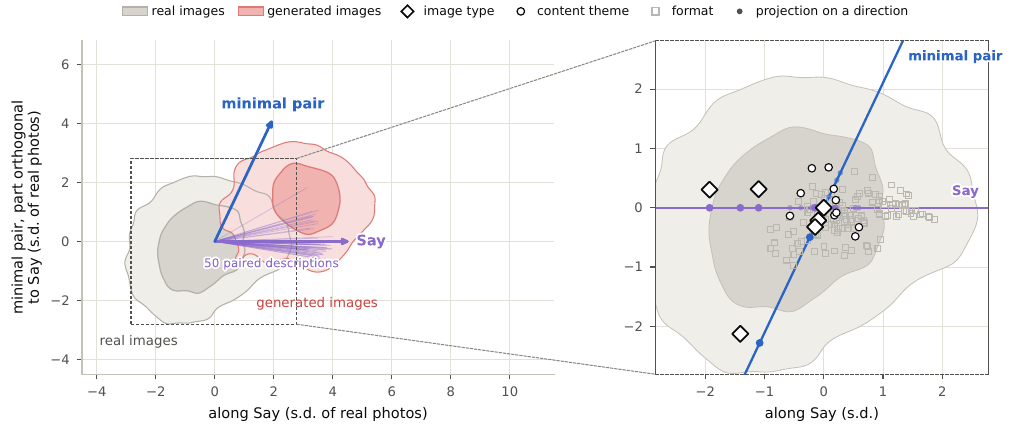}
\caption{\textbf{What pairing changes in the generation direction (PE).} As in Fig.~5.}
\label{fig:s09}
\end{figure*}

\begin{figure*}[t]
\centering
\includegraphics[width=\textwidth]{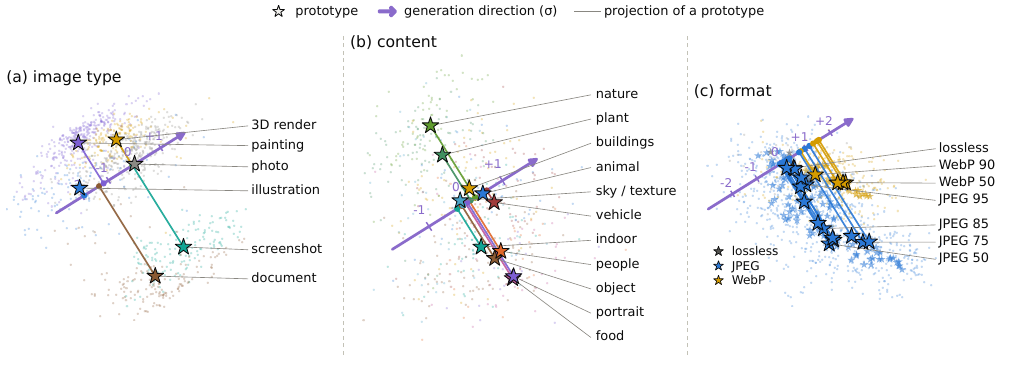}
\caption{\textbf{Prototypes of \see along the generation direction (PE).} As in \cref{fig:s05}.}
\label{fig:s10}
\end{figure*}

\begin{figure*}[t]
\centering
\includegraphics[width=\textwidth]{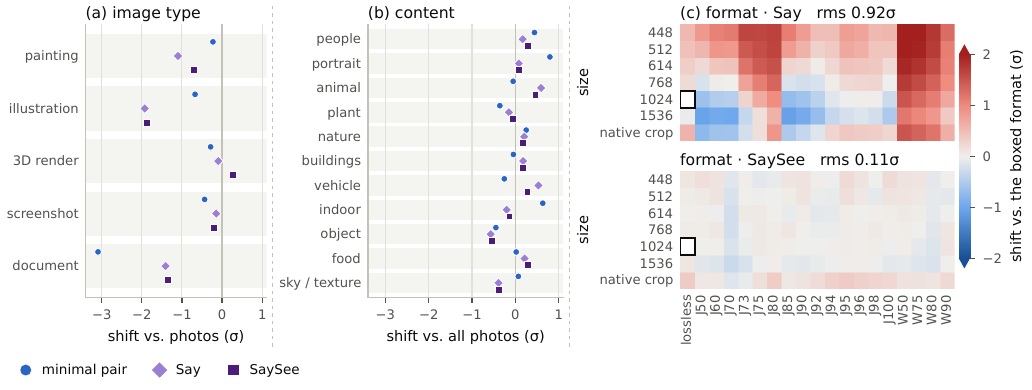}
\caption{\textbf{Shifts of the scores of real images over the three families of conditions, after \say and after \see (PE).} As in \cref{fig:s06}.}
\label{fig:s11}
\end{figure*}

\paragraph{Results with PE.}
\cref{fig:s07,fig:s08,fig:s09,fig:s10,fig:s11} are the PE versions of Figs.~3 and~5 of the main paper and of \cref{fig:s05,fig:s06}.
PE differs from MoE-ViE in the image type.
Along the directions of both the minimal pair and \say, non-photo images all score below photos with PE (\cref{tab:s28}), and the prototypes of the five image types lie between $-1.9\sigma$ and $-0.08\sigma$, so the image-type family makes no correction.
The format family behaves as with MoE-ViE: the root mean square of the offsets falls from $0.92\sigma$ with \say to $0.11\sigma$ with \saysee, and the share of conditions with an absolute offset above $0.2\sigma$ falls from 85\% to 8\%.

\begin{figure*}[t]
\centering
\includegraphics[width=\textwidth]{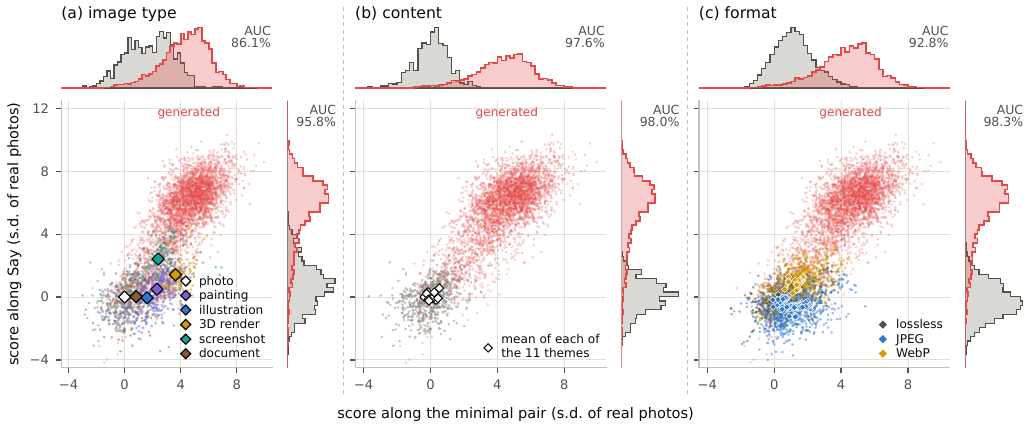}
\caption{\textbf{After \say, real and generated images are better separated under every family of conditions (MoE-ViE).} The horizontal axis is the score along the direction of the minimal pair and the vertical axis the score along the generation direction of \say, both with the mean of real photographs as zero and their standard deviation as unit. Red dots are the generated images of the 24 evaluation sets; the others are real images: (a) the 6 image types; (b) photographs of the 11 content themes; (c) the versions of 80 photographs under the 133 format conditions, colored by codec. Diamonds are the means of the conditions. The margins show the distributions of the two kinds of images along the two directions.}
\label{fig:s12}
\end{figure*}

\begin{figure*}[t]
\centering
\includegraphics[width=\textwidth]{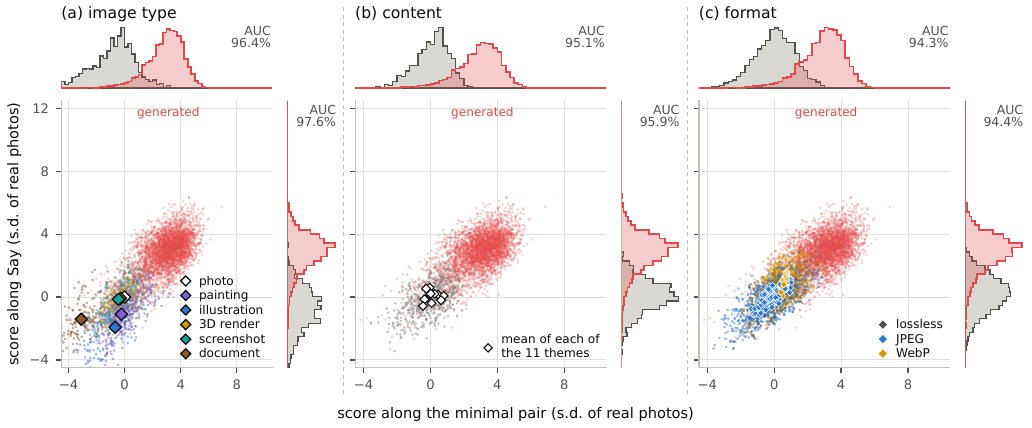}
\caption{\textbf{PE version of \cref{fig:s12}.}}
\label{fig:s13}
\end{figure*}

\paragraph{Separability within each family of conditions.}
\cref{fig:s12,fig:s13} compare the minimal pair and \say from another angle.
For the real images of each condition, we compute the AUC between the generated images of the 24 evaluation sets and these real images, and average over the conditions.
\say separates real from generated images better for all three families with MoE-ViE, and for the image-type and content families with PE.
The format family with PE is essentially unchanged (94.3\% and 94.4\%), and its bias is corrected by \see.
With MoE-ViE, the image-type family rises from 86.1\% to 95.8\% (3D renders from 67.3\% to 94.4\%), the format family from 92.8\% to 98.3\%, and the content family from 97.6\% to 98.0\%.
With PE, the image-type and content families rise from 96.4\% and 95.1\% to 97.6\% and 95.9\%.

\clearpage\suppfullcol
\section{Results on a Huawei Ascend NPU}
\label{sec:supp-npu}

\saysee requires no training.
At test time, a frozen backbone computes the image embedding, which is projected, corrected by \see and compared with a fixed threshold.
Operator implementations and numerical precision differ across hardware, so image embeddings and scores may differ slightly from one device to another.
If the scores drifted across devices, the fixed threshold would have to be set again on each device, and \saysee could not be deployed as is on other computing platforms.
To test this, we ran \saysee again on a Huawei Ascend NPU and compared the results with those on an NVIDIA RTX PRO 6000, set by set.

\cref{tab:s29} gives the AUC of \saysee on each evaluation set on the two devices.
The bAcc at the fixed threshold is the same on both devices, the mAUC differs by at most 0.1 points, and the AUC of each evaluation set differs by at most 0.3 points.

\begin{table}[H]
\centering
\small
\caption{\textbf{AUC (\%) of \saysee on each evaluation set on an NVIDIA RTX PRO 6000 and on a Huawei Ascend NPU.} The last row is the bAcc at the fixed threshold.}
\label{tab:s29}
\fittab{
\begin{tabular}{lcccc}
\toprule
 & \multicolumn{2}{c}{MoE-ViE} & \multicolumn{2}{c}{PE} \\
\cmidrule(lr){2-3}\cmidrule(lr){4-5}
Set & RTX PRO 6000 & Ascend NPU & RTX PRO 6000 & Ascend NPU \\
\midrule
GenImage & 99.6 & 99.6 & 99.3 & 99.3 \\
AIGCDB & 99.7 & 99.6 & 99.3 & 99.3 \\
DRCT-2M & 100.0 & 100.0 & 98.7 & 98.7 \\
Synthbuster & 100.0 & 100.0 & 99.9 & 99.9 \\
GenImage++ & 99.9 & 99.9 & 99.8 & 99.8 \\
B-Free-NG & 100.0 & 100.0 & 100.0 & 100.0 \\
SynthCLIC & 98.7 & 98.6 & 97.3 & 97.1 \\
Chameleon & 99.5 & 99.5 & 97.1 & 96.9 \\
WildRF & 99.7 & 99.7 & 99.4 & 99.4 \\
SocialRF & 99.3 & 99.3 & 98.5 & 98.5 \\
CommunityAI & 99.8 & 99.8 & 98.5 & 98.5 \\
B-Free-Online & 99.4 & 99.4 & 92.2 & 92.3 \\
CO-SPY-Wild & 100.0 & 100.0 & 98.1 & 98.1 \\
OpenFake-Reddit & 95.4 & 95.5 & 94.0 & 93.8 \\
RRDataset & 96.7 & 96.7 & 96.1 & 96.1 \\
HiRes-50K & 97.5 & 97.5 & 94.7 & 94.7 \\
DailyBench & 95.7 & 95.5 & 87.7 & 87.5 \\
So-Fake-OOD & 98.2 & 98.2 & 97.3 & 97.3 \\
OpenFake-Test & 99.4 & 99.4 & 99.4 & 99.4 \\
AIGenImages2026 & 96.3 & 96.3 & 94.1 & 94.1 \\
AIGI-Now & 98.8 & 98.8 & 99.2 & 99.2 \\
MLLMGenSet & 87.5 & 87.2 & 86.6 & 86.5 \\
RealChain & 96.1 & 96.1 & 98.0 & 98.0 \\
Mirage-Test & 98.1 & 98.1 & 98.3 & 98.3 \\
mAUC & 98.2 & 98.1 & 96.8 & 96.8 \\
bAcc & 93.7 & 93.7 & 91.3 & 91.3 \\
\bottomrule
\end{tabular}
}
\end{table}

\FloatBarrier

\end{document}